\documentclass{article}
\usepackage{iclr2027_conference,times}

\usepackage{amsmath,amsfonts,bm}

\def\eqref#1{equation~\ref{#1}}

\def\plaineqref#1{\ref{#1}}

\def\1{\bm{1}}

\DeclareMathAlphabet{\mathsfit}{\encodingdefault}{\sfdefault}{m}{sl}
\SetMathAlphabet{\mathsfit}{bold}{\encodingdefault}{\sfdefault}{bx}{n}

\usepackage{colortbl}
\usepackage{hyperref}
\usepackage{url}
\usepackage{booktabs}
\usepackage{graphicx}
\usepackage{float}
\usepackage{multirow}
\usepackage{longtable}
\usepackage{etoolbox}
\usepackage{placeins}

\makeatletter
\pretocmd{\@makecaption}{\normalsize}{}{\errmessage{Caption font patch failed}}
\patchcmd{\LT@makecaption}{\reset@font}{\reset@font\normalsize}{}{\errmessage{Longtable caption font patch failed}}
\makeatother

\definecolor{improw}{gray}{0.88}

\newcommand{\termreview}[1]{#1}

\newcommand{\revisionblue}[1]{#1}

\title{MarkovTS: Pretrained State--Transition Representations for
Zero-Shot Time Series Generation}

\author{%
\textbf{Yujie Li}$^{1,2}$, \textbf{Zezhi Shao}$^{1}$,
\textbf{Jingbo Zhou}$^{3}$, \textbf{Yongjun Xu}$^{1,2}$,
\textbf{Xueqi Cheng}$^{1,2}$ \& \textbf{Fei Wang}$^{1,2}$\thanks{Corresponding author: Fei Wang.} \\
\normalfont $^{1}$State Key Laboratory of AI Safety, Institute of Computing Technology, \\
\normalfont Chinese Academy of Sciences, Beijing, China \\
\normalfont $^{2}$University of Chinese Academy of Sciences, Beijing, China \\
\normalfont $^{3}$Baidu Frontier Research Department, Beijing, China \\
\normalfont\texttt{\{liyujie23s,shaozezhi,xyj,cxq,wangfei\}@ict.ac.cn} \\
\normalfont\texttt{zhoujingbo@baidu.com}
}

\hypersetup{
  hidelinks,
  pdftitle={MarkovTS: Pretrained State-Transition Representations for Zero-Shot Time Series Generation},
  pdfauthor={Yujie Li, Zezhi Shao, Jingbo Zhou, Yongjun Xu, Xueqi Cheng, Fei Wang}
}

\iclrfinalcopy
\begin{document}

\maketitle
\fancyhead{}

\begin{abstract}
Across heterogeneous real-world domains, time-series generation supports simulation, model development, and data augmentation by generating sequences that preserve target-domain statistics and temporal dependencies when real observations are scarce or costly. However, existing methods predominantly address in-domain generation, limiting their reuse in unseen domains. We study reference-conditioned zero-shot time series generation: target domains are unseen during pretraining, generation uses a single target-domain reference sequence, and no model parameters are updated. MarkovTS characterizes sequence evolution through local morphologies and their ordered transitions at multiple temporal scales. Its pretrained encoder quantizes normalized time-series patches into multiscale morphological states with a state codebook, while a transition codebook encodes \revisionblue{contextual transitions between adjacent intervals}. These transferable state--transition representations condition a pretrained flow-matching model. At generation, the frozen model combines a reference-derived condition with independent Gaussian noise to produce stochastic samples that retain the reference's temporal organization. Across 12 unseen target datasets, MarkovTS outperforms strong baselines in most distributional comparisons and generates diverse, reference-consistent sequences. As downstream augmentation, its samples yield overall gains across forecasting, classification, and anomaly detection benchmarks and diverse model families.
\mbox{Our code is available at \url{https://github.com/blisky-li/MarkovTS}.}
\end{abstract}

\section{Introduction}
\label{sec:introduction}

Time-series generation estimates the distribution of observed sequences and generates new sequences with corresponding statistical properties and temporal dependencies. It is particularly valuable when representative measurements are scarce, costly to collect, or difficult to share across healthcare, energy, transportation, finance, and environmental monitoring \citep{yoon2019timegan,li2023crvae,gonen2025imagenfew}. The resulting generated time series support simulation, model development, and downstream data augmentation \citep{yuan2024diffusionts,huang2025timedp}.

\begin{figure}[t]
  \centering
  \includegraphics[width=0.95\textwidth,pagebox=cropbox]{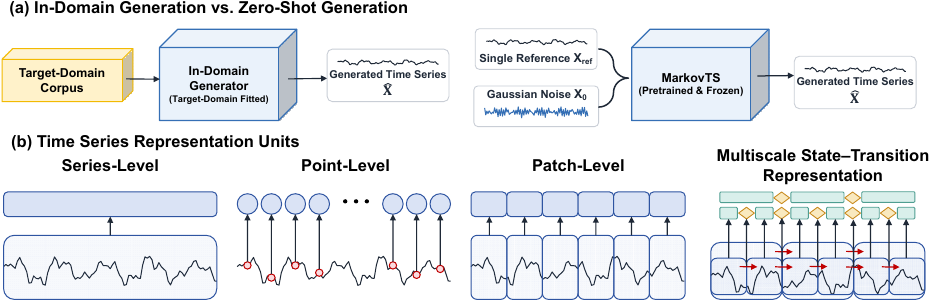}
  \caption{Generation settings and representation units.
  	\textbf{(a)} In-domain training versus single-reference generation with frozen MarkovTS.
  	\textbf{(b)} Series-, point-, and patch-level units versus the proposed multiscale state-transition representation.}
  \label{fig:intro-motivation}
\end{figure}

\revisionblue{Existing generators typically acquire target-specific temporal structure from collections of target-domain sequences (Figure~\ref{fig:intro-motivation}(a)). Dataset-specific models fit their parameters to these observations \citep{yoon2019timegan,desai2021timevae,yuan2024diffusionts}, while pretrained generators use target examples for domain conditioning or few-shot adaptation \citep{huang2025timedp,gonen2025imagenfew,yao2026timemode}. Reusing pretrained parameters thus leaves the challenge of obtaining sufficient temporal information from a single reference in an unseen domain.}

\noindent\begin{minipage}{\linewidth} Real-world scenarios often lack sufficient data for model fitting and instead call for generating additional task-specific samples directly from a single \termreview{reference}. We term this \termreview{zero-shot time series generation}, in which the generator must capture and exploit transferable temporal structure within the reference without target-domain parameter updates.
\end{minipage}

\revisionblue{Representation granularity determines which temporal information can be transferred from a single reference (Figure~\ref{fig:intro-motivation}(b)). Global sequence codes summarize the entire series but offer limited flexibility across lengths and complexities \citep{fabius2014vrae,desai2021timevae}; point-level states preserve numerical detail but require costly contextual modeling \citep{yoon2019timegan}. \mbox{Fixed-scale} patches provide a local compromise, yet expose only one temporal scale \citep{zhou2026winformer}.} Beyond granularity, tokenization and its learning objectives \citep{talukder2024totem,wen2024vqshape,tao2025tokencast} raise a further concern: downstream objectives may bias tokenizers toward local morphology while neglecting sequence evolution. \revisionblue{A representation that captures local shapes but omits how they evolve provides an incomplete condition for reproducing the reference's temporal organization.}

\revisionblue{To address these limitations,} we propose \textbf{MarkovTS}, combining a \termreview{Markov-inspired} multiscale state--transition encoder with a conditional flow-matching generator for zero-shot time series generation. The pretrained encoder uses a \termreview{state codebook} to quantize normalized multiscale patches into morphological states and a \termreview{transition codebook} to encode features between continuous patch-pairs and corresponding state-pairs. Reference-derived codebook vectors condition the pretrained generator to control temporal organization, while independent Gaussian sources provide stochastic variation. Together, local morphology and contextual transitions provide a reference-specific condition for controlling temporal organization during zero-shot transfer to unseen domains. Our contributions are as follows:
\begin{itemize}
\item We formalize single-reference zero-shot time series generation as transferring temporal organization from one unseen-domain reference without target-domain parameter updates.
\item \revisionblue{We propose MarkovTS, which couples a transferable multiscale state--transition representation of local morphology and contextual transitions with conditional flow matching to preserve stochastic variation across generated samples.}
\item Experiments across 12 unseen target datasets show strong distributional matching, while fixed-noise reference swaps show that generated outputs follow the selected joint condition. The generated data also support forecasting, classification, and anomaly detection.
\end{itemize}

\section{Related Work and Background}
\label{sec:related-work}

\subsection{Time-Series Generation}

Time-series generators model temporal dependence through adversarial learning \citep{yoon2019timegan,xu2020cotgan,jeon2022gtgan,seyfi2022coscigan}, continuous or discrete latent dynamics \citep{desai2021timevae,li2023crvae,zhou2023ls4,lee2023timevqvae}, and iterative denoising \citep{yuan2024diffusionts,cao2024timedit}.  TimeDP learns a multi-domain diffusion prior and derives a domain prompt from target examples without parameter updates~\citep{huang2025timedp}; Winformer emphasizes window-level periodic alignment~\citep{zhou2026winformer}. ImagenFew instead adapts pretrained diffusion models through target-domain few-shot fine-tuning \citep{gonen2025imagenfew,li2025sta}.

Representation units also differ in temporal granularity. Sequence-level models use global latent codes \citep{fabius2014vrae,desai2021timevae}, whereas TimeGAN uses recurrent point-level states \citep{yoon2019timegan}. ImagenFew maps observations into invertible image representations \citep{gonen2025imagenfew}. Patch-level models aggregate local subseries and model inter-patch context; Winformer aligns periodic windows \citep{zhou2026winformer}, while TimeVQVAE quantizes low- and high-frequency features into local time-frequency tokens with bidirectional priors \citep{lee2023timevqvae}.

Collapse toward morphology in time-series representations is subtle because in-domain tasks or training objectives place limited emphasis on complex \revisionblue{evolution}. VQShape and UniShape learn generalizable morphological representations for classification \citep{wen2024vqshape,liu2026unishape}. Morphology-based forecasting improves performance \citep{yu2025geometric,lee2022pmmemnet,shao2025blast,wang2025aries} largely on smaller \revisionblue{in-domain} corpora, where other components can supply temporal context.
\revisionblue{Together, QUALS, ReCast, and LatentTSF highlight the risk that task-driven optimization favors a limited subset of temporal information, encouraging shortcut learning and representation degradation at the expense of transferable temporal structure~\citep{li2027quals, ma2026recast,yang2026latenttsf}.}

\subsection{Flow Matching}

Flow matching trains a continuous normalizing flow by regressing a vector field that transports a simple source distribution to data distribution. Its simulation-free regression objective avoids integrating the flow during training and supports non-diffusion paths, including linear conditional paths that facilitate efficient sampling \citep{lipman2023flowmatching}. Let $X_0\sim p_0=\mathcal{N}(0,I)$ be a source sample, $X_1\sim q_{\mathrm{data}}$ a real time series, $t\sim\mathcal{U}(0,1)$. For the linear conditional path used in this work,
\begin{equation}
    X_t=(1-t)X_0+tX_1,
    \qquad
    U_t=\frac{\mathrm{d}X_t}{\mathrm{d}t}=X_1-X_0.
    \label{eq:fm-path}
\end{equation}
Given an optional condition $C$, a velocity network $v_\theta$ is optimized by
\begin{equation}
    \mathcal{L}_{\mathrm{FM}}(\theta)
    =\mathbb{E}_{X_0,X_1,t}
    \left[\left\|v_\theta(X_t,t;C)-U_t\right\|_2^2\right].
    \label{eq:fm-objective}
\end{equation}
After training, generation starts from $\widetilde{X}_0\sim p_0$ and solves
\begin{equation}
    \frac{\mathrm{d}\widetilde{X}_t}{\mathrm{d}t}
    =v_\theta(\widetilde{X}_t,t;C),
    \qquad
    \widehat{X}=\widetilde{X}_1.
    \label{eq:fm-sampling}
\end{equation}
In MarkovTS, $C$ is instantiated by the multiscale state and transition codes extracted from a reference time series; Section~\ref{sec:method} specifies how these codes are learned and injected into the velocity network.

\section{Problem Formulation}
\label{sec:problem-formulation}

Let $\mathcal{D}_{\mathrm{pre}}=\bigcup_{i=1}^{M}\mathcal{D}_i$ be a pretraining corpus drawn from source-domain distributions $\mathcal{Q}_{\mathrm{pre}}=\{q_i\}_{i=1}^{M}$. Given a \termreview{reference sequence} $X_{\mathrm{ref}}\sim q_*$ from an unseen target distribution $q_*\notin\mathcal{Q}_{\mathrm{pre}}$, \termreview{zero-shot time series generation} uses fixed pretrained parameters $\Theta$ and independent Gaussian sources to generate new sequences:
\begin{equation}
    Z_j\sim p_0=\mathcal{N}(0,I),
    \qquad
    \widehat{X}^{(j)}=G_{\Theta}\bigl(Z_j;X_{\mathrm{ref}}\bigr),
    \quad j=1,\ldots,N.
    \label{eq:problem-formulation}
\end{equation}
The resulting conditional distribution $p_{\Theta}(\widehat{X}\mid X_{\mathrm{ref}})$ should preserve the statistical properties and temporal organization expressed by $X_{\mathrm{ref}}$, while $Z_j$ provides variation among generated sequences.


\section{Method}
\label{sec:method}

\subsection{Overview}
\label{sec:method-overview}

\begin{figure}[t]
  \centering
  \includegraphics[width=0.8\textwidth,pagebox=cropbox]{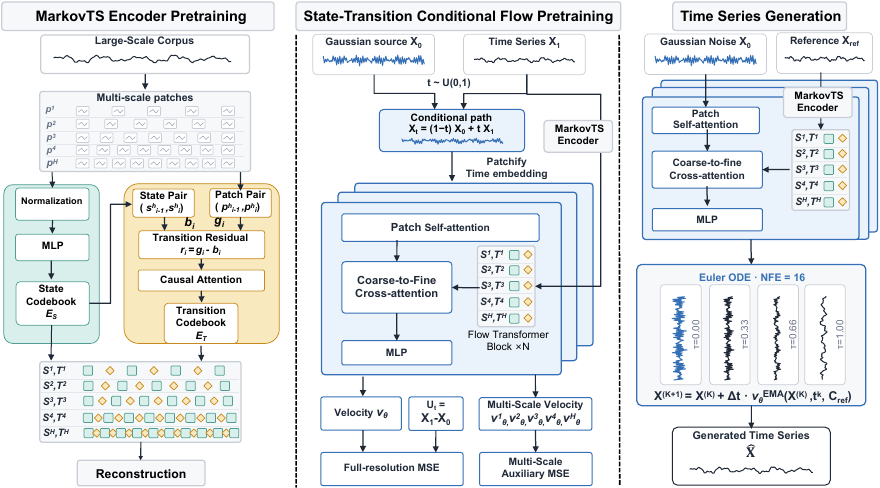}
  \caption{Overview of MarkovTS. \textbf{Left:} multiscale state--transition representation learning. \textbf{Middle:} conditional flow pretraining. \textbf{Right:} single-reference zero-shot generation.}
  \label{fig:markovts-overview}
\end{figure}

\revisionblue{MarkovTS combines a pretrained multiscale state--transition encoder with a conditional flow-matching generator (Figure~\ref{fig:markovts-overview}). At inference, the encoder extracts a fixed condition from a single reference, and the generator produces multiple sequences from independent Gaussian sources.}

\subsection{Multiscale State--Transition Representation Pretraining}
\label{sec:state-transition-representation}

Inspired by hidden Markov models, MarkovTS uses a state codebook to describe local morphology and a transition codebook to quantize transition queries derived from relations between adjacent patches and contextualized by causal attention over the full preceding context. This construction does not assume a first-order transition matrix.

\revisionblue{MarkovTS processes each variable independently with shared parameters. For a length-$L$ time series with $D$ aligned variables, $X\in\mathbb R^{D\times L}$, standardization gives $\bar X$. At scale $h$, with $w_1>\cdots>w_H$, each variable sequence (row) is partitioned into $n_h$ non-overlapping intervals of span $w_h$, each represented by a $P$-dimensional patch $p_i^h$, $i=0,\ldots,n_h-1$. We omit the variable index $d$ below.}

\revisionblue{The state branch isolates local morphology before assigning a discrete code. Let $\mu_{\mathrm v}(p_i^h)$ and $\sigma_{\mathrm v}(p_i^h)$ denote the mean and population standard deviation over nonempty bins only. It uses $\widetilde p_i^h=(p_i^h-\mu_{\mathrm v}(p_i^h))/\max(\sigma_{\mathrm v}(p_i^h),\epsilon)$; empty bins are filled with $\mu_{\mathrm v}(p_i^h)$ and therefore map to zero.} A shared multilayer perceptron maps $\widetilde p_i^h$ to a unit-normalized query $q_i^{S,h}$, which is assigned to the most similar entry in the state codebook $E_S=\{e_k^S\}_{k=1}^{N_S}$:
\begin{equation}
  k_i^{S,h}=\arg\max_k\operatorname{cos}(q_i^{S,h},e_k^S),
  \qquad s_i^h=e_{k_i^{S,h}}^S.
  \label{eq:state-assignment}
\end{equation}
\revisionblue{Patch-independent encoding and a scale-shared codebook provide a common vocabulary for local morphologies at different temporal extents.}

The transition branch complements normalized states with changes between adjacent intervals, indexed by $i=1,\ldots,n_h-1$. A separate encoder maps patches without local standardization to continuous features $a_i^h=f_{\mathrm{full}}(p_i^h)$, retaining relative level and amplitude. The patch-pair and state-pair encoders concatenate adjacent features and their differences to form the continuous patch-pair relation $g_i^h$, state-pair relation $b_i^h$, and transition residual $r_i^h$:
\begin{equation}
\begin{aligned}
  g_i^h&=f_P([a_{i-1}^h,a_i^h,a_i^h-a_{i-1}^h]),\\
  b_i^h&=f_S([s_{i-1}^h,s_i^h,s_i^h-s_{i-1}^h]),\qquad
  r_i^h=g_i^h-b_i^h.
\end{aligned}
  \label{eq:transition-residual}
\end{equation}

\revisionblue{Projected residual queries are interleaved with state vectors at interval boundaries and centers, respectively. Causal attention uses preceding states and residual queries to produce normalized transition queries $q_i^{T,h}$. Quantization selects from the separate, scale-shared transition codebook $E_T=\{e_k^T\}_{k=1}^{N_T}$:}

\begin{equation}
  k_i^{T,h}=\arg\max_k\operatorname{cos}(q_i^{T,h},e_k^T),
  \qquad \tau_i^h=e_{k_i^{T,h}}^T.
  \label{eq:transition-assignment}
\end{equation}
Thus, identical local state pairs can receive different transition codes when their continuous relations or preceding contexts differ.

\revisionblue{Pretraining first learns morphological states through normalized-patch reconstruction and code commitment~\citep{oord2017vqvae}. The state tokenizer and codebook are then frozen. State--transition learning reconstructs full-resolution and multiscale waveforms, while residual matching trains transition codes to retain relational direction and magnitude. Auxiliary regularization stabilizes the queries and discourages collapse.}

For variable $d$, let $S_d^h=(s_{d,0}^h,\ldots,s_{d,n_h-1}^h)$ and $T_d^h=(\tau_{d,1}^h,\ldots,\tau_{d,n_h-1}^h)$ denote ordered sequences of assigned state and transition vectors. The complete representation supplied to the generator is
\begin{equation}
  C(\bar X)=\left(C_d(\bar X)\right)_{d=1}^{D},\qquad
  C_d(\bar X)=\left((S_d^h,T_d^h)\right)_{h=1}^{H}.
  \label{eq:markov-condition}
\end{equation}

\subsection{State--Transition Conditional Flow Pretraining}
\label{sec:state-transition-flow}

\revisionblue{The conditional flow model converts discrete state--transition representations into continuous waveforms by injecting temporally aligned conditions from coarse to fine. The encoder and codebooks remain fixed, while flow matching trains the generator from Gaussian sources.}

\revisionblue{Let $X_1$ denote a standardized training input and $X_{1,d}\in\mathbb R^L$ its variable-$d$ sequence, paired with $C_d(X_1)$. An independent Gaussian source $X_{0,d}$ and time $t$ define $X_{t,d}$ through Equation~\ref{eq:fm-path}. Each forward pass uses one variable and its condition. Separate projections map the interleaved state and transition vectors into a common conditioning space while preserving their temporal coordinates. Each fine-scale element incorporates the average of overlapping projected elements from the nearest available coarser scale, and within-scale self-attention produces the scale-$h$ condition memory $C_d^h$.}

\revisionblue{The flow Transformer embeds patches of $X_{t,d}$ with flow-time information. Each block applies waveform self-attention, then injects conditions through coarse-to-fine cross-attention. At each scale, waveform features are pooled over aligned intervals and attend to condition memory within overlapping temporal spans. The update is interpolated back and added before the next finer scale, allowing fine conditions to refine the coarse-conditioned waveform features. For waveform features $H_\ell$ in block $\ell$, the scale-$h$ update is}
\begin{equation}
 H_\ell\leftarrow H_\ell+\operatorname{Up}_h\!\left[
 \operatorname{CA}\!\left(\operatorname{Pool}_h(H_\ell),C_d^h\right)\right].
\label{eq:main-cross-scale-conditioning}
\end{equation}
\revisionblue{The operators $\operatorname{Pool}_h$ and $\operatorname{Up}_h$ denote aligned pooling and interpolation, and $\operatorname{CA}$ denotes cross-attention. Pooled queries are recomputed after each scale update, so finer-scale conditioning builds on the preceding coarser-scale update. A feed-forward layer completes the block.}

\revisionblue{The output predicts the full-resolution velocity $v_\theta(X_{t,d},t,C_d(X_1))\in\mathbb R^L$ against $X_{1,d}-X_{0,d}$ under conditional flow matching~\citep{lipman2023flowmatching}. Auxiliary heads predict pooled versions of the same target, linking multiscale conditions to the full-resolution field used for sampling.}

\subsection{\revisionblue{Reference-Conditioned Zero-Shot Generation}}
\label{sec:reference-zero-shot-generation}

\revisionblue{At inference, the pretrained generator produces new sequences from a fixed reference-derived condition and fresh Gaussian sources. Each aligned variable uses its own condition; sources are shared across variables within a sample and independent across samples.}

\revisionblue{Reference statistics normalize $X_{\mathrm{ref}}$ to $\bar X_{\mathrm{ref}}$, which is encoded once into $C(\bar X_{\mathrm{ref}})$. For sample $j$, $\epsilon^{(j)}\sim\mathcal N(0,I_L)$ initializes each variable. Euler integration on $0=t_0<\cdots<t_K=1$ uses the exponential-moving-average velocity field $v_{\bar\theta}$:}

\begin{equation}
\begin{aligned}
 Z_{d,0}^{(j)}&=\epsilon^{(j)},\\
 Z_{d,k+1}^{(j)}&=Z_{d,k}^{(j)}
 +(t_{k+1}-t_k)\,
 v_{\bar\theta}\!\left(Z_{d,k}^{(j)},t_k,C_d(\bar X_{\mathrm{ref}})\right).
\end{aligned}
 \label{eq:shared-source-generation}
\end{equation}
\revisionblue{The reference mean $\mu_d$ and effective scale $s_d$ restore the original units as $\widehat X_d^{(j)}=\mu_d+s_d Z_{d,K}^{(j)}$. Implementation details, training objectives, and theories are provided in Appendices~\ref{app:method-details}--\ref{app:theory}.}

\section{Experiments}
\label{sec:experiments}

\revisionblue{We organize the evaluation around three research questions:}

\noindent\revisionblue{\textbf{RQ1: Generation quality.} How well does MarkovTS match unseen target distributions while preserving reference-specific temporal organization and stochastic diversity?}

\noindent\revisionblue{\textbf{RQ2: Representation analysis.} What do state and transition codes represent, and how do representation design and code assignment affect generation quality?}

\noindent\revisionblue{\textbf{RQ3: Downstream utility.} Can MarkovTS-generated data improve downstream tasks?}

\subsection{Experimental Setup}
\label{sec:experimental-setup}

\paragraph{\revisionblue{Data and generation settings.}}
\revisionblue{We evaluate generation on 12 datasets covering energy, transportation, environmental, and economic data. The MarkovTS encoder and generator are pretrained on GIFT-Eval Pretrain~\citep{aksu2024gifteval},\footnote{\url{https://huggingface.co/datasets/Salesforce/GiftEvalPretrain}} excluding the 12 evaluation datasets. We follow the evaluation settings of TimeDP and Winformer, using the target training split only to provide references, without fine-tuning. In the main comparison, deterministic nearest-code state and transition assignments from each length-512 reference condition length-512 outputs generated with independent Gaussian sources and 16-step Euler integration. Each output is partitioned into three non-overlapping length-168 evaluation windows, with the final eight points discarded. Results report means and sample standard deviations over seeds 2021--2025.}

\paragraph{Baselines.}
We compare with nine generators: TimeGAN and GT-GAN \citep{yoon2019timegan,jeon2022gtgan}; TimeVAE, TimeVQVAE, and TimeVQVAE-C \citep{desai2021timevae,lee2023timevqvae,huang2025timedp}; and Diffusion-TS, TimeDP, Winformer, and ImagenFew \citep{yuan2024diffusionts,huang2025timedp,zhou2026winformer,gonen2025imagenfew}. \revisionblue{All main-table baselines are trained or adapted with target-domain training data under their reported protocols; notably, TimeDP and its reproduced baselines are trained jointly on all 12 evaluation datasets, with unseen-domain generation evaluated only in a separate experiment \citep{huang2025timedp}.}

\paragraph{Evaluation metrics.}
\revisionblue{We use Maximum Mean Discrepancy (MMD) \citep{gretton2012kernel} and a histogram-based flattened Kullback--Leibler divergence (flat-KL) \citep{kullback1951information} for marginal target-distribution agreement; lower values are better. Fixed-noise reference swaps assess conditional fidelity using shape-normalized TS2Vec, autocorrelation, spectral distance, and reference retrieval.}

\revisionblue{Appendices~\ref{app:training-configuration} and~\ref{app:generation-datasets}--\ref{app:evaluation-metrics} provide training configurations and evaluation details covering datasets, protocols, metrics, and baselines.}

\begin{table*}[!t]
\caption{Time-series generation quality on non-overlapping length-168 windows from length-512 outputs conditioned on length-512 references. Lower is better; values are five-seed mean $\pm$ sample standard deviation. Best and second-best methods are \textbf{bold} and \underline{underlined}.}
\label{tab:generation-l168}
\centering
\setlength{\tabcolsep}{2.2pt}
\renewcommand{\arraystretch}{0.927}
\resizebox{\textwidth}{!}{%
\begin{tabular}{cl*{10}{c}}
\toprule
  Metric & Dataset & MarkovTS
& Winformer & ImagenFew & TimeDP & Diffusion-TS & TimeGAN & GT-GAN & TimeVAE & TimeVQVAE & TimeVQVAE-C \\
\midrule
\multirow{12}{*}{\rotatebox[origin=c]{90}{Maximum Mean Discrepancy}}
& Electricity & \textbf{0.001}$\pm$0.001 & \textbf{0.001}$\pm$0.002 & \underline{0.002}$\pm$0.003 & \textbf{0.001}$\pm$0.001 & 0.003$\pm$0.002 & 0.367$\pm$0.255 & 0.254$\pm$0.166 & 0.577$\pm$0.006 & 0.152$\pm$0.024 & \underline{0.002}$\pm$0.001 \\
& Solar       & \textbf{0.033}$\pm$0.003 & \underline{0.035}$\pm$0.004 & 0.039$\pm$0.011 & 0.041$\pm$0.011 & 0.050$\pm$0.012 & 0.628$\pm$0.053 & 0.578$\pm$0.039 & 0.353$\pm$0.014 & 0.437$\pm$0.020 & 0.058$\pm$0.005 \\
& Wind        & \textbf{0.013}$\pm$0.009 & 0.034$\pm$0.014 & 0.071$\pm$0.050 & 0.025$\pm$0.017 & 0.035$\pm$0.006 & 0.213$\pm$0.017 & 0.170$\pm$0.040 & 0.170$\pm$0.004 & 0.131$\pm$0.014 & \underline{0.018}$\pm$0.007 \\
& Traffic     & 0.087$\pm$0.002 & \textbf{0.071}$\pm$0.005 & \underline{0.080}$\pm$0.009 & 0.083$\pm$0.034 & 0.111$\pm$0.031 & 0.567$\pm$0.057 & 0.538$\pm$0.078 & 0.218$\pm$0.007 & 0.213$\pm$0.016 & 0.089$\pm$0.001 \\
& Taxi        & 0.110$\pm$0.008 & \textbf{0.085}$\pm$0.010 & 0.108$\pm$0.042 & \underline{0.095}$\pm$0.023 & 0.131$\pm$0.014 & 0.275$\pm$0.054 & 0.319$\pm$0.032 & 0.139$\pm$0.007 & 0.128$\pm$0.004 & 0.109$\pm$0.014 \\
& Pedestrian  & \textbf{0.032}$\pm$0.006 & \underline{0.040}$\pm$0.008 & 0.041$\pm$0.012 & 0.044$\pm$0.020 & 0.071$\pm$0.019 & 0.090$\pm$0.030 & 0.112$\pm$0.019 & 0.065$\pm$0.002 & 0.067$\pm$0.007 & 0.058$\pm$0.002 \\
& Air         & \underline{0.015}$\pm$0.002 & \textbf{0.011}$\pm$0.002 & 0.023$\pm$0.013 & \textbf{0.011}$\pm$0.003 & 0.022$\pm$0.011 & 0.120$\pm$0.045 & 0.211$\pm$0.041 & 0.089$\pm$0.016 & 0.028$\pm$0.002 & 0.041$\pm$0.008 \\
& Temperature & \textbf{0.211}$\pm$0.002 & 0.230$\pm$0.021 & 0.275$\pm$0.059 & \underline{0.219}$\pm$0.022 & 0.241$\pm$0.049 & 0.926$\pm$0.042 & 0.809$\pm$0.081 & 1.002$\pm$0.014 & 0.323$\pm$0.008 & 0.259$\pm$0.043 \\
& Rain        & \textbf{0.031}$\pm$0.002 & \underline{0.036}$\pm$0.016 & 0.052$\pm$0.003 & 0.057$\pm$0.039 & 0.079$\pm$0.058 & 0.329$\pm$0.285 & 0.111$\pm$0.109 & 0.292$\pm$0.019 & 0.074$\pm$0.007 & 0.080$\pm$0.004 \\
& NN5         & \textbf{0.145}$\pm$0.000 & \underline{0.147}$\pm$0.008 & 0.164$\pm$0.005 & 0.164$\pm$0.010 & 0.186$\pm$0.043 & 0.874$\pm$0.088 & 0.632$\pm$0.074 & 0.821$\pm$0.061 & 0.327$\pm$0.012 & 0.243$\pm$0.041 \\
& Fred-MD     & \textbf{0.001}$\pm$0.000 & \underline{0.002}$\pm$0.001 & 0.005$\pm$0.002 & \underline{0.002}$\pm$0.001 & 0.006$\pm$0.002 & 0.043$\pm$0.021 & 0.133$\pm$0.102 & 0.059$\pm$0.008 & 0.008$\pm$0.002 & 0.005$\pm$0.002 \\
& Exchange & \textbf{0.124}$\pm$0.008 & \underline{0.137}$\pm$0.012 & 0.160$\pm$0.027 & 0.151$\pm$0.024 & 0.206$\pm$0.035 & 0.530$\pm$0.154 & 0.475$\pm$0.116 & 0.543$\pm$0.149 & 0.342$\pm$0.050 & 0.233$\pm$0.107 \\
\midrule
\multirow{12}{*}{\rotatebox[origin=c]{90}{K-L Divergence}}
& Electricity & \textbf{0.001}$\pm$0.000 & \underline{0.008}$\pm$0.010 & 0.154$\pm$0.094 & 0.012$\pm$0.016 & 0.315$\pm$0.247 & 0.488$\pm$0.175 & 0.407$\pm$0.079 & 0.734$\pm$0.023 & 0.280$\pm$0.051 & 0.027$\pm$0.015 \\
& Solar       & \textbf{0.009}$\pm$0.002 & \underline{0.013}$\pm$0.005 & 0.079$\pm$0.010 & 0.016$\pm$0.005 & 0.066$\pm$0.055 & 0.612$\pm$0.447 & 0.120$\pm$0.041 & 0.260$\pm$0.016 & 0.865$\pm$0.108 & 0.234$\pm$0.062 \\
& Wind        & \underline{0.111}$\pm$0.018 & 0.202$\pm$0.044 & 0.374$\pm$0.295 & 0.152$\pm$0.034 & 0.548$\pm$0.372 & 1.924$\pm$1.233 & \textbf{0.107}$\pm$0.016 & 0.484$\pm$0.015 & 0.483$\pm$0.066 & 0.183$\pm$0.047 \\
& Traffic     & \textbf{0.007}$\pm$0.002 & 0.011$\pm$0.002 & 0.010$\pm$0.004 & \underline{0.009}$\pm$0.003 & 0.120$\pm$0.074 & 1.305$\pm$0.320 & 1.409$\pm$0.251 & 0.211$\pm$0.014 & 0.178$\pm$0.026 & 0.016$\pm$0.003 \\
& Taxi        & \underline{0.010}$\pm$0.004 & \textbf{0.005}$\pm$0.003 & 0.020$\pm$0.011 & 0.011$\pm$0.004 & 0.075$\pm$0.034 & 0.650$\pm$0.180 & 0.950$\pm$0.197 & 0.110$\pm$0.020 & 0.110$\pm$0.026 & 0.038$\pm$0.010 \\
& Pedestrian  & \textbf{0.007}$\pm$0.001 & \underline{0.009}$\pm$0.004 & 0.020$\pm$0.008 & 0.014$\pm$0.010 & 0.133$\pm$0.069 & 0.417$\pm$0.181 & 0.411$\pm$0.096 & 0.065$\pm$0.005 & 0.405$\pm$0.051 & 0.039$\pm$0.008 \\
& Air         & \textbf{0.026}$\pm$0.011 & \textbf{0.026}$\pm$0.011 & 0.054$\pm$0.026 & \underline{0.027}$\pm$0.016 & 0.106$\pm$0.079 & 0.348$\pm$0.093 & 0.578$\pm$0.049 & 0.164$\pm$0.012 & 0.054$\pm$0.012 & 0.093$\pm$0.025 \\
& Temperature & \textbf{0.137}$\pm$0.041 & 0.176$\pm$0.027 & 0.235$\pm$0.070 & \underline{0.171}$\pm$0.073 & 0.342$\pm$0.131 & 8.892$\pm$2.681 & 3.174$\pm$2.685 & 2.183$\pm$0.110 & 0.735$\pm$0.066 & 0.379$\pm$0.110 \\
& Rain        & \textbf{0.002}$\pm$0.001 & \underline{0.011}$\pm$0.003 & 0.012$\pm$0.002 & 0.013$\pm$0.012 & 0.061$\pm$0.072 & 0.506$\pm$0.174 & 0.432$\pm$0.099 & 0.160$\pm$0.022 & 0.047$\pm$0.018 & 0.065$\pm$0.018 \\
& NN5         & \textbf{0.045}$\pm$0.003 & \textbf{0.045}$\pm$0.007 & 0.057$\pm$0.017 & \underline{0.054}$\pm$0.014 & 0.165$\pm$0.076 & 4.928$\pm$4.112 & 1.386$\pm$0.520 & 1.337$\pm$0.220 & 1.063$\pm$0.274 & 0.220$\pm$0.151 \\
& Fred-MD     & \textbf{0.176}$\pm$0.001 & \underline{0.201}$\pm$0.014 & 0.528$\pm$0.445 & 0.203$\pm$0.035 & 0.835$\pm$0.554 & 0.512$\pm$0.290 & 0.380$\pm$0.070 & 0.346$\pm$0.041 & 0.831$\pm$0.077 & 1.118$\pm$0.276 \\
& Exchange & \underline{1.794}$\pm$0.137 & \textbf{1.621}$\pm$0.126 & 3.498$\pm$3.061 & 1.866$\pm$0.132 & 2.337$\pm$0.714 & 8.861$\pm$3.397 & 7.201$\pm$4.380 & 10.404$\pm$1.434 & 5.052$\pm$1.385 & 8.475$\pm$3.056 \\
\midrule
\multicolumn{2}{l}{Wins / win rate} & 18 / 75.0\% & 8 / 33.3\% & 0 / 0.0\% & 2 / 8.3\% & 0 / 0.0\% & 0 / 0.0\% & 1 / 4.2\% & 0 / 0.0\% & 0 / 0.0\% & 0 / 0.0\% \\
\bottomrule
\end{tabular}%
}
\end{table*}

\subsection{\revisionblue{Generation Quality (RQ1)}}
\label{sec:generation-quality}

\paragraph{\revisionblue{Distributional fidelity.}}
\revisionblue{MarkovTS is best or tied-best in 18 of 24 MMD and flat-KL comparisons in Table~\ref{tab:generation-l168}, leading on 9 of 12 datasets per metric. Representative results are MMD 0.013 versus 0.018 on Wind and flat-KL 0.002 versus 0.011 on Rain.}

\revisionblue{MMD and flat-KL support agreement with target length-168 window and pooled-value distributions. The same pretrained model serves 12 datasets without target updates; Appendix~\ref{app:additional-generation-quality} gives complementary controls.}

\paragraph{\revisionblue{Reference-specific conditional fidelity.}}
\revisionblue{MarkovTS aims to generate diverse samples faithful to a selected reference, which marginal MMD and flat-KL cannot test. Table~\ref{tab:reference-swap-main} fixes Gaussian noise and reference $i$'s affine restoration, then replaces only the joint state--transition condition with that of a deranged reference $j\ne i$; a condition-agnostic generator would be unaffected. On every dataset, swaps increase distance from $i$ in shape-normalized TS2Vec, ACF, and PSD. Active-condition retrieval reaches 93.7\% in the binary task and about 63\% in the 100-way task, versus chance levels of 50\% and 1\%; nearly equal correct/swapped 100-way accuracy indicates redirection rather than information loss. Thus, the intervention directly supports the reference-specific use of the joint condition required by our single-reference claim, without isolating states from transitions.}

\begin{table*}[!ht]
\caption{Reference-specific fidelity under fixed-noise condition swaps. Entries are five-seed means; Appendix~\ref{app:reference-swap} reports standard deviations.}
\label{tab:reference-swap-main}
\centering
\fontsize{7.5}{9}\selectfont
\setlength{\tabcolsep}{2.0pt}
\renewcommand{\arraystretch}{0.765}
\setlength{\aboverulesep}{0.3ex}
\setlength{\belowrulesep}{0.45ex}
\begin{tabular*}{\textwidth}{@{\extracolsep{\fill}}lrrrrrrrrrrrrr}
\toprule
Metric & Elec. & Solar & Wind & Traffic & Taxi & Ped. & Air & Temp. & Rain & NN5 & Fred-MD & Exchange & Mean \\
\midrule
$\Delta$ Shape & 0.016 & 0.017 & 0.093 & 0.056 & 0.041 & 0.127 & 0.074 & 0.036 & 0.040 & 0.018 & 0.032 & 0.057 & 0.051 \\
$\Delta$ ACF & 0.230 & 0.085 & 0.234 & 0.124 & 0.152 & 0.177 & 0.237 & 0.078 & 0.056 & 0.093 & 0.233 & 0.242 & 0.162 \\
$\Delta$ PSD & 0.250 & 0.122 & 0.242 & 0.160 & 0.214 & 0.218 & 0.300 & 0.148 & 0.220 & 0.113 & 0.171 & 0.236 & 0.199 \\
Binary retrieval & 95.5 & 95.7 & 93.8 & 98.8 & 99.4 & 92.0 & 95.8 & 92.3 & 96.7 & 75.9 & 92.2 & 96.0 & 93.7 \\
100-way correct & 58.9 & 51.6 & 71.2 & 84.1 & 89.1 & 45.8 & 86.6 & 45.1 & 78.3 & 26.0 & 46.3 & 77.3 & 63.4 \\
100-way swapped & 59.2 & 51.9 & 69.9 & 84.2 & 89.4 & 46.7 & 84.0 & 45.0 & 78.0 & 25.6 & 46.5 & 76.6 & 63.1 \\
\bottomrule
\end{tabular*}
\end{table*}

\paragraph{\revisionblue{Reference fidelity and stochastic diversity.}}
\label{sec:reference-fidelity-diversity}

\begin{figure}[t]
  \centering
  \includegraphics[width=1\textwidth]{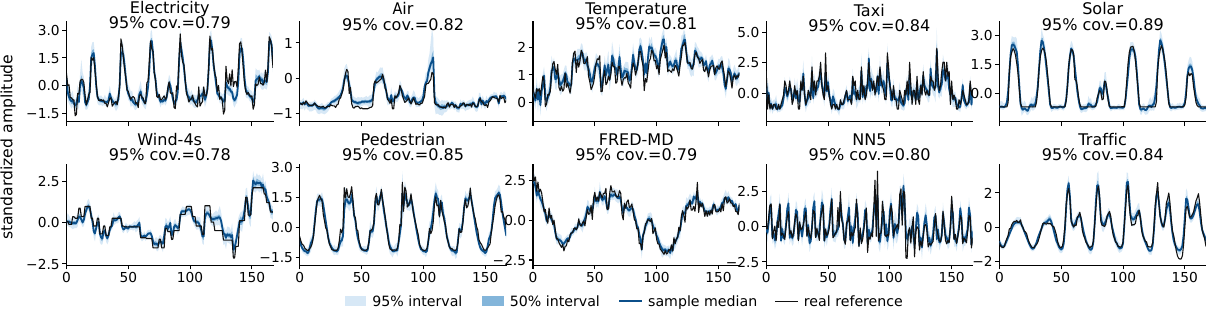}
  \caption{Conditional distributions across ten held-out datasets. Panels show references, medians, and 50\%/95\% empirical intervals; inclusion rate denotes reference coverage by the 95\% interval.}
  \label{fig:conditional-stability-l168}
\end{figure}

\revisionblue{Figure~\ref{fig:conditional-stability-l168} summarizes 50 generated samples per reference across ten held-out datasets. Recurring median traces and nonzero 50\%/95\% intervals show reference alignment with stochastic variation; Appendix~\ref{app:generation-visualizations} gives details and samples.}

\subsection{State--Transition Analysis (RQ2)}
\label{sec:generation-analysis}

\textbf{Code and model replacement.} Table~\ref{tab:code-randomization} compares deterministic assignment with uniform code replacement (Random) and reference-dependent probability sampling (Stochastic), applied to states, transitions, or both with model parameters frozen. This paired intervention uses a deterministic control distinct from the three-window main estimate in Table~\ref{tab:generation-l168}. Randomizing both streams causes greater degradation than either single-stream intervention in most comparisons, indicating that both provide useful reference information. Stochastic sampling instead maintains scores close to the deterministic control, suggesting that reference-dependent code selection preserves useful conditioning information despite assignment changes.

Patch-Flow replaces the state--transition representation with a standard patch Transformer and single codebook while retaining the remaining model and training settings. Under the paired intervention protocol, MarkovTS has lower MMD and flat-KL on eleven of twelve datasets, supporting state--transition over single-codebook patch conditioning.

\begin{table*}[t]
\caption{Generation quality under state--transition code interventions and patch-based comparison.}
\label{tab:code-randomization}
\centering
\fontsize{7.5}{9}\selectfont
\setlength{\tabcolsep}{2.0pt}
\renewcommand{\arraystretch}{0.765}
\setlength{\aboverulesep}{0.3ex}
\setlength{\belowrulesep}{0.45ex}
\begin{tabular*}{\textwidth}{@{\extracolsep{\fill}}llrrrrrrrrrrrr}
\toprule
Variant & Metric & Elec. & Solar & Wind & Traffic & Taxi & Ped. & Air & Temp. & Rain & NN5 & Fred-MD & Exchange \\
\midrule
MarkovTS & MMD & 0.001 & 0.032 & 0.019 & 0.087 & 0.101 & 0.032 & 0.009 & 0.142 & 0.031 & 0.180 & 0.002 & 0.124 \\
 & flat-KL & 0.001 & 0.009 & 0.147 & 0.003 & 0.002 & 0.007 & 0.019 & 0.161 & 0.007 & 0.043 & 0.244 & 1.631 \\
\midrule
Random states & MMD & 0.002 & 0.236 & 0.023 & 0.168 & 0.118 & 0.049 & 0.016 & 0.134 & 0.046 & 0.205 & 0.003 & 0.148 \\
 & flat-KL & 0.059 & 0.689 & 0.189 & 0.144 & 0.075 & 0.088 & 0.039 & 0.134 & 0.010 & 0.106 & 0.364 & 5.030 \\
Random transitions & MMD & 0.002 & 0.190 & 0.023 & 0.130 & 0.119 & 0.054 & 0.017 & 0.152 & 0.053 & 0.231 & 0.002 & 0.147 \\
 & flat-KL & 0.061 & 0.842 & 0.165 & 0.131 & 0.146 & 0.148 & 0.046 & 0.109 & 0.015 & 0.193 & 0.200 & 3.068 \\
Random both & MMD & 0.003 & 0.516 & 0.023 & 0.302 & 0.148 & 0.082 & 0.022 & 0.213 & 0.051 & 0.292 & 0.002 & 0.148 \\
 & flat-KL & 0.068 & 1.099 & 0.178 & 0.247 & 0.184 & 0.192 & 0.057 & 0.253 & 0.050 & 0.301 & 0.236 & 3.351 \\
\midrule
Stochastic states & MMD & 0.001 & 0.034 & 0.023 & 0.080 & 0.091 & 0.034 & 0.013 & 0.134 & 0.036 & 0.166 & 0.003 & 0.138 \\
  & flat-KL & 0.005 & 0.015 & 0.177 & 0.007 & 0.016 & 0.010 & 0.030 & 0.098 & 0.004 & 0.047 & 0.256 & 1.760 \\
Stochastic transitions & MMD & 0.002 & 0.034 & 0.023 & 0.080 & 0.091 & 0.035 & 0.013 & 0.134 & 0.036 & 0.186 & 0.003 & 0.142 \\
  & flat-KL & 0.012 & 0.014 & 0.177 & 0.007 & 0.017 & 0.015 & 0.030 & 0.098 & 0.004 & 0.051 & 0.291 & 1.842 \\
Stochastic both & MMD & 0.002 & 0.034 & 0.023 & 0.080 & 0.091 & 0.037 & 0.013 & 0.134 & 0.036 & 0.186 & 0.003 & 0.143 \\
  & flat-KL & 0.015 & 0.020 & 0.177 & 0.007 & 0.016 & 0.020 & 0.030 & 0.098 & 0.004 & 0.051 & 0.288 & 1.883 \\
\midrule
Patch-Flow  & MMD & 0.003 & 0.045 & 0.033 & 0.084 & 0.105 & 0.057 & 0.011 & 0.204 & 0.040 & 0.236 & 0.004 & 0.142 \\
 & flat-KL & 0.054 & 0.173 & 0.203 & 0.003 & 0.012 & 0.163 & 0.021 & 0.224 & 0.008 & 0.215 & 0.369 & 2.013 \\
\bottomrule
\end{tabular*}
\end{table*}

\begin{figure}[t]
\centering
\includegraphics[width=1\textwidth]{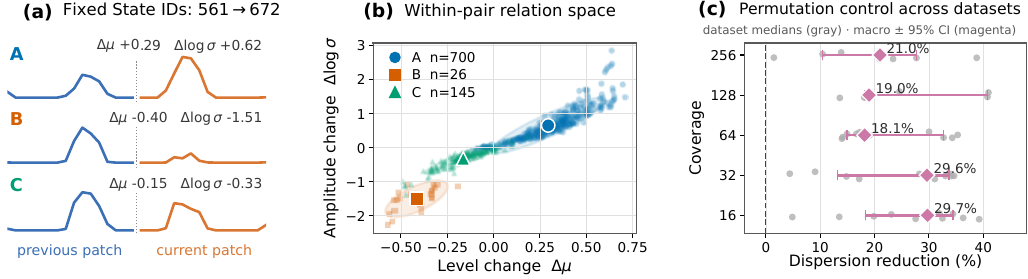}
\caption{State--transition complementarity on Solar at span 32. (a) Real patch pairs sharing state IDs but assigned distinct transition codes. (b) Occurrence centroids and covariance ellipses. (c) Scale-level within-code dispersion reduction against count-preserving permutations across datasets; gray points are dataset medians, and magenta diamonds are macro-medians with 95\% bootstrap CIs.}
\label{fig:transition-complementarity}
\end{figure}

\textbf{State--transition complementarity.} Using real training references and the frozen encoder, we test whether transition codes distinguish changes within the same ordered state pair. Figure~\ref{fig:transition-complementarity}(a--b) illustrates distinct level and amplitude changes associated with different transition codes. Across 3,589 groups defined by dataset, scale, and state pair, we compare within-code dispersion in level, amplitude, slope, and boundary changes against count-preserving label permutations. The scale-level macro-median reduction is 18.1--29.7\%, with all five 95\% bootstrap confidence intervals excluding zero (Figure~\ref{fig:transition-complementarity}(c)). This indicates that context-conditioned transition codes capture variation not resolved by the state pair, complementing the morphological states.

\subsection{\revisionblue{Downstream Utility (RQ3)}}
\label{sec:downstream-applications}
\label{sec:downstream-forecasting}

\revisionblue{We use MarkovTS-generated data for downstream training. \emph{Real + MarkovTS} pairs each real training example with one generated counterpart and replaces exactly half of every mini-batch: complete sequences for classification and position-matched windows for forecasting and anomaly detection. Batch size and updates per epoch match \emph{Real only}; Arsenal uses a fixed 50\% replacement set of equal cardinality. All comparisons share maximum training budgets, model selection, real validation indices, and real test data.}

\revisionblue{\textbf{Forecasting.} Following the BasicTS protocol~\citep{shao2024exploring,li2026apt}\footnote{\url{https://github.com/GestaltCogTeam/BasicTS}}, we evaluate five backbones on six datasets with lookback 336 and prediction horizons $\{96,192,336,720\}$. Table~\ref{tab:forecast-aug} reports mean absolute error (MAE) and mean squared error (MSE), averaged over the four horizons.}

\revisionblue{Augmentation reduces both errors for every backbone on multiple datasets; ETTh1 MSE drops from 0.485 to 0.421 for iTransformer and Exchange MSE from 0.478 to 0.390 for SparseTSF.}

For forecasting only, \emph{Online} regenerates one counterpart per real training window each epoch, trains only on generated windows, and matches the per-epoch batch and update counts of \emph{Real only} (Appendix~\ref{app:forecast-aug}).

\begin{table*}[!t]
\caption{Forecasting performance averaged over four horizons.
Cells show \emph{Real only}/\emph{Real + MarkovTS}; improvements are bold.
Imp. counts improved dataset--metric pairs out of 12.}
\label{tab:forecast-aug}
\centering
\fontsize{7}{9}\selectfont
\setlength{\tabcolsep}{0pt}
\renewcommand{\arraystretch}{0.80}
\setlength{\aboverulesep}{0.3ex}
\setlength{\belowrulesep}{0.45ex}
\begin{tabular*}{\textwidth}{@{\extracolsep{\fill}}l@{\hspace{3pt}}*{6}{c@{\hspace{1pt}}c@{\hspace{1.5pt}}}c@{}}
\toprule
& \multicolumn{2}{c}{ECL} & \multicolumn{2}{c}{ETTh1}
& \multicolumn{2}{c}{ETTh2} & \multicolumn{2}{c}{Exchange}
& \multicolumn{2}{c}{Traffic} & \multicolumn{2}{c}{Weather} & \\
\cmidrule(lr){2-3}\cmidrule(lr){4-5}\cmidrule(lr){6-7}
\cmidrule(lr){8-9}\cmidrule(lr){10-11}\cmidrule(lr){12-13}
Backbone & MAE$\downarrow$ & MSE$\downarrow$ & MAE$\downarrow$ & MSE$\downarrow$ & MAE$\downarrow$ & MSE$\downarrow$
& MAE$\downarrow$ & MSE$\downarrow$ & MAE$\downarrow$ & MSE$\downarrow$ & MAE$\downarrow$ & MSE$\downarrow$ & Imp. \\
\midrule
CATS
& .258/\textbf{.245} & .165/\textbf{.161}
& .443/\textbf{.423} & .448/\textbf{.425}
& .425/\textbf{.384} & .392/\textbf{.355}
& .420/.421 & .388/\textbf{.387}
& .284/\textbf{.252} & .421/\textbf{.402}
& .258/\textbf{.249} & .221/.225  & 10/12 \\
iTrans.
& .259/.261 & .164/.169
& .473/\textbf{.433} & .485/\textbf{.421}
& .439/\textbf{.427} & .421/\textbf{.409}
& .478/\textbf{.447} & .464/\textbf{.419}
& .289/.297 & .412/.429
& .267/\textbf{.258} & .233/\textbf{.230}  & 8/12 \\
SparseTSF
& .265/\textbf{.261} & .173/\textbf{.171}
& .429/\textbf{.417} & .427/\textbf{.412}
& .427/\textbf{.414} & .399/\textbf{.389}
& .481/\textbf{.425} & .478/\textbf{.390}
& .295/\textbf{.279} & .445/.458
& .283/\textbf{.276} & .252/.255  & 10/12 \\
TQNet
& .246/.247 & .158/.158
& .425/.428 & .422/.429
& .417/\textbf{.416} & .391/.393
& .454/\textbf{.427} & .428/\textbf{.387}
& .265/\textbf{.259} & .410/\textbf{.398}
& .249/.252 & .222/.223  & 5/12 \\
OLinear
& .298/\textbf{.297} & .188/\textbf{.186}
& .428/.429 & .414/.415
& .416/\textbf{.412} & .389/\textbf{.386}
& .456/\textbf{.439} & .435/\textbf{.402}
& .328/\textbf{.321} & .465/\textbf{.453}
& .250/.254 & .221/.223  & 8/12 \\
\bottomrule
\end{tabular*}
\end{table*}

\label{sec:downstream-classification}

\begin{table*}[!t]
\caption{Classification on 23 UEA datasets: \emph{Real only}/\emph{Real + MarkovTS}.
 Bold: improvements; Imp.: improved metrics. Percentages except LL.}
\label{tab:classification-augmentation}
\centering
\fontsize{7.5}{9}\selectfont
\setlength{\tabcolsep}{3.0pt}
\renewcommand{\arraystretch}{0.80}
\setlength{\aboverulesep}{0.3ex}
\setlength{\belowrulesep}{0.45ex}
\begin{tabular*}{\textwidth}{@{\extracolsep{\fill}}lrrrrrrrr}
\toprule
Classifier & Acc.$\uparrow$ & BAcc.$\uparrow$ & F1$\uparrow$ & AUROC$\uparrow$
& LL$\downarrow$ & Sens.$\uparrow$ & Spec.$\uparrow$ & Imp. \\
\midrule
ResNet1D & 68.15/\textbf{68.94} & 67.94/\textbf{68.86}
& 65.05/\textbf{67.12} & 83.71/\textbf{85.05}
& 1.13/\textbf{1.07} & 66.02/\textbf{69.33} & 69.95/69.21 & 6/7 \\
TimesNet & 65.84/\textbf{68.36} & 65.03/\textbf{66.94}
& 63.86/\textbf{65.26} & 83.32/\textbf{84.13}
& 1.16/\textbf{1.10} & 66.28/\textbf{66.73}
& 65.03/\textbf{68.30} & 7/7 \\
ConvTran & 69.31/\textbf{70.23} & 68.01/\textbf{69.16}
& 67.93/67.49 & 85.52/\textbf{85.65} & 0.82/\textbf{0.79}
& 70.49/69.10 & 67.36/\textbf{70.65} & 5/7 \\
Arsenal & 73.21/73.17 & 71.34/71.32 & 71.16/\textbf{71.31}
& 84.39/\textbf{84.85} & 5.56/\textbf{5.12}
& 71.19/\textbf{71.53} & 74.17/73.77 & 4/7 \\
\bottomrule
\end{tabular*}
\end{table*}

\label{sec:downstream-anomaly}

\begin{table*}[!t]
\caption{Anomaly-detection on 26 datasets. Cells show
\emph{Real only}/\emph{Real + MarkovTS}; improvements are bold.
Higher is better; Imp. counts improved metrics.
A-P/A-R: pointwise PR-/ROC-AUC; R-A-P/R-A-R: range-aware AUCs;
V-PR/V-ROC: PR/ROC volumes under surface.}
\label{tab:anomaly-augmentation}
\centering
\fontsize{7.5}{9}\selectfont
\setlength{\tabcolsep}{3.1pt}
\renewcommand{\arraystretch}{0.80}
\setlength{\aboverulesep}{0.3ex}
\setlength{\belowrulesep}{0.45ex}
\begin{tabular*}{\textwidth}{@{\extracolsep{\fill}}lrrrrrrr}
\toprule
Detector & A-P$\uparrow$ & A-R$\uparrow$ & R-A-P$\uparrow$ & R-A-R$\uparrow$
& V-PR$\uparrow$ & V-ROC$\uparrow$ & Imp. \\
\midrule
DLinear & .233/\textbf{.253} & .672/\textbf{.679}
& .255/\textbf{.273} & .713/\textbf{.715}
& .254/\textbf{.271} & .710/\textbf{.714} & 6/6 \\
TranAD & .278/.278 & .674/\textbf{.676} & .267/.261
& .676/\textbf{.677} & .264/.259 & .673/\textbf{.675} & 3/6 \\
CATCH & .248/\textbf{.254} & .675/\textbf{.683}
& .278/\textbf{.284} & .720/\textbf{.726}
& .275/\textbf{.281} & .717/\textbf{.723} & 6/6 \\
ModernTCN & .237/\textbf{.241} & .660/\textbf{.664}
& .259/\textbf{.262} & .704/\textbf{.708}
& .257/\textbf{.261} & .702/\textbf{.705} & 6/6 \\
TimesNet & .268/.267 & .689/.685 & .286/\textbf{.290}
& .731/.730 & .285/\textbf{.289} & .729/.727 & 2/6 \\
\bottomrule
\end{tabular*}
\end{table*}

\FloatBarrier

\revisionblue{\textbf{Classification.} On 23 UEA datasets~\citep{bagnall2018uea} from Multiverse~\citep{middlehurst2026multiverse}\footnote{\url{https://github.com/aeon-toolkit/multiverse}}, four classifiers are evaluated using seven metrics; generated samples inherit their references' class labels. \emph{Real + MarkovTS} improves most aggregates in Table~\ref{tab:classification-augmentation}, including all seven for TimesNet; improvements average 2.69\%, with no decline exceeding 2\%.}

\revisionblue{\textbf{Anomaly detection.} Following TAB/OpenTS~\citep{qiu2025tab}\footnote{\url{https://decisionintelligence.github.io/OpenTS/leaderboards/}}, five detectors are evaluated on 215 series from 26 datasets; generated counterparts match their references' lengths and index grids and inherit their pointwise labels. \emph{Real + MarkovTS} improves all six aggregates for DLinear, CATCH, and ModernTCN; improvements average 1.89\%, with mixed results for TranAD and TimesNet.}

Appendix~\ref{app:experimental-details} provides protocols; Appendices~\ref{app:forecast-aug}--\ref{app:anomaly-augmentation} report complete results.

\section{Conclusion}
\label{sec:conclusion}

We introduced MarkovTS for reference-conditioned zero-shot time series generation. Its pretrained multiscale state--transition encoder separates locally normalized morphology from contextual transitions to condition a flow-matching generator for stochastic generation from one reference without target-domain tuning. Across 12 unseen datasets, MarkovTS is best or tied-best in most MMD and flat-KL comparisons. Fixed-noise swaps show outputs follow the selected joint condition; code interventions and within-state-pair analyses support complementary state and transition components. MarkovTS-generated data improve forecasting, classification, and anomaly detection on average under matched training budgets, with gains varying across models and metrics. Together, the results support reference-specific generation beyond marginal distribution matching.

\bibliography{iclr2027_conference}
\bibliographystyle{iclr2027_conference}

\clearpage
\appendix
\section{Notation}
\label{app:notation}

\revisionblue{Table~\ref{tab:notation} summarizes the notation. Indices \(h\), \(i\), and \(d\) index scales, intervals, and variables.}

\begin{table}[h!]
\small
\caption{\revisionblue{Notation for the MarkovTS formulation, representation, and generator.}}
\label{tab:notation}
\centering
\small
\setlength{\tabcolsep}{5pt}
\newcommand{\notationrule}{\arrayrulecolor{black!20}\specialrule{0.25pt}{1.2pt}{1.2pt}\arrayrulecolor{black}}
\renewcommand{\arraystretch}{0.75}
\begin{tabular}{lp{0.72\textwidth}}
\toprule
Symbol & Meaning \\
\midrule
\(\mathcal D_{\mathrm{pre}},\mathcal Q_{\mathrm{pre}}\) &
Pretraining corpus and its set of source-domain distributions. \\
\notationrule
\(q_*,R_*,T_*\) &
Unseen target distribution, evaluation reference pool, and disjoint target test
set. Each generated sample uses one member of $R_*$. \\
\notationrule
\(X_{\mathrm{ref}},K_{\mathrm{eval}}\) &
\revisionblue{Inference reference and number of references used in repeated
single-reference evaluation.} \\
\notationrule
\(X\in\mathbb R^{D\times L}\) &
\revisionblue{A length-$L$ time series comprising $D$ aligned variable sequences, each modeled independently.} \\
\notationrule
\revisionblue{$\bar X,P,n_h$} & \revisionblue{Standardized input, canonical patch dimension, and number of padded intervals at scale $h$.} \\
\notationrule
\(\mathcal H,H,h\) &
Set, number, and index of temporal scales. \\
\notationrule
\(w_h\) & \revisionblue{Temporal span at scale $h$, ordered from coarse to fine.} \\
\notationrule
\(p_i^h,\widetilde p_i^h,\mathcal V_i^h\) &
Fixed-dimensional interval patch, its locally standardized form, and its nonempty-bin index set. \\
\notationrule
\(q_i^{S,h},E_S,e_k^S,s_i^h\) &
State query, state codebook, its \(k\)-th entry, and assigned state codebook vector. \\
\notationrule
\revisionblue{$N_S,N_T$} & \revisionblue{State and transition codebook sizes; denoted $K_S,K_T$ in the implementation configuration.} \\
\notationrule
\(g_i^h,b_i^h,r_i^h\) &
Patch-pair relation, state-pair relation, and transition residual. \\
\notationrule
\(q_i^{T,h},E_T,e_k^T,\tau_i^h\) &
Transition query, transition codebook, its \(k\)-th entry, and assigned transition codebook vector. \\
\notationrule
\revisionblue{$a_i^h,R$} & \revisionblue{Continuous patch feature and common pair-relation output dimension.} \\
\notationrule
\revisionblue{$S_d^h,T_d^h$} & \revisionblue{Ordered sequences of state and transition vectors for variable $d$ at scale $h$.} \\
\notationrule
\revisionblue{$C(\bar X),C_d(\bar X)$} &
\revisionblue{Condition collection and single-variable condition.} \\
\notationrule
\revisionblue{$C_{\mathrm{seq},d}^h,C_d^h$} &
\revisionblue{Ordered scale-$h$ condition sequence and processed condition memory.} \\
\notationrule
\(X_0,X_1,X_t,U_t\) &
Gaussian source, clean sample, interpolation state, and target velocity \(X_1-X_0\). \\
\notationrule
\(v_\theta,v_{\bar\theta}\) &
\revisionblue{Velocity fields with learned and exponential-moving-average parameters.} \\
\notationrule
\revisionblue{$M,M^h,P_h$} &
\revisionblue{Full-resolution mask, scale-$h$ mask, and masked pooling operator.} \\
\notationrule
\revisionblue{$j,Z_{d,k}^{(j)},t_k$} & \revisionblue{Sample index, variable-$d$ integration state, and time at step $k$.} \\
\notationrule
\revisionblue{$\mu_d,\sigma_d,s_d$} & \revisionblue{Reference mean, standard deviation, and effective rescaling factor.} \\
\notationrule
\(N_{\mathrm{FE}}\) &
Number of Euler function evaluations for sampling. \\
\bottomrule
\end{tabular}
\end{table}

\section{Additional Method Details}
\label{app:method-details}

\revisionblue{We specify preprocessing, representation objectives, and conditional flow training and sampling.}

\subsection{Preprocessing and aligned patchification}
\label{app:preprocessing}

\revisionblue{For each variable $d$, let $\Omega_d$ index the finite observations in the reference window. The reference mean and population standard deviation are}
\begin{equation}
    \mu_d=\frac{1}{|\Omega_d|}\sum_{t\in\Omega_d}x_{t,d},\qquad
    \sigma_d=\sqrt{\frac{1}{|\Omega_d|}\sum_{t\in\Omega_d}(x_{t,d}-\mu_d)^2},
    \label{eq:app-normalization}
\end{equation}
\revisionblue{For a nondegenerate reference, $s_d=\sigma_d$. Normalization uses $\bar{x}_{t,d}=(x_{t,d}-\mu_d)/(s_d+\epsilon)$ with stabilizer $\epsilon$; output rescaling uses $\hat{x}_{t,d}=s_d\hat{\bar{x}}_{t,d}+\mu_d$. Flow pretraining instead computes statistics over each finite training row before window extraction, without further standardization.}

\revisionblue{For padded length $L_{\mathrm{pad}}$, scale $h$ contains $n_h=L_{\mathrm{pad}}/w_h$ intervals. Each interval $i$ is partitioned into $P$ consecutive bins $B_{i,b}^h$ of $w_h/P$ positions. With validity mask $M_t$, let $c_{i,b}^h=\sum_{t\in B_{i,b}^h}M_t$, $\check p_{i,b}^h=\sum_{t\in B_{i,b}^h}M_t\bar x_t/\max(1,c_{i,b}^h)$, and $\mathcal V_i^h=\{b:c_{i,b}^h>0\}$. The canonical patch is}
\begin{equation}
p_{i,b}^h=
\begin{cases}
\check p_{i,b}^h, & b\in\mathcal V_i^h,\\
\dfrac{\sum_{b'\in\mathcal V_i^h}\check p_{i,b'}^h}
{\max(1,|\mathcal V_i^h|)},
& b\notin\mathcal V_i^h.
\end{cases}
\label{eq:app-canonical-patch}
\end{equation}
\revisionblue{When $\mathcal V_i^h$ is nonempty, the second case fills an empty bin with the mean of the nonempty-bin means in its interval; a fully empty interval evaluates to zero and is masked as inactive. The state branch computes local mean and variance from $\mathcal V_i^h$ only, so filled entries become exactly zero after local standardization and remain in normalized-shape reconstruction. Raw and multiscale waveform reconstruction instead remove them with bin-level masks. Patch-pair features retain source-standardized levels and amplitudes without local standardization.}

\revisionblue{At scales $h\in\mathcal H$, aligned intervals have temporal spans $w_h\in\{256,128,64,32,16\}$ and use a base patch size of $16$. An interval and its state are active when at least half its samples are valid. Transitions require both adjacent states to be valid; invalid intervals break the transition sequence. Aligned variables are separately encoded on shared interval boundaries.}

\subsection{\revisionblue{Representation learning and codebook optimization}}

The patch-pair and state-pair networks $f_P$ and $f_S$ in Eq.~\plaineqref{eq:transition-residual} have independent parameters and a shared $R$-dimensional output space. State-code inputs to $f_S$ are detached, but $f_S$ remains trainable.

\label{app:representation-learning}
\begingroup

State warm-up learns local morphology before transition learning freezes the state tokenizer and codebook. Both codebooks span all scales; decoders serve only representation pretraining.

\subsubsection{State warm-up}
\label{app:state-objectives}

Let $m_{ui}^h$ indicate patch validity for window $u$, interval $i$, and scale $h$, with $N_h=\max(1,\sum_{u,i}m_{ui}^h)$ and query dimension $d_q=64$. State warm-up minimizes
\begin{equation}
\begin{split}
 \mathcal L_S={}&\mathcal L_{\mathrm{shape}}
 +\lambda_S\mathcal L_{\mathrm{commit}}^S
 +\lambda_{\mathrm{iso}}\mathcal L_{\mathrm{iso}}\\
 &+\lambda_{\mathrm{loc}}\mathcal L_{\mathrm{loc}}
 +\lambda_-\rho_S(s)\mathcal L_-.
\end{split}
\label{eq:app-state-objective}
\end{equation}
A shared linear decoder maps states to $\widehat p_{ui}^h\in\mathbb R^P$, with reconstruction and commitment losses
\begin{equation}
\begin{aligned}
 \mathcal L_{\mathrm{shape}}
 &=\frac1H\sum_{h=1}^H
 \frac{\sum_{u,i}m_{ui}^h
 \|\widehat p_{ui}^h-\widetilde p_{ui}^h\|_1}{P N_h},\\
 \mathcal L_{\mathrm{commit}}^S
 &=\frac1H\sum_{h=1}^H
 \frac{\sum_{u,i}m_{ui}^h
 \|q_{ui}^{S,h}-\operatorname{sg}(s_{ui}^h)\|_2^2}{d_q N_h}.
\end{aligned}
\label{eq:app-state-reconstruction-commitment}
\end{equation}
The operator $\operatorname{sg}$ stops gradients. The denominator $P N_h$ reflects that normalized-shape reconstruction averages all $P$ bins of each active patch, including empty-bin targets that equal zero after local standardization based only on nonempty bins. Commitment updates queries, while codebooks use exponential moving averages (EMA).

Query isotropy discourages concentration in a small number of embedding directions. For $n\ge2$ sampled valid queries at one scale, let $z_a=\operatorname{norm}(q_a^{S,h})$, $\mu=n^{-1}\sum_a z_a$, and $\Sigma=n^{-1}\sum_a(z_a-\mu)(z_a-\mu)^\top$. Define
\begin{equation}
 \delta=\max\!\left(\operatorname{sg}
 \left(\frac{\operatorname{tr}\Sigma}{d_q}\right),10^{-6}\right),
 \qquad
 A_{jk}=\frac{\Sigma_{jk}}
 {\sqrt{\max(\Sigma_{jj},10^{-6})\max(\Sigma_{kk},10^{-6})}}.
\end{equation}
The per-scale penalty is
\begin{equation}
 I_h=\|\mu\|_2^2
 +\frac1{d_q}\sum_{j=1}^{d_q}
 \left(\frac{\Sigma_{jj}}{\delta}-1\right)^2
 +\frac1{d_q}\sum_{j\ne k}A_{jk}^2.
\label{eq:app-state-isotropy}
\end{equation}
$\mathcal L_{\mathrm{iso}}$ averages $I_h$ equally over scales with at least two valid queries, with at most 2,048 queries each.

Local consistency regularizes queries and assignments under shape perturbations. For a normalized patch $\widetilde p$, construct $\widetilde p'=\operatorname{StdNorm}(\widetilde p+0.03\epsilon)$, where $\epsilon\sim\mathcal N(0,I_P)$ and $\operatorname{StdNorm}$ subtracts the patch mean and divides by its population standard deviation clamped below at $10^{-6}$. Let $q,q'$ be their state queries. With detached, unit-normalized prototypes and temperature $\eta=0.1$, define
\begin{equation}
 p_k(q)=\frac{\exp(q^\top\operatorname{sg}(e_k^S)/\eta)}
 {\sum_l\exp(q^\top\operatorname{sg}(e_l^S)/\eta)},\qquad
 \ell_{\mathrm{loc}}=1-\cos(q,q')+\operatorname{JS}(p(q),p(q')).
\label{eq:app-state-consistency}
\end{equation}
Jensen--Shannon divergence $\operatorname{JS}$ clamps probabilities at $10^{-8}$ before logarithms. Both queries receive gradients. $\mathcal L_{\mathrm{loc}}$ averages up to 256 valid patches per scale, then nonempty scales equally.

Delayed negative separation distinguishes dissimilar shapes after warm-up. From at most 128 valid state queries per scale, construct the set $\mathcal P_h$ of unordered pairs that satisfy all four criteria: different training windows, different selected code IDs, prototype cosine similarity at most $0.5$, and normalized-shape cosine similarity at most $0.5$. The loss is
\begin{equation}
 \mathcal L_-=
 \frac1{|\mathcal H_-|}\sum_{h\in\mathcal H_-}
 \frac1{|\mathcal P_h|}\sum_{(a,b)\in\mathcal P_h}
 [\cos(q_a^{S,h},q_b^{S,h})-0.2]_+^2,
\label{eq:app-state-negative}
\end{equation}
where $\mathcal H_-$ contains scales with eligible pairs and $[x]_+=\max(x,0)$. Pair selection is detached; gradients pass only through query similarities. At optimizer step $s$, the weight is
\begin{equation}
 \rho_S(s)=\min\!\left(1,\max\!\left(0,\frac{s-5000}{5000}\right)\right).
\label{eq:app-state-negative-ramp}
\end{equation}
All state-query subsampling is uniform without replacement within each device's batch. A regularizer contributes zero when no scale supplies enough valid state queries or pairs.

\subsubsection{Transition learning}
\label{app:transition-objectives}

Transition learning optimizes the continuous patch-pair encoder, state-pair network, causal context encoder, and decoder using waveform reconstruction, residual matching, and query regularization:
\begin{equation}
 \mathcal L_{\mathrm{enc}}=
 \mathcal L_{\mathrm{raw}}
 +\lambda_{\mathrm{ms}}\mathcal L_{\mathrm{multiscale}}
 +\lambda_T\mathcal L_{\mathrm{match}}^T
 +\lambda_{\mathrm{vc}}\mathcal L_{\mathrm{varcov}}^T.
\label{eq:app-encoder-objective}
\end{equation}

The decoder reconstructs waveforms coarse to fine using learned queries at the coarsest scale, an interpolation bridge with a residual multilayer perceptron at finer scales, and scale-specific linear heads producing $P$ values per interval. Each window follows one route across all scales: joint, state, or transition-driven reconstruction, with probabilities $0.30$, $0.50$, and $0.20$. The state route uses only the current state without the cross-scale bridge. The transition-driven route uses preceding states and transitions through the current interval, initializing from a state when preceding information is unavailable; the joint route additionally accesses the current state.

Let $v_{uib}^h$ indicate that bin $b$ and its enclosing patch are valid. For route predictions $\widehat p_{uib}^h$,
\begin{equation}
\begin{aligned}
 R_h&=\frac{\sum_{u,i,b}v_{uib}^h
 \left(|\widehat p_{uib}^h-p_{uib}^h|
 +(\widehat p_{uib}^h-p_{uib}^h)^2\right)}
 {2\max(1,\sum_{u,i,b}v_{uib}^h)},\\
 \mathcal L_{\mathrm{raw}}&=R_H,\qquad
 \mathcal L_{\mathrm{multiscale}}=\frac1{H-1}\sum_{h=1}^{H-1}R_h.
\end{aligned}
\label{eq:app-waveform-reconstruction}
\end{equation}
Scale $H$ is finest. Because $v_{uib}^h$ combines bin and patch validity, both waveform losses exclude empty bins and normalize by the number of valid bins. This differs from normalized-shape reconstruction, which retains their zero-valued targets. Each loss is estimated over the sampled decoding route.

For each valid adjacent pair, the layer-normalized residual $r_i^h$ (Eq.~\plaineqref{eq:transition-residual}) and $\log(1+\|r_i^h\|_2)$ are projected to a local query, contextualized by causal attention before quantization. The selected vector predicts $\widehat r_i^h=W_R\tau_i^h$, matched to the detached target $r_*=\operatorname{sg}(r_i^h)$ by
\begin{equation}
 \ell_T(r_*,\widehat r)=2-2\cos(r_*,\widehat r)
 +0.1\left[\log(1+\|r_*\|_2)
 -\log(1+\|\widehat r\|_2)\right]^2.
\label{eq:app-transition-match}
\end{equation}
Cosine similarity uses $\epsilon=10^{-6}$; residual norms are clamped below at $10^{-6}$ before $\log(1+\cdot)$. $\mathcal L_{\mathrm{match}}^T$ averages over valid transitions within each scale, then over scales. Gradients pass through the straight-through transition vector and prediction head; only this loss detaches its target.

Variance--covariance regularization follows the spread and decorrelation principles of VICReg~\citep{bardes2022vicreg} and acts on both local and contextual transition queries. For either query set at scale $h$, sample $n\ge2$ valid queries and set $z_a=\sqrt{d_q}\operatorname{norm}(q_a)$, $\bar z=n^{-1}\sum_a z_a$. Define
\begin{equation}
 \sigma_j=\sqrt{\frac1n\sum_a(z_{aj}-\bar z_j)^2+10^{-4}},\qquad
 V=\frac1{n-1}\sum_a(z_a-\bar z)(z_a-\bar z)^\top,
\end{equation}
and
\begin{equation}
 J_h(q)=\frac1{d_q}\sum_j[1-\sigma_j]_+
 +\frac1{d_q}\sum_{j\ne k}V_{jk}^2,\qquad
 \mathcal L_{\mathrm{varcov}}^T
 =\frac12\left(\langle J_h(q^{T,\mathrm{local}})\rangle_h
 +\langle J_h(q^{T,\mathrm{context}})\rangle_h\right).
\label{eq:app-transition-varcov}
\end{equation}
Each average uses 2--2,048 valid queries per eligible scale and query set.

\begin{table}[t]
\small
\centering
\caption{Objective weights used in state warm-up and transition learning.
The negative-separation weight follows the ramp $\rho_S(s)$.}
\label{tab:representation-loss-weights}
\renewcommand{\arraystretch}{0.75}
\begin{tabular}{llr}
\toprule
Stage & Objective & Weight \\
\midrule
State & Normalized-shape reconstruction & 1.00 \\
 & Code commitment & 1.00 \\
 & Query isotropy & 0.10 \\
 & Perturbation consistency & 0.05 \\
 & Delayed negative separation & $0.02\rho_S(s)$ \\
\midrule
Transition & Full-resolution reconstruction & 1.00 \\
 & Multiscale reconstruction & 1.00 \\
 & Direction--magnitude matching & 1.00 \\
 & Query variance--covariance & 0.05 \\
\bottomrule
\end{tabular}
\end{table}

\subsubsection{Codebook optimization}
\label{app:codebook-optimization}

Both branches use unit-normalized queries and codebook entries with deterministic nearest-code assignment (Eqs.~\plaineqref{eq:state-assignment} and~\plaineqref{eq:transition-assignment}) and straight-through gradients~\citep{oord2017vqvae}, e.g., $q_i^{S,h}+\operatorname{sg}(s_i^h-q_i^{S,h})$. EMA statistics instead use Sinkhorn-balanced hard assignments $a_{ik}$ over equal numbers of valid queries sampled from each nonempty scale. With decay $\beta=0.99$,
\begin{equation}
\begin{aligned}
 n_k&\leftarrow\beta n_k+(1-\beta)\sum_i a_{ik},\\
 m_k&\leftarrow\beta m_k+(1-\beta)\sum_i a_{ik}q_i,\\
 e_k&\leftarrow\operatorname{norm}\!\left(\frac{m_k}{n_k+\epsilon}\right),
\end{aligned}
\label{eq:app-state-ema}
\end{equation}
where $\operatorname{norm}(z)=z/\max(\|z\|_2,10^{-6})$. EMA uses nearest-code assignments when sampled queries are fewer than codebook entries and reinitializes persistently inactive entries from current queries. Neighbor smoothing, periodic whitening and repulsion, and covariance correction modify EMA codebook updates separately from gradient-based objectives. Objective weights appear in Table~\ref{tab:representation-loss-weights}; training schedules and optimizer settings are in Appendix~\ref{app:training-configuration}.
\endgroup

\subsection{Multiscale state--transition condition}

\revisionblue{After representation pretraining, the encoder and both codebooks are frozen. For variable $d$ at scale $h$, the ordered condition sequence interleaves states and transitions in temporal order:}
\begin{equation}
    C_{\mathrm{seq},d}^h=[s_{d,0}^h,\tau_{d,1}^h,s_{d,1}^h,\ldots,\tau_{d,n_h-1}^h,s_{d,n_h-1}^h].
    \label{eq:app-condition-sequence}
\end{equation}
\revisionblue{State $s_{d,i}^h$ lies at interval center $(i+\tfrac12)w_h$ and transition $\tau_{d,i}^h$ at boundary $iw_h$. Masks retain active states and transitions within valid segments.}

\revisionblue{Separate state and transition projections map the elements of $C_{\mathrm{seq},d}^h$ into a common conditioning space. For any active hierarchy suffix, each fine-scale element incorporates the mean of overlapping projected elements from its nearest active coarser scale $\pi(h)$. A learned layer fuses this summary with the local element, and within-scale self-attention produces the scale-$h$ condition memory $C_d^h$.}

\revisionblue{Condition injection follows Eq.~\plaineqref{eq:main-cross-scale-conditioning} with masked interval pooling. Active scales update coarse to fine with query repooling. Attention respects validity, segment identity, and temporal overlap.}

\subsection{Conditional flow optimization and zero-shot sampling}

\revisionblue{For single-variable inputs $X_0,X_1,X_t\in\mathbb R^L$, condition $C$, validity mask $M$, and target $U_t=X_1-X_0$ (Eq.~\plaineqref{eq:fm-path}), the full-resolution loss averages over valid positions per sequence, then over the batch:}
\begin{equation}
    \mathcal{L}_{\mathrm{full}}
    =\mathbb{E}\left[
    \frac{\|M\odot(v_\theta(X_t,t,C)-U_t)\|_2^2}
    {\|M\|_1+\epsilon}\right].
    \label{eq:app-full-flow-loss}
\end{equation}
\revisionblue{For every active scale $h$, masked pooling defines the target $U_t^h=\operatorname{Pool}_h(U_t)$ for an auxiliary velocity head $v_{\theta,h}$. The complete objective is}
\begin{equation}
    \mathcal{L}_{\mathrm{flow}}=\mathcal{L}_{\mathrm{full}}
    +\sum_{h\in\mathcal{H}_{\mathrm{act}}}\lambda_h
    \mathbb{E}\left[
    \frac{\|M^h\odot(v_{\theta,h}(X_t,t,C)-U_t^h)\|_2^2}
    {\|M^h\|_1+\epsilon}\right].
    \label{eq:app-complete-flow-loss}
\end{equation}
\revisionblue{Training uses all five scales or a suffix starting at span $128$, $64$, $32$, or $16$; inactive scales contribute neither loss nor gradient. For $Y=(X_t,t,C)$ and fixed linear masked pooling $P_h$, squared-loss population targets satisfy $v^\star(Y)=\mathbb E[U_t\mid Y]$ and $v_h^\star(Y)=\mathbb E[P_hU_t\mid Y]=P_hv^\star(Y)$ by linearity. This consistency applies on the training-condition support~\citep{lipman2023flowmatching}.}

\revisionblue{Sampling follows Section~\ref{sec:reference-zero-shot-generation}, with reference normalization and rescaling defined in Appendix~\ref{app:preprocessing}.}

\subsection{\revisionblue{Training and implementation settings}}
\label{app:training-configuration}

\revisionblue{Tables~\ref{tab:pretraining-stages} and~\ref{tab:app-implementation} summarize training and model settings with $P=16$ canonical values. State warm-up starts randomly; transition learning resets the optimizer and scheduler, following Appendix~\ref{app:representation-learning}. Flow optimization freezes the encoder and both codebooks.}

\revisionblue{Flow training uses univariate windows of length 512--4096 in increments of 256, with one length per batch. The final phase assigns probability $0.5$ to length 4096 and total probability $0.7$ to lengths at least 3072. The full-resolution loss has unit weight; auxiliary weights appear in Table~\ref{tab:app-implementation}.}

\revisionblue{The final flow phase initializes from a step-2,000 generator EMA checkpoint pretrained on the same GIFT-Eval Pretrain corpus, resetting the optimizer, scheduler, and EMA state. Linear warm-up precedes cosine decay over a planned 4,000-step schedule to a minimum learning-rate ratio of $0.1$. Evaluation uses the checkpoint after 2,400 additional updates. Sampling uses EMA parameters, nearest-code assignments, and the integration settings in Table~\ref{tab:app-implementation}; the main length-512-to-168 evaluation uses seeds 2021--2025.}

\revisionblue{Pretraining uses NVIDIA A800 80GB GPUs; downstream experiments and generation use NVIDIA RTX 4090 24GB GPUs. Before stage freezing, the representation model, including its reconstruction decoder, has 22.87M parameters; the flow generator has 24.57M. The transition context encoder has six causal Transformer layers of width 384 with eight heads; the reconstruction decoder has three layers per scale. Encoder training standardizes a finite segment of at least 512 points selected from each 4096-point row, whereas flow training standardizes the complete training row before cropping. Both use population standard deviations and a constant-scale fallback of one.}

\begin{table}[t]
\small
\caption{\revisionblue{Pretraining settings by stage. All stages use AdamW, bfloat16, and gradient clipping at norm 1.}}
\label{tab:pretraining-stages}
\centering
\small
\setlength{\tabcolsep}{4pt}
\renewcommand{\arraystretch}{0.75}
\begin{tabular}{lrrrr}
\toprule
Stage & Updates & Learning rate & LR warm-up & Effective batch \\
\midrule
State warm-up & 20,000 & $10^{-4}$ & 200 & 84 ($4\times21$) \\
State--transition & 160,000 & $10^{-4}$ & 1,600 & 84 ($4\times21$) \\
Flow, final phase & 2,400 & $10^{-5}$ & 100 & 10,240 ($8\times1280$) \\
\bottomrule
\end{tabular}
\end{table}

\revisionblue{AdamW uses $(\beta_1,\beta_2)=(0.9,0.95)$, weight decay $0.01$, and $\epsilon=10^{-8}$ throughout. Encoder schedules use linear warm-up and zero final rates. State--transition training resumes the same schedule at steps 120,000--160,000 with covariance-regularized codebook updates. Flow training uses activation checkpointing.}

\begin{table}[t]
\small
    \caption{\revisionblue{Representation and conditional flow configurations used by MarkovTS.}}
    \label{tab:app-implementation}
    \centering
    \small
    \renewcommand{\arraystretch}{0.75}
    \begin{tabular}{ll}
        \toprule
        Component & Configuration \\
        \midrule
        \multicolumn{2}{l}{\revisionblue{Representation configuration}} \\
        Base patch / temporal spans & $16$ / $\{16,32,64,128,256\}$ \\
        State codebook & $K_S=1024$, dimension $64$, EMA $0.99$ \\
        Transition codebook & $K_T=1024$, dimension $64$, EMA $0.99$ \\
        \midrule
        \multicolumn{2}{l}{\revisionblue{Conditional flow: final optimization phase}} \\
        Flow backbone & width $288$, depth $8$, heads $8$, MLP ratio $4$ \\
        Auxiliary flow losses & $\lambda_h=0.1$ for each active scale \\
        Flow parameter EMA & $0.9995$ \\
        Euler sampling & $16$ function evaluations \\
        \bottomrule
    \end{tabular}
\end{table}

\subsection{Generation throughput}
\label{app:generation-throughput}

\revisionblue{Table~\ref{tab:generation-throughput} reports throughput for the selected EMA model with 16 Euler steps on four RTX 4090 24GB GPUs. Values are medians over 12 datasets and five seeds, aggregated across GPUs and counting complete sequences before evaluation-window extraction. Flow-only timing covers numerical integration; end-to-end timing also includes condition preparation and output processing, including CPU-side checks and file writing.}

\begin{table}[t]
\small
\caption{\revisionblue{Generation throughput aggregated across four RTX 4090 24GB GPUs.
Rates count complete sequences per second; batch sizes are per GPU.}}
\label{tab:generation-throughput}
\centering
\small
\setlength{\tabcolsep}{7pt}
\renewcommand{\arraystretch}{0.75}
\begin{tabular}{rrrr}
\toprule
Length & Batch/GPU & Flow-only & End-to-end \\
\midrule
512 & 128 & 323.02 & 128.89 \\
1024 & 96 & 221.13 & 74.40 \\
2048 & 64 & 166.73 & 43.75 \\
\bottomrule
\end{tabular}
\end{table}

\section{Theoretical Properties of MarkovTS}
\label{app:theory}

\revisionblue{We analyze state-identity stability and reconstruction error under transition residual matching.}

\subsection{Invariant and stable state identities}

\paragraph{Proposition 1 (Affine-invariant state assignment).}
\revisionblue{Let $p\in\mathbb{R}^{P}$ be a canonical patch with nonempty-bin set $\mathcal V$ and $\sigma_{\mathrm v}(p)>0$, where $\mu_{\mathrm v}$ and $\sigma_{\mathrm v}$ use only indices in $\mathcal V$. Define $N_{\mathrm v}(p)=(p-\mu_{\mathrm v}(p)\mathbf{1})/\sigma_{\mathrm v}(p)$. For any $a>0$ and $b\in\mathbb{R}$,}
\begin{equation}
    N_{\mathrm v}(ap+b\mathbf{1})=N_{\mathrm v}(p).
    \label{eq:app-affine-invariance}
\end{equation}
\revisionblue{Thus, a state query computed solely from $N_{\mathrm v}(p)$ retains its deterministic code assignment under positive rescaling and additive offsets. For a unit-normalized query $q$ and unit-normalized code vectors $e_k$, maximizing cosine similarity is equivalent to nearest-neighbor assignment on the unit sphere. If}
\begin{equation}
    \gamma=q^\top e_{k^\star}-\max_{\ell\ne k^\star}q^\top e_\ell>0,
    \label{eq:app-code-margin}
\end{equation}
then every unit query $q'$ satisfying $\|q'-q\|_2<\gamma/2$ retains assignment $k^\star$.

\emph{Proof.} The validity mask, and hence $\mathcal V$, is unchanged by the affine transformation. For $a>0$, $\mu_{\mathrm v}(ap+b\mathbf{1})=a\mu_{\mathrm v}(p)+b$ and $\sigma_{\mathrm v}(ap+b\mathbf{1})=a\sigma_{\mathrm v}(p)$, which proves Eq.~\plaineqref{eq:app-affine-invariance}. For unit vectors, $\|q-e_k\|_2^2=2-2q^\top e_k$, so maximizing cosine similarity minimizes Euclidean distance. For every $\ell\ne k^\star$,
\begin{equation}
    (q')^\top(e_{k^\star}-e_\ell)
    \ge \gamma-2\|q'-q\|_2>0,
\end{equation}
by Cauchy--Schwarz and $\|e_k\|_2=1$. Thus $k^\star$ remains the unique maximizer. \hfill$\square$

\revisionblue{Invariance requires an inactive standard-deviation floor for both patches. With the assignment margin, it supports morphological-code reuse under level and amplitude changes.}

\subsection{Transition residual and relation fidelity}

\paragraph{Proposition 2 (Transition-residual control).}
Let $g$ be the patch-pair relation, $b$ the corresponding state-pair relation, $r=g-b$, and $\hat g=b+\hat r$ the relation reconstructed from an assigned transition vector. Then
\begin{equation}
    \|g-\hat g\|_2^2=\|r-\hat r\|_2^2.
    \label{eq:app-residual-identity}
\end{equation}
\revisionblue{Assume $0<m\le\|r\|_2,\|\hat r\|_2\le M$ and $\alpha>0$. For}
\begin{equation}
    \ell_T^\alpha(r,\hat r)=2-2\cos(r,\hat r)
    +\alpha\big[\log(1+\|r\|_2)-\log(1+\|\hat r\|_2)\big]^2,
\end{equation}
the residual error satisfies
\begin{equation}
    \|r-\hat r\|_2^2
    \le \kappa\ell_T^\alpha(r,\hat r),\qquad
    \kappa=\max\left\{M^2,\frac{(1+M)^2}{\alpha}\right\}.
    \label{eq:app-transition-bound}
\end{equation}

\emph{Proof.} The first identity follows from $g-(b+\hat r)=r-\hat r$. Writing $a=\|r\|_2$, $\hat a=\|\hat r\|_2$, and $\vartheta=\cos(r,\hat r)$ gives
\begin{equation}
    \|r-\hat r\|_2^2=a\hat a(2-2\vartheta)+(a-\hat a)^2.
\end{equation}
The first term is at most $M^2(2-2\vartheta)$. By the mean-value theorem for $z\mapsto\log(1+z)$ on $[0,M]$, $|a-\hat a|\le(1+M)|\log(1+a)-\log(1+\hat a)|$. Combining the bounds proves Eq.~\plaineqref{eq:app-transition-bound}. \hfill$\square$

\revisionblue{Under the stated norm bounds, the direction--magnitude objective in Eq.~\plaineqref{eq:app-transition-match} bounds the error of patch-pair relations reconstructed from state-pair relations and predicted transition residuals.}

\section{Experimental Details}
\label{app:experimental-details}

\subsection{Pretraining corpus and generation datasets}
\label{app:generation-datasets}

We pretrain on GIFT-Eval Pretrain~\citep{aksu2024gifteval}, excluding the 12 generation datasets in Table~\ref{tab:generation-data} before training-window extraction. The pretraining corpus follows the official GIFT-Eval Pretrain dataset listing linked in Section~\ref{sec:experimental-setup}.

\revisionblue{The 12 held-out public GluonTS datasets~\citep{huang2025timedp} cover energy, transportation, environmental measurements, and economic activity. Table~\ref{tab:generation-data} summarizes their source data, sampling frequencies, and temporal coverage before window extraction. The main comparison evaluates non-overlapping length-168 windows extracted from length-512 references and generated outputs.}

\begin{table*}[h!]
\small
\caption{Held-out datasets for generation evaluation.}
\label{tab:generation-data}
\centering
\small
\setlength{\tabcolsep}{2.4pt}
\renewcommand{\arraystretch}{0.75}
\begin{tabular}{llllp{0.43\textwidth}}
\toprule
Dataset & Application & Frequency & Scale & Content and temporal extent \\
\midrule
Electricity & Energy & Hourly & 321 clients & Electricity consumption, 2012--2014 \\
Solar & Energy & Hourly & 137 series & Solar power production in Alabama, 2006 \\
Wind & Energy & 4 seconds & 1 long series & AEMO wind production, beginning August 2019 \\
Traffic & Transportation & 10 minutes & 963 sensors & Freeway-lane occupancy in the San Francisco Bay Area over 440 days \\
Taxi & Transportation & 30 minutes & 1,214 locations & New York City taxi rides in January 2015 and January 2016 \\
Pedestrian & Transportation & Hourly & 66 sensors & Pedestrian counts in Melbourne, May 2009--April 2020 \\
Air Quality & Environment & Hourly & 270 series & Pollutants from 59 Beijing/London stations, 2017--2018 \\
Temperature & Environment & Daily & 422 stations & Mean temperature measurements across Australia, 2015--2017 \\
Rain & Environment & Daily & 422 stations & Rainfall measurements from the same Australian station collection \\
NN5 & Banking & Daily & 111 series & Cash withdrawals from automated teller machines \\
Fred-MD & Macroeconomics & Monthly & 107 series & Transformed U.S. macroeconomic indicators \\
Exchange & Finance & Daily & 8 series & Exchange rates among eight currencies \\
\bottomrule
\end{tabular}
\end{table*}

\revisionblue{Generation uses target training references and frozen MarkovTS parameters. Test data are excluded from generation and downstream fitting; validation follows the task-specific protocols below.}

\subsection{Generation evaluation protocol and baselines}
\label{app:generation-protocol}
\label{app:generation-baselines}

\revisionblue{For the main evaluation, we select $K_{\mathrm{eval}}=100$ valid length-512 univariate reference windows extracted from the channels of the target training split. The frozen MarkovTS encoder assigns deterministic nearest-code states and transitions, and one fixed flow checkpoint generates length-512 outputs from independent Gaussian sources. Each output yields three fixed, non-overlapping length-168 windows in temporal order; the final eight points are discarded. For each reference, we generate outputs until at least 50 windows are available, order the candidates first by Gaussian draw and then by temporal position, and retain the first 50 (thus discarding the final candidate window). This gives $N=5000$ evaluation windows per dataset and seed. Table~\ref{tab:generation-l168} evaluates these windows against the test split over seeds 2021--2025. GraphFormer and reference-dependent codebook sampling are not used in this main protocol.}

\revisionblue{TimeVQVAE-C uses dataset identity as the class label during multi-domain training. ImagenFew results report five-seed means and sample standard deviations under the same length-168 evaluation \citep{gonen2025imagenfew}. The main table compares RBF-MMD and flat-KL. Wins are counted over the 24 MMD/flat-KL comparisons after rounding means to three decimals, crediting all tied methods.}

\subsection{Evaluation metrics}
\label{app:evaluation-metrics}

\paragraph{Generation fidelity.}
Let $\mathcal{X}_{r}=\{\mathbf{x}_{i}\}_{i=1}^{n_r}$ and $\mathcal{X}_{g}=\{\widetilde{\mathbf{x}}_{j}\}_{j=1}^{n_g}$ denote the real and generated evaluation windows, where each $\mathbf{x}\in\mathbb{R}^{L\times C}$. In the main length-512-to-168 benchmark, \revisionblue{channels are evaluated as univariate windows ($C=1$).} Before computing MMD and flat-KL, both sets are standardized by the pooled mean $\mu_r$ and standard deviation $\sigma_r$ of the real evaluation set:
\begin{equation}
\mathbf{x}'=\frac{\mathbf{x}-\mu_r}{\sigma_r+10^{-7}}.
\label{eq:metric-standardization}
\end{equation}
Both metrics are lower-is-better.

Maximum Mean Discrepancy (MMD) compares complete vectorized windows. With $\mathbf{z}_i=\operatorname{vec}(\mathbf{x}'_i)$, $\widetilde{\mathbf{z}}_j=\operatorname{vec}(\widetilde{\mathbf{x}}'_j)$, and the radial-basis-function kernel $k(\mathbf{u},\mathbf{v})=\exp(-\gamma\lVert\mathbf{u}-\mathbf{v}\rVert_2^2)$, the evaluator reports the biased squared estimate
\begin{align}
\widehat{\operatorname{MMD}}^{2}
={}&\frac{1}{n_r^2}\sum_{i,i'}k(\mathbf{z}_i,\mathbf{z}_{i'})
+\frac{1}{n_g^2}\sum_{j,j'}k(\widetilde{\mathbf{z}}_j,
\widetilde{\mathbf{z}}_{j'}) \notag\\
&-\frac{2}{n_rn_g}\sum_{i,j}k(\mathbf{z}_i,
\widetilde{\mathbf{z}}_j),
\label{eq:metric-mmd}
\end{align}
\revisionblue{The kernel uses $\gamma=1/(LC)$, and tables retain the conventional label MMD.}

Flattened Kullback--Leibler divergence (flat-KL) discards the sample, time, and channel indices and pools all scalar observations. The range of the real set defines $B=50$ equal-width histogram bins. Let $p_b$ and $q_b$ be the normalized real and generated masses in bin $b$, and let $\bar q_b=(q_b+\epsilon)/\sum_{a=1}^{B}(q_a+\epsilon)$ with $\epsilon=10^{-9}$. The reported divergence is
\begin{equation}
\operatorname{flat\mbox{-}KL}(\mathcal{X}_{r}\Vert\mathcal{X}_{g})
=\sum_{b=1}^{B}p_b\log\frac{p_b}{\bar q_b}.
\label{eq:metric-flat-kl}
\end{equation}

Context-Fr\'echet Inception Distance (Context-FID) supplies the complementary representation-space comparison with ImagenFew~\citep{gonen2025imagenfew}. For each dataset $d$, a separately trained TS2Vec encoder $E_d$ \citep{yue2022ts2vec} maps a sequence to a contextual embedding. If $(\boldsymbol{\mu}_{r},\boldsymbol{\Sigma}_{r})$ and $(\boldsymbol{\mu}_{g},\boldsymbol{\Sigma}_{g})$ are the empirical means and covariances of real and generated embeddings, respectively, then
\begin{equation}
\operatorname{ContextFID}
=\lVert\boldsymbol{\mu}_{r}-\boldsymbol{\mu}_{g}\rVert_2^2
+\operatorname{Tr}\!\left(
\boldsymbol{\Sigma}_{r}+\boldsymbol{\Sigma}_{g}
-2(\boldsymbol{\Sigma}_{r}\boldsymbol{\Sigma}_{g})^{1/2}\right).
\label{eq:context-fid}
\end{equation}
\revisionblue{Lower Context-FID is better. TS2Vec only encodes evaluation samples; MarkovTS remains fixed.}

\paragraph{Forecasting utility.}
For $N$ test windows, forecast horizon $H$, and $C$ variables, let $y_{ihc}$ and $\widehat y_{ihc}$ denote a target and its prediction. We report
\begin{equation}
\operatorname{MAE}=\frac{1}{NHC}\sum_{i,h,c}
\lvert y_{ihc}-\widehat y_{ihc}\rvert,\qquad
\operatorname{MSE}=\frac{1}{NHC}\sum_{i,h,c}
(y_{ihc}-\widehat y_{ihc})^2.
\label{eq:forecast-metrics}
\end{equation}
\revisionblue{Both metrics are lower-is-better. Table~\ref{tab:forecast-aug} averages the scores over four forecast horizons.}

\paragraph{Classification utility.}
Let $n_c$ be the number of test instances in class $c$, $w_c=n_c/N$, and $P_c,R_c$ denote one-vs-rest precision and recall. With $F_c=2P_cR_c/(P_c+R_c)$, the reported metrics are
\begin{align}
\operatorname{Accuracy}&=\frac1N\sum_i\mathbb I[\widehat y_i=y_i],
&\operatorname{BAcc}&=\frac1C\sum_c R_c,\notag\\
\operatorname{F1}&=\begin{cases}F_{c_{\min}},&C=2,\\
\sum_c w_cF_c,&C>2,\end{cases}
&\operatorname{AUROC}&=\begin{cases}\operatorname{AUROC}_{1},&C=2,\\
\sum_c w_c\operatorname{AUROC}_c,&C>2.
\end{cases}
\label{eq:classification-metrics}
\end{align}
\revisionblue{The indices $c_{\min}$ and $c_{\max}$ identify the least and most frequent test classes, with ties resolved by the smallest and largest encoded labels, respectively. Binary F1 treats $c_{\min}$ as positive, whereas binary AUROC uses the probability of encoded class 1. Multiclass AUROC uses prevalence-weighted one-vs-rest scores. Undefined precision, recall, and F1 contributions are set to zero.}

Sensitivity and specificity use $R_{c_{\min}}$ and $R_{c_{\max}}$ for binary tasks, respectively. Following Multiverse, both use prevalence-weighted recall $\sum_c w_cR_c$ for multiclass tasks.

For predicted class probabilities $\widehat p_{ic}$, log loss is
\begin{equation}
\operatorname{LL}=-\frac1N\sum_i\log\widehat p_{i,y_i}.
\label{eq:classification-logloss}
\end{equation}
\revisionblue{All measures except log loss are higher-is-better. Dataset-level results follow the binary or multiclass convention before equal-weight averaging over the 23 datasets.}

\paragraph{Anomaly-detection utility.}
Let $s_t$ be an anomaly score, $y_t\in\{0,1\}$ the pointwise label, and $\widehat y_t(\tau)=\mathbb{I}[s_t\geq\tau]$. At threshold $\tau$, precision, recall, and false-positive rate are
\begin{equation}
P(\tau)=\frac{\mathrm{TP}}{\mathrm{TP}+\mathrm{FP}},\qquad
R(\tau)=\frac{\mathrm{TP}}{\mathrm{TP}+\mathrm{FN}},\qquad
F(\tau)=\frac{\mathrm{FP}}{\mathrm{FP}+\mathrm{TN}}.
\label{eq:anomaly-point-rates}
\end{equation}
Pointwise PR-AUC (A-P, implemented as average precision) and pointwise ROC-AUC (A-R) summarize the threshold sweep as
\begin{equation}
\operatorname{A\mbox{-}P}=\sum_k(R_k-R_{k-1})P_k,\qquad
\operatorname{A\mbox{-}R}=\int_{0}^{1}R(F)\,dF.
\label{eq:anomaly-point-auc}
\end{equation}
Following TAB/OpenTS~\citep{qiu2025tab}, the range-aware operator softly expands each ground-truth anomaly interval by a tolerance window and combines overlap with an existence reward. Let $P_w(\tau)$, $R_w(\tau)$, and $F_w(\tau)$ be the resulting range-aware rates for tolerance $w$, and let $\ell$ be the median ground-truth anomaly-range length. Range-aware PR-AUC and ROC-AUC are
\begin{equation}
\operatorname{R\mbox{-}A\mbox{-}P}=\int P_{\ell}\,dR_{\ell},\qquad
\operatorname{R\mbox{-}A\mbox{-}R}=\int R_{\ell}\,dF_{\ell}.
\label{eq:anomaly-range-auc}
\end{equation}
\revisionblue{The volumes under the PR and ROC surfaces average range-aware areas over tolerance widths~\citep{paparrizos2022volume}:}
\begin{equation}
\operatorname{V\mbox{-}PR}=\frac{1}{2\ell+1}\sum_{w=0}^{2\ell}
\int P_w\,dR_w,\qquad
\operatorname{V\mbox{-}ROC}=\frac{1}{2\ell+1}\sum_{w=0}^{2\ell}
\int R_w\,dF_w.
\label{eq:anomaly-vus}
\end{equation}
\revisionblue{All six anomaly-detection metrics are threshold-free and higher-is-better.}

\subsection{Forecasting datasets and evaluation protocol}
\label{app:forecast-protocol}

\revisionblue{The forecasting evaluation uses six datasets~\citep{li2026apt}; Table~\ref{tab:forecast-data} lists training statistics and chronological splits. ETTh1 and ETTh2 contain two-year transformer load and oil-temperature records.}

\revisionblue{Each forecasting dataset contains one continuous multivariate training sequence. Dataset size counts its length times its channel count before augmentation or sliding-window extraction.}

\begin{table}[h!]
\small
\caption{Forecasting datasets and chronological splits. Mean length counts training time steps; training values count scalar observations.}
\label{tab:forecast-data}
\centering
\small
\setlength{\tabcolsep}{3pt}
\renewcommand{\arraystretch}{0.75}
\begin{tabular}{lllrrrr}
\toprule
Dataset & Application & Frequency & Ch. & Mean length & Training values & Split \\
\midrule
ECL & Energy & Hourly & 321 & 18,412 & 5,910,252 & 7:1:2 \\
ETTh1 & Energy & Hourly & 7 & 8,640 & 60,480 & 6:2:2 \\
ETTh2 & Energy & Hourly & 7 & 8,640 & 60,480 & 6:2:2 \\
Exchange & Economy & Daily & 8 & 5,311 & 42,488 & 7:1:2 \\
Traffic & Transportation & Hourly & 862 & 12,280 & 10,585,360 & 7:1:2 \\
Weather & Environment & 10 minutes & 21 & 36,887 & 774,627 & 7:1:2 \\
\bottomrule
\end{tabular}
\end{table}

\revisionblue{Forecasting uses CATS, iTransformer, and TQNet~\citep{kim2024cats,liu2024itransformer,lin2025tqnet}, SparseTSF~\citep{lin2024sparsetsf}, and OLinear~\citep{yue2025olinear}. Table~\ref{tab:forecast-model-config} lists configurations.}

\begin{table*}[h!]
\small
\caption{Forecasting-backbone configurations.}
\label{tab:forecast-model-config}
\centering
\small
\setlength{\tabcolsep}{4pt}
\setbox0=\hbox\bgroup
\renewcommand{\arraystretch}{0.75}
\begin{tabular}{llll}
\toprule
Backbone & Architecture & Regularization & Learning rate \\
\midrule
CATS &
\(d=128\); FFN 256; 16 heads; 3 layers; patch/stride 24/24 &
dropout 0.20; RevIN &
\(3\times10^{-4}\) \\
iTransformer &
\(d=256\); FFN 512; 1 head; 1 layer &
dropout 0.25; RevIN &
\(5\times10^{-5}\) \\
SparseTSF &
period 4 &
--- &
\(6\times10^{-4}\) \\
TQNet &
\(d=512\); 4 heads; dataset cycle length &
dropout 0; RevIN &
\(7.5\times10^{-5}\) \\
OLinear &
\(d=256\); FFN 256; 2 layers; patch/stride 16/8 &
dropout 0.20; embedding 8 &
\(3.75\times10^{-5}\) \\
\bottomrule
\end{tabular}
\egroup\leavevmode
\ifdim\wd0>\textwidth
\resizebox{\textwidth}{!}{\box0}%
\else
\box0
\fi
\end{table*}

\revisionblue{Forecasting batch sizes are 64, except 16 for ECL and 32 for Weather; TQNet uses 16 throughout. Weight decay is $10^{-4}$ for iTransformer and zero otherwise.}

\revisionblue{Forecasting uses lookback 336 and horizons \(\{96,192,336,720\}\), with channel-wise normalization and seed 42. Models minimize MAE for at most 100 epochs, using a MultiStep schedule (milestones 1 and 25, factor 0.5), gradient-norm clipping at 5, and early-stopping patience 15. Validation is performed each epoch; the selected checkpoint is evaluated once on the real test split.}

\revisionblue{For forecasting augmentation, reference standard deviations below $10^{-6}$ are replaced by the corresponding training-split standard deviations, or by one if the training-split values are also below $10^{-6}$.}

\revisionblue{For forecasting, a real training sequence is partitioned into consecutive, non-overlapping reference segments: full segments contain exactly 4096 time steps, followed by one optional remainder segment. A remainder of length $L_s<4096$ is masked and right-padded to $G_s=\max(512,256\lceil L_s/256\rceil)$, and its generated output is cropped back to $L_s$. Each reference segment conditions exactly one generated segment. We concatenate the generated segments in their original order, producing one generated-only sequence with exactly the same length as the real sequence; no real values are inserted into this sequence. Sliding windows are extracted from the complete real and generated sequences. Windows may cross a join between two adjacent generated segments, but no window contains both real and generated values. A real window and its generated counterpart use the same start and end indices, so the two window pools have equal cardinality.}

\revisionblue{Under \emph{Real + MarkovTS}, we first sample the real-window indices for a batch, then uniformly choose exactly half of those indices and substitute the position-matched generated windows. Batch size and the number of updates per epoch are therefore identical to \emph{Real only}. Maximum epochs, the learning-rate schedule, validation indices, the test set, and early-stopping rules are also shared. \emph{Online} uses the same segmentation, generated-only concatenation, and window extraction. Its only difference is that fresh Gaussian sources regenerate all generated counterparts before every epoch; the epoch then uses only these generated windows while retaining the \emph{Real only} number of batches and updates.}

\subsection{Classification datasets and evaluation protocol}
\label{app:classification-protocol}

The classification study uses the 23 datasets shared by all four selected classifiers in the fixed Multiverse archive snapshot \citep{middlehurst2026multiverse}. Table~\ref{tab:classification-data} reports \revisionblue{official train/test sizes, task metadata, and mean training lengths and scalar counts after length equalization; parentheses show original variable-length ranges. The original test split is reserved for final evaluation.}

\begingroup
\fontsize{7.7}{8.8}\selectfont
\setlength{\tabcolsep}{3pt}
\renewcommand{\arraystretch}{1.00}
\begin{longtable}{lrrrrrrl}
\caption{Classification datasets used for downstream evaluation.}\label{tab:classification-data}\\
\toprule
Dataset & Train & Test & Ch. & Mean length & Training values & Cls. & Type \\
\midrule
\endfirsthead
\multicolumn{8}{c}{\normalsize\tablename\ \thetable\ (continued)}\\
\toprule
Dataset & Train & Test & Ch. & Mean length & Training values & Cls. & Type \\
\midrule
\endhead
\midrule
\multicolumn{8}{r}{Continued on next page}\\
\endfoot
\bottomrule
\endlastfoot
ArticularyWordRecognition & 275 & 300 & 9 & 144 & 356,400 & 25 & Motion \\
AtrialFibrillation & 15 & 15 & 2 & 640 & 19,200 & 3 & Biosignal \\
CharacterTrajectories & 1422 & 1436 & 3 & 119 (60--182) & 507,654 & 20 & Motion \\
Cricket & 108 & 72 & 6 & 1,197 & 775,656 & 12 & Motion \\
DuckDuckGeese & 50 & 50 & 1345 & 270 & 18,157,500 & 5 & Audio \\
ERing & 30 & 270 & 4 & 65 & 7,800 & 6 & Motion \\
Epilepsy & 137 & 138 & 3 & 206 & 84,666 & 4 & Motion \\
EthanolConcentration & 261 & 263 & 3 & 1,751 & 1,371,033 & 4 & Spectroscopy \\
HandMovementDirection & 160 & 74 & 10 & 400 & 640,000 & 4 & Biosignal \\
Handwriting & 150 & 850 & 3 & 152 & 68,400 & 26 & Motion \\
Heartbeat & 204 & 205 & 61 & 405 & 5,039,820 & 2 & Audio \\
JapaneseVowels & 270 & 370 & 12 & 25 (7--29) & 81,000 & 9 & Audio \\
LSST & 2459 & 2466 & 6 & 36 & 531,144 & 14 & Sensor \\
Libras & 180 & 180 & 2 & 45 & 16,200 & 15 & Motion \\
MotorImagery & 278 & 100 & 64 & 3,000 & 53,376,000 & 2 & Biosignal \\
NATOPS & 180 & 180 & 24 & 51 & 220,320 & 6 & Motion \\
PEMS-SF & 267 & 173 & 963 & 144 & 37,025,424 & 7 & Sensor \\
PhonemeSpectra & 3315 & 3353 & 11 & 217 & 7,912,905 & 39 & Audio \\
RacketSports & 151 & 152 & 6 & 30 & 27,180 & 4 & Motion \\
SelfRegulationSCP1 & 268 & 293 & 6 & 896 & 1,440,768 & 2 & Biosignal \\
SpokenArabicDigits & 6599 & 2199 & 13 & 65 (4--93) & 5,576,155 & 10 & Audio \\
StandWalkJump & 12 & 15 & 4 & 2,500 & 120,000 & 3 & Biosignal \\
UWaveGestureLibrary & 120 & 320 & 3 & 315 & 113,400 & 8 & Motion \\
\end{longtable}
\endgroup

\paragraph{Selected classifiers.}
\revisionblue{We use ResNet1D~\citep{wang2017timeseries}, TimesNet~\citep{wu2023timesnet}, ConvTran~\citep{foumani2024convtran}, and Arsenal~\citep{middlehurst2021hivecote}.}

\paragraph{Augmentation and evaluation.}
\revisionblue{Each real classification sequence $X_i$ has one generated counterpart $\widehat X_i$ with the same length and class label. For a reference of length $L\le4096$, generation uses reference-only normalization and masked right-padding to $G=\max(512,256\lceil L/256\rceil)$, followed by cropping to $L$ and inverse normalization. Thus the real and generated pools have the same cardinality and class counts, without temporal rescaling.}

\revisionblue{For each neural-classifier batch, we first sample real-sequence indices and then uniformly choose exactly half of them, replacing $X_i$ by $\widehat X_i$ at the selected indices. The batch size, number of updates per epoch, maximum training budget, and model-selection rule match \emph{Real only}. Arsenal does not use mini-batches: before fitting, we uniformly select exactly half of the original training indices once and replace those real sequences by their generated counterparts. Its mixed training set therefore has the same cardinality as the \emph{Real only} set.}

\revisionblue{The two settings reuse exactly the same real validation indices. A sequence assigned to validation and its generated counterpart are both excluded from checkpoint-selection training. ResNet1D uses validation performance to select the training duration and then refits on all original training indices; for \emph{Real + MarkovTS}, the same fixed-cardinality 50\% replacement rule is applied during this refit. TimesNet and ConvTran select checkpoints on the shared real validation subset. Arsenal fits once on all original training indices, using either all-real sequences or its fixed 50\% mixed set. Table~\ref{tab:classification-training-config} summarizes the training configurations.}

\revisionblue{Metrics follow the benchmark definitions in Appendix~\ref{app:evaluation-metrics}.}

\begin{table}[t]
\small
\centering
\caption{Classification training configurations.}
\label{tab:classification-training-config}
\small
\setlength{\tabcolsep}{5pt}
\renewcommand{\arraystretch}{0.75}
\begin{tabular}{llrr}
\toprule
Model & Optimizer / learning rate & Batch size & Maximum epochs \\
\midrule
ResNet1D & Adam / $10^{-3}$ & 64 & 100 \\
TimesNet & RAdam / $10^{-3}$ & 16 & 30 \\
ConvTran & RAdam / $10^{-3}$ & 16 & 100 \\
Arsenal & 25 members; 2,000 kernels each & --- & --- \\
\bottomrule
\end{tabular}
\end{table}

\subsection{Anomaly-detection datasets and evaluation protocol}
\label{app:anomaly-protocol}

\revisionblue{The anomaly-detection study uses 26 multivariate datasets from TAB/OpenTS~\citep{qiu2025tab}, comprising 215 series, listed in Table~\ref{tab:anomaly-augmentation-full}. Genesis, KDDcup99, and OPP are excluded from evaluation and aggregation because their official results are incomplete.}

\revisionblue{Table~\ref{tab:anomaly-data} summarizes the real training segments of these 215 series. Mean length is averaged over series within each dataset; channel ranges reflect differences between series. Training values sum scalar observations across all real training segments.}

\begin{table}[t]
\small
\centering
\caption{Anomaly-detection dataset statistics for the original training segments.}
\label{tab:anomaly-data}
\small
\setlength{\tabcolsep}{6pt}
\renewcommand{\arraystretch}{0.75}
\begin{tabular}{lrrrr}
\toprule
Dataset & Series & Mean length & Ch. & Training values \\
\midrule
ASD & 12 & 8,527.6 & 19 & 1,944,289 \\
CalIt2 & 1 & 2,520.0 & 2 & 5,040 \\
CATSv2 & 4 & 28,471.8 & 17 & 1,936,079 \\
CICIDS & 1 & 85,115.0 & 72 & 6,128,280 \\
Creditcard & 1 & 142,403.0 & 29 & 4,129,687 \\
DAPHNET & 7 & 29,241.0 & 9 & 1,842,183 \\
DLR & 1 & 11,565.0 & 9 & 104,085 \\
ECG & 1 & 56,104.0 & 32 & 1,795,328 \\
Exathlon & 30 & 15,804.4 & 7--31 & 10,022,209 \\
GECCO & 1 & 69,260.0 & 9 & 623,340 \\
GHL & 25 & 49,347.5 & 19 & 23,440,072 \\
GutenTAG & 48 & 10,000.0 & 2--20 & 3,940,000 \\
LTDB & 5 & 12,715.2 & 2--3 & 147,152 \\
MITDB & 6 & 35,416.7 & 2 & 425,000 \\
MSL & 1 & 58,317.0 & 55 & 3,207,435 \\
NYC & 1 & 13,104.0 & 3 & 39,312 \\
PSM & 1 & 132,481.0 & 25 & 3,312,025 \\
PUMP & 1 & 76,901.0 & 44 & 3,383,644 \\
SKAB & 34 & 9,405.0 & 8 & 2,558,160 \\
SMAP & 1 & 135,183.0 & 25 & 3,379,575 \\
SMD & 1 & 708,405.0 & 38 & 26,919,390 \\
SVDB & 6 & 23,760.0 & 2 & 285,120 \\
SWAN & 1 & 60,000.0 & 38 & 2,280,000 \\
SWAT & 1 & 495,000.0 & 51 & 25,245,000 \\
TAO & 12 & 2,000.0 & 3 & 72,000 \\
TODS & 12 & 20,000.0 & 5 & 1,200,000 \\
\bottomrule
\end{tabular}
\end{table}

\paragraph{Selected detectors.}
\revisionblue{The detectors include DLinear~\citep{zeng2023dlinear}, TranAD~\citep{tuli2022tranad}, CATCH~\citep{wu2025catch}, ModernTCN~\citep{luo2024moderntcn}, and TimesNet~\citep{wu2023timesnet}.}

\paragraph{Augmentation and evaluation.}
\revisionblue{Each anomaly-detection training sequence is partitioned into consecutive, non-overlapping reference segments: full segments contain exactly 4096 time steps, followed by one optional remainder segment. A remainder of length $L_s<4096$ is masked and right-padded to $G_s=\max(512,256\lceil L_s/256\rceil)$, and its generated output is cropped back to $L_s$. We generate every segment once and concatenate only generated segments in their original order. The resulting sequence has the same length and index grid as its real counterpart; no real values are inserted. The anomaly label at each real index is copied to the generated sequence at the same index.}

\revisionblue{Stride-1 windows are extracted from the complete real and generated sequences, including across joins between adjacent generated segments. Each generated window uses the same start and end indices, and therefore the same label slice, as its real counterpart. Under \emph{Real + MarkovTS}, we first sample the real-window indices for a batch and then uniformly choose exactly half of those indices for replacement by their generated counterparts. The batch size, number of updates per epoch, maximum training budget, and model-selection rules match \emph{Real only}. Generated samples are used only for training; both settings reuse the same real validation indices and real test data. Table~\ref{tab:anomaly-training-config} lists their shared detector configurations.}

\revisionblue{The six metrics follow Appendix~\ref{app:evaluation-metrics}. Series-level scores are averaged within each dataset, followed by an equal-weight average over the 26 datasets.}

\begin{table}[t]
\small
\centering
\caption{Anomaly-detection training configurations. CATCH uses separate
optimizers for its main and mask networks.}
\label{tab:anomaly-training-config}
\small
\setlength{\tabcolsep}{5pt}
\renewcommand{\arraystretch}{0.75}
\begin{tabular}{lr lrr}
\toprule
Model & Window & Optimizer / learning rate & Batch size & Maximum epochs \\
\midrule
DLinear & 100 & Adam / $10^{-4}$ & 64 & 10 \\
TranAD & 100 & Adam / $10^{-4}$ & 64 & 5 \\
CATCH & 96 & Adam / $10^{-4}$, $10^{-5}$ & 64 & 3 \\
ModernTCN & 100 & Adam / $10^{-3}$ & 64 & 1 \\
TimesNet & 100 & Adam / $10^{-4}$ & 32 & 10 \\
\bottomrule
\end{tabular}
\end{table}

\section{Additional Generation Analyses}
\label{app:generation-analysis}

\subsection{Additional generation-quality evaluation}
\label{app:additional-generation-quality}

\revisionblue{We complement the main MMD and flat-KL results with TS2Vec-based Context-Fr\'echet Inception Distance (Context-FID). It compares temporal embedding distributions (Appendix~\ref{app:evaluation-metrics}).}

\begin{table}[!ht]
\small
\caption{TS2Vec-based Context-FID at sequence length 168. Lower is better;
bold marks the better method on each dataset.}
\label{tab:context-fid}
\centering
\small
\setlength{\tabcolsep}{7pt}
\renewcommand{\arraystretch}{0.75}
\begin{tabular*}{\textwidth}{@{\extracolsep{\fill}}lrr}
\toprule
Dataset & ImagenFew & MarkovTS \\
\midrule
Electricity & 0.5226 & \textbf{0.3845} \\
Solar & \textbf{0.1857} & 0.4350 \\
Wind & 2.3173 & \textbf{0.6314} \\
Traffic & \textbf{0.6135} & 0.6471 \\
Taxi & 0.6176 & \textbf{0.4889} \\
Pedestrian & \textbf{0.4465} & 2.2981 \\
Air & 1.0353 & \textbf{0.4145} \\
Temperature & 3.1750 & \textbf{1.4524} \\
Rain & 3.7823 & \textbf{1.0151} \\
NN5 & 2.2606 & \textbf{1.4191} \\
Fred-MD & 1.2467 & \textbf{1.2070} \\
Exchange & 8.8181 & \textbf{3.3709} \\
\midrule
Wins (all 12 datasets) & 3 & \textbf{9} \\
\bottomrule
\end{tabular*}
\end{table}

\revisionblue{MarkovTS achieves lower Context-FID on 9 of 12 datasets, while ImagenFew leads on Solar, Traffic, and Pedestrian. These complement Table~\ref{tab:generation-l168}'s window-level and pooled-value results.}

\subsubsection{Fixed-noise reference-swap analysis}
\label{app:reference-swap}

MMD and flat-KL compare distributions aggregated over references, so they cannot determine whether an output follows its selected reference. We therefore pair every generation with a fixed-noise intervention. For reference $i$, frozen deterministic state--transition condition $C_i$, and Gaussian source $z$, the correct output is $Y_i=G(z,C_i)$. A seed-specific derangement selects $j\ne i$, and the swapped output is $Y_{i\leftarrow j}=G(z,C_j)$. The pair shares $z$, model parameters, and reference $i$'s affine restoration; only the complete multiscale state--transition condition changes. This controls noise and prevents affine scale from revealing the swapped reference.

The experiment uses the main flow checkpoint, 16 Euler steps, and seeds 2021--2025. Per dataset and seed, 100 length-512 training references each produce 50 correct/swapped pairs. Each output is evaluated through its three non-overlapping length-168 windows, with the final eight points ignored. GraphFormer and probability sampling remain disabled.

For shape retrieval, the dataset-specific 320-dimensional TS2Vec encoders used in the ImagenFew comparison remain frozen. Each length-168 reference and output window is independently z-standardized before TS2Vec embedding. For each representation or structural statistic, Table~\ref{tab:reference-swap} reports the swapped-minus-correct distance to original reference $i$: cosine distance between TS2Vec embeddings, ACF RMS over lags 1--40, and Hellinger distance between normalized non-DC periodograms. Binary retrieval chooses the active condition from references $i$ and $j$; the 100-way task searches among the 100 reference windows at the same temporal offset within their length-512 references, targeting $i$ for the correct output and $j$ after swapping.

All 12 datasets have positive distance changes under all three measures. Across dataset means, the changes are 0.0506 for Shape-TS2Vec, 0.1616 for ACF, and 0.1994 for PSD. Mean binary retrieval is 93.67\%, compared with 50\% chance. Correct and swapped 100-way retrieval are 63.36\% and 63.09\%, respectively, compared with 1\% chance. Their near equality shows that the swap preserves approximately the same amount of identifiable condition information while redirecting it to a different reference. Correct and swapped marginal MMD remain similar because derangement preserves the empirical condition pool; these small differences are treated as a sanity check rather than conditional-fidelity evidence. Because the complete state--transition condition is swapped jointly, the intervention establishes sensitivity to the joint condition but does not separate the semantic roles of states and transitions.

\begin{table*}[t]
\centering
\caption{Fixed-noise reference-swap analysis. Distance columns are swapped-minus-correct distances to the original reference; positive values indicate that changing only the condition moves the output away from its original owner. Retrieval targets the active condition. Values are five-seed mean $\pm$ sample standard deviation; chance is 50\% for binary and 1\% for 100-way retrieval.}
\label{tab:reference-swap}
\fontsize{7.5}{9}\selectfont
\setlength{\tabcolsep}{3.5pt}
\renewcommand{\arraystretch}{0.92}
\begin{tabular*}{\textwidth}{@{\extracolsep{\fill}}lccccc}
\toprule
Dataset & $\Delta$ Shape-TS2Vec & $\Delta$ ACF & $\Delta$ PSD & Binary (\%) & 100-way correct/swapped (\%) \\
\midrule
Electricity & $0.0157\pm0.0017$ & $0.2299\pm0.0195$ & $0.2500\pm0.0243$ & $95.5\pm1.1$ & $58.9\pm3.8$ / $59.2\pm4.3$ \\
Solar & $0.0165\pm0.0025$ & $0.0852\pm0.0120$ & $0.1218\pm0.0102$ & $95.7\pm1.1$ & $51.6\pm2.7$ / $51.9\pm2.9$ \\
Wind & $0.0934\pm0.0034$ & $0.2341\pm0.0109$ & $0.2424\pm0.0095$ & $93.8\pm0.8$ & $71.2\pm1.6$ / $69.9\pm2.1$ \\
Traffic & $0.0559\pm0.0005$ & $0.1235\pm0.0107$ & $0.1595\pm0.0116$ & $98.8\pm0.3$ & $84.1\pm0.7$ / $84.2\pm1.2$ \\
Taxi & $0.0414\pm0.0016$ & $0.1523\pm0.0063$ & $0.2138\pm0.0058$ & $99.4\pm0.2$ & $89.1\pm0.8$ / $89.4\pm0.9$ \\
Pedestrian & $0.1271\pm0.0167$ & $0.1774\pm0.0073$ & $0.2178\pm0.0066$ & $92.0\pm1.4$ & $45.8\pm1.1$ / $46.7\pm1.7$ \\
Air & $0.0735\pm0.0045$ & $0.2368\pm0.0128$ & $0.2995\pm0.0087$ & $95.8\pm1.3$ & $86.6\pm0.8$ / $84.0\pm2.6$ \\
Temperature & $0.0364\pm0.0013$ & $0.0780\pm0.0022$ & $0.1482\pm0.0029$ & $92.3\pm0.4$ & $45.1\pm2.9$ / $45.0\pm2.8$ \\
Rain & $0.0404\pm0.0037$ & $0.0555\pm0.0012$ & $0.2198\pm0.0028$ & $96.7\pm0.6$ & $78.3\pm1.1$ / $78.0\pm0.2$ \\
NN5 & $0.0180\pm0.0011$ & $0.0929\pm0.0062$ & $0.1133\pm0.0038$ & $75.9\pm0.8$ & $26.0\pm0.2$ / $25.6\pm0.7$ \\
Fred-MD & $0.0324\pm0.0017$ & $0.2325\pm0.0205$ & $0.1707\pm0.0094$ & $92.2\pm0.3$ & $46.3\pm0.6$ / $46.5\pm0.6$ \\
Exchange & $0.0569\pm0.0027$ & $0.2417\pm0.0090$ & $0.2361\pm0.0109$ & $96.0\pm0.8$ & $77.3\pm1.5$ / $76.6\pm1.5$ \\
\midrule
Mean & 0.0506 & 0.1616 & 0.1994 & 93.67 & 63.36 / 63.09 \\
\bottomrule
\end{tabular*}
\end{table*}

\paragraph{Reference replication and perturbation.}
We compare generation with reference replication and Gaussian perturbation, neither requiring an encoder or generator. For each seed in 2021--2025, 100 length-512 references are sampled with replacement from the training pool. Replication uses the complete reference as a length-512 control output, whereas perturbation adds $\sigma\,\mathrm{std}(X_{\mathrm{ref}}^{512})\epsilon$, where $\epsilon\sim\mathcal{N}(0,I)$ and $\sigma\in\{0.05,0.1,0.2,0.5,1\}$; the same noise realization is reused across strengths within each seed. As in the main protocol, every control output is split into three non-overlapping length-168 windows, its final eight points are discarded, and outputs are repeated or sampled until the first 50 windows per reference can be retained in output-then-temporal order. Thus, each control also contributes 5,000 length-168 evaluation windows per dataset and seed. Table~\ref{tab:reference-replication} shows that MarkovTS improves over replication under both metrics on Solar, Wind, Pedestrian, Temperature, Rain, Fred-MD, and Exchange. The gains are not universal: replication retains lower MMD on Traffic, Taxi, and Air, and lower flat-KL on NN5. Among the simple controls, replication gives the lowest Traffic MMD, whereas Gaussian perturbation with $\sigma=0.1$ gives the lowest NN5 flat-KL.

Gaussian perturbation can improve distributional matching, but its effects vary across datasets and metrics. MarkovTS retains lower MMD than every tested noise strength on Solar, Wind, Traffic, Pedestrian, Rain, NN5, Fred-MD, and Exchange, whereas noise achieves lower MMD on Taxi, Air, and Temperature. Noise achieves lower flat-KL on Traffic, Air, Temperature, NN5, and Exchange. Together, these controls show that MarkovTS can improve test-distribution matching beyond reference replication and independent Gaussian perturbation on many, but not all, datasets and metrics.

\begin{table}[!htbp]
\centering
\caption{Generation versus reference replication and Gaussian perturbation. Values are five-seed mean $\pm$ standard deviation; lower is better. Bold marks the best means at displayed precision.}
\label{tab:reference-replication}
\scriptsize
\setlength{\tabcolsep}{2pt}
\renewcommand{\arraystretch}{0.9}
\begin{tabular*}{\textwidth}{@{\extracolsep{\fill}}lccccccc}
\toprule
Dataset & Replication & $\sigma=0.05$ & $\sigma=0.1$ & $\sigma=0.2$ & $\sigma=0.5$ & $\sigma=1$ & MarkovTS \\
\midrule
\multicolumn{8}{c}{MMD$\downarrow$} \\
\midrule
Electricity & \textbf{0.001}$\pm$0.001 & \textbf{0.001}$\pm$0.001 & \textbf{0.001}$\pm$0.002 & \textbf{0.001}$\pm$0.001 & \textbf{0.001}$\pm$0.002 & 0.002$\pm$0.002 & \textbf{0.001}$\pm$0.001 \\
Solar & 0.035$\pm$0.005 & 0.037$\pm$0.005 & 0.038$\pm$0.006 & 0.040$\pm$0.006 & 0.049$\pm$0.009 & 0.193$\pm$0.011 & \textbf{0.033}$\pm$0.003 \\
Wind & 0.026$\pm$0.015 & 0.028$\pm$0.020 & 0.024$\pm$0.018 & 0.023$\pm$0.016 & 0.022$\pm$0.007 & 0.021$\pm$0.005 & \textbf{0.013}$\pm$0.009 \\
Traffic & \textbf{0.070}$\pm$0.006 & 0.092$\pm$0.003 & 0.092$\pm$0.003 & 0.092$\pm$0.003 & 0.098$\pm$0.004 & 0.165$\pm$0.009 & 0.087$\pm$0.002 \\
Taxi & 0.094$\pm$0.011 & 0.099$\pm$0.000 & 0.099$\pm$0.000 & 0.098$\pm$0.000 & \textbf{0.093}$\pm$0.000 & 0.106$\pm$0.000 & 0.110$\pm$0.008 \\
Pedestrian & 0.050$\pm$0.016 & 0.053$\pm$0.014 & 0.053$\pm$0.017 & 0.057$\pm$0.024 & 0.060$\pm$0.021 & 0.061$\pm$0.024 & \textbf{0.032}$\pm$0.006 \\
Air & \textbf{0.012}$\pm$0.003 & \textbf{0.012}$\pm$0.004 & 0.013$\pm$0.002 & 0.014$\pm$0.002 & 0.013$\pm$0.002 & 0.017$\pm$0.002 & 0.015$\pm$0.002 \\
Temperature & 0.222$\pm$0.025 & 0.146$\pm$0.000 & 0.146$\pm$0.000 & 0.143$\pm$0.000 & \textbf{0.128}$\pm$0.000 & 0.134$\pm$0.000 & 0.211$\pm$0.002 \\
Rain & 0.033$\pm$0.003 & 0.035$\pm$0.003 & 0.035$\pm$0.003 & 0.034$\pm$0.003 & 0.032$\pm$0.003 & 0.039$\pm$0.003 & \textbf{0.031}$\pm$0.002 \\
NN5 & 0.153$\pm$0.004 & 0.156$\pm$0.005 & 0.167$\pm$0.005 & 0.179$\pm$0.007 & 0.201$\pm$0.008 & 0.193$\pm$0.007 & \textbf{0.145}$\pm$0.000 \\
Fred-MD & 0.003$\pm$0.002 & 0.003$\pm$0.002 & 0.003$\pm$0.002 & 0.003$\pm$0.002 & 0.003$\pm$0.001 & 0.003$\pm$0.001 & \textbf{0.001}$\pm$0.000 \\
Exchange & 0.132$\pm$0.020 & 0.127$\pm$0.015 & 0.137$\pm$0.024 & 0.142$\pm$0.025 & 0.145$\pm$0.028 & 0.149$\pm$0.024 & \textbf{0.124}$\pm$0.008 \\
\midrule
\multicolumn{8}{c}{flat-KL$\downarrow$} \\
\midrule
Electricity & 0.106$\pm$0.048 & 0.110$\pm$0.000 & 0.102$\pm$0.000 & 0.100$\pm$0.000 & 0.098$\pm$0.000 & 0.084$\pm$0.000 & \textbf{0.001}$\pm$0.000 \\
Solar & 0.037$\pm$0.037 & 0.073$\pm$0.000 & 0.105$\pm$0.000 & 0.260$\pm$0.001 & 0.595$\pm$0.002 & 0.863$\pm$0.002 & \textbf{0.009}$\pm$0.002 \\
Wind & 0.223$\pm$0.063 & 0.224$\pm$0.065 & 0.219$\pm$0.060 & 0.201$\pm$0.047 & 0.177$\pm$0.024 & 0.151$\pm$0.008 & \textbf{0.111}$\pm$0.018 \\
Traffic & 0.012$\pm$0.003 & 0.017$\pm$0.002 & 0.007$\pm$0.002 & \textbf{0.004}$\pm$0.002 & 0.017$\pm$0.005 & 0.065$\pm$0.008 & 0.007$\pm$0.002 \\
Taxi & 0.030$\pm$0.025 & 0.021$\pm$0.000 & 0.018$\pm$0.001 & 0.018$\pm$0.000 & 0.025$\pm$0.000 & 0.049$\pm$0.000 & \textbf{0.010}$\pm$0.004 \\
Pedestrian & 0.020$\pm$0.008 & 0.022$\pm$0.008 & 0.027$\pm$0.009 & 0.032$\pm$0.012 & 0.079$\pm$0.017 & 0.153$\pm$0.075 & \textbf{0.007}$\pm$0.001 \\
Air & 0.029$\pm$0.005 & 0.029$\pm$0.005 & 0.027$\pm$0.004 & 0.029$\pm$0.005 & \textbf{0.016}$\pm$0.003 & 0.037$\pm$0.008 & 0.026$\pm$0.011 \\
Temperature & 0.164$\pm$0.022 & 0.170$\pm$0.002 & 0.158$\pm$0.001 & 0.141$\pm$0.000 & 0.090$\pm$0.000 & \textbf{0.071}$\pm$0.000 & 0.137$\pm$0.041 \\
Rain & 0.007$\pm$0.002 & 0.007$\pm$0.003 & 0.008$\pm$0.004 & 0.007$\pm$0.003 & 0.011$\pm$0.006 & 0.030$\pm$0.008 & \textbf{0.002}$\pm$0.001 \\
NN5 & 0.034$\pm$0.011 & 0.035$\pm$0.013 & \textbf{0.033}$\pm$0.011 & 0.041$\pm$0.014 & 0.047$\pm$0.016 & 0.038$\pm$0.013 & 0.045$\pm$0.003 \\
Fred-MD & 0.517$\pm$0.259 & 0.346$\pm$0.142 & 0.340$\pm$0.135 & 0.308$\pm$0.124 & 0.243$\pm$0.008 & 0.187$\pm$0.004 & \textbf{0.176}$\pm$0.001 \\
Exchange & 1.934$\pm$0.372 & \textbf{1.766}$\pm$0.028 & 2.168$\pm$0.122 & 2.490$\pm$0.142 & 2.957$\pm$0.238 & 3.454$\pm$0.393 & 1.794$\pm$0.137 \\
\bottomrule
\end{tabular*}
\end{table}

\subsection{State--transition ablations}
\label{app:code-randomization}

We examine code-assignment interventions and a single-codebook patch-based alternative. Uniform randomization and probability sampling modify the state stream, transition stream, or both while keeping the pretrained encoder, codebooks, and flow generator frozen. Patch-Flow instead uses a separately trained patch encoder with a single codebook. Table~\ref{tab:code-randomization-full} summarizes the results.

Joint randomization produces higher mean MMD and flat-KL than either single-stream intervention on most datasets, including Solar, Traffic, and NN5, supporting both reference-derived streams. Effects vary: joint randomization does not yield the highest MMD on Rain or the highest flat-KL on Wind. The three probability-sampling variants yield similar distributional scores at the selected temperature. Under this paired intervention protocol, the deterministic MarkovTS control has lower MMD and flat-KL than Patch-Flow on eleven of twelve datasets.

Both code-assignment experiments use seeds 2021--2025 on all 12 datasets. Per dataset and seed, 100 length-512 references each condition 50 Gaussian sources, yielding 5,000 sequences per variant; one sampled code sequence is shared across each reference's 50 sources. Generation uses 16 Euler steps, restores reference normalization statistics, and extracts the length-168 crop $[168,336)$ for evaluation against the target test split. Within each experiment and seed, variants share references, Gaussian sources, checkpoints, data files, and evaluation controls. The deterministic control in this paired experiment is therefore not the three-window main estimate in Table~\ref{tab:generation-l168}. Seeds vary reference selection, Gaussian sources, and code sampling; standard deviations measure cross-seed variation.

\subsubsection{Uniform code randomization}

Random states replaces every valid state index with an independent uniform draw from all 1,024 entries, retaining reference-derived transition indices; Random transitions does the converse, and Random both replaces both streams. Draws may coincide with original indices; invalid positions and codebook vectors remain unchanged. Random both shares its randomized state and transition streams with the respective single-stream variants. All interventions follow deterministic encoding, without recomputing transitions after state replacement.

\subsubsection{Reference-dependent code sampling}
\label{app:full-codebook-sampling}

Probability sampling draws from all 1,024 entries using the frozen query's cosine energies $E_k$:
\begin{equation}
p(k\mid q)=\frac{\exp(-E_k(q)/T)}
{\sum_{\ell=1}^{1024}\exp(-E_\ell(q)/T)},\qquad T=0.005.
\label{eq:app-full-codebook-probability}
\end{equation}
Stochastic states, Stochastic transitions, and Stochastic both sample every valid position in the corresponding streams, without top-$k$ truncation. Unsampled streams use nearest-code assignment. State assignments are obtained first; their selected vectors are then used to recompute transition queries and energies. Thus, Stochastic states can change transition indices even when transitions use nearest-code assignment, preserving the encoder's state-dependent transition construction rather than fixing the other stream as in uniform randomization.

\subsubsection{Single-codebook patch representation}

Patch-Flow uses a standard patch Transformer with a single codebook to condition a flow-matching generator. The patch representation is trained using codebook optimization and reconstruction losses, with the remaining model and training configurations matching MarkovTS.

\begin{table*}[t]
\centering
\caption{Generation quality under code interventions and patch-based comparison. Values are five-seed mean $\pm$ sample standard deviation. Lower is better.}
\label{tab:code-randomization-full}
\label{tab:full-codebook-sampling}
\scriptsize
\setlength{\tabcolsep}{1.2pt}
\renewcommand{\arraystretch}{1.12}
\begin{tabular}{@{}p{0.08\textwidth}p{0.06\textwidth}*{8}{>{\centering\arraybackslash}p{\dimexpr(\textwidth-0.165\textwidth-12\tabcolsep)/8\relax}}@{}}
\toprule
 & & & \multicolumn{3}{c}{Random} & \multicolumn{3}{c}{Stochastic} & \\
\cmidrule(lr){4-6}\cmidrule(lr){7-9}
Dataset & Metric & MarkovTS & States & Transitions & Both & States & Transitions & Both & \shortstack{Patch-\\Flow}\\
\midrule
Elec. & MMD & 0.001$\pm$0.000 & 0.002$\pm$0.002 & 0.002$\pm$0.002 & 0.003$\pm$0.002 & 0.001$\pm$0.001 & 0.002$\pm$0.002 & 0.002$\pm$0.002 & 0.003$\pm$0.002\\
 & flat-KL & 0.001$\pm$0.000 & 0.059$\pm$0.044 & 0.061$\pm$0.041 & 0.068$\pm$0.039 & 0.005$\pm$0.003 & 0.012$\pm$0.008 & 0.015$\pm$0.013 & 0.054$\pm$0.031\\
\addlinespace[2pt]
Solar & MMD & 0.032$\pm$0.003 & 0.236$\pm$0.008 & 0.190$\pm$0.011 & 0.516$\pm$0.008 & 0.034$\pm$0.004 & 0.034$\pm$0.003 & 0.034$\pm$0.003 & 0.045$\pm$0.006 \\
 & flat-KL & 0.009$\pm$0.000 & 0.689$\pm$0.036 & 0.842$\pm$0.027 & 1.099$\pm$0.046 & 0.015$\pm$0.008 & 0.014$\pm$0.007 & 0.020$\pm$0.010 & 0.173$\pm$0.018 \\
\addlinespace[2pt]
Wind & MMD & 0.019$\pm$0.001 & 0.023$\pm$0.026 & 0.023$\pm$0.025 & 0.023$\pm$0.026 & 0.023$\pm$0.026 & 0.023$\pm$0.026 & 0.023$\pm$0.026 & 0.033$\pm$0.030 \\
 & flat-KL & 0.147$\pm$0.006 & 0.189$\pm$0.061 & 0.165$\pm$0.051 & 0.178$\pm$0.056 & 0.177$\pm$0.059 & 0.177$\pm$0.057 & 0.177$\pm$0.057 & 0.203$\pm$0.068\\
\addlinespace[2pt]
Traffic & MMD & 0.087$\pm$0.002 & 0.168$\pm$0.010 & 0.130$\pm$0.008 & 0.302$\pm$0.009 & 0.080$\pm$0.007 & 0.080$\pm$0.007 & 0.080$\pm$0.007 & 0.084$\pm$0.003 \\
 & flat-KL & 0.003$\pm$0.000 & 0.144$\pm$0.016 & 0.131$\pm$0.040 & 0.247$\pm$0.085 & 0.007$\pm$0.004 & 0.007$\pm$0.004 & 0.007$\pm$0.004 & 0.003$\pm$0.001\\
\addlinespace[2pt]
Taxi & MMD & 0.101$\pm$0.001 & 0.118$\pm$0.007 & 0.119$\pm$0.007 & 0.148$\pm$0.007 & 0.091$\pm$0.011 & 0.091$\pm$0.012 & 0.091$\pm$0.011 & 0.105$\pm$0.003 \\
 & flat-KL & 0.002$\pm$0.000 & 0.075$\pm$0.029 & 0.146$\pm$0.055 & 0.184$\pm$0.053 & 0.016$\pm$0.011 & 0.017$\pm$0.012 & 0.016$\pm$0.011 & 0.012$\pm$0.004 \\
\addlinespace[2pt]
Ped. & MMD & 0.032$\pm$0.001 & 0.049$\pm$0.003 & 0.054$\pm$0.006 & 0.082$\pm$0.003 & 0.034$\pm$0.003 & 0.035$\pm$0.004 & 0.037$\pm$0.010 & 0.057$\pm$0.007 \\
 & flat-KL & 0.007$\pm$0.001 & 0.088$\pm$0.017 & 0.148$\pm$0.017 & 0.192$\pm$0.011 & 0.010$\pm$0.004 & 0.015$\pm$0.006 & 0.020$\pm$0.007 & 0.163$\pm$0.015 \\
\addlinespace[2pt]
Air & MMD & 0.009$\pm$0.001 & 0.016$\pm$0.002 & 0.017$\pm$0.004 & 0.022$\pm$0.005 & 0.013$\pm$0.002 & 0.013$\pm$0.002 & 0.013$\pm$0.002 & 0.011$\pm$0.002 \\
 & flat-KL & 0.019$\pm$0.002 & 0.039$\pm$0.020 & 0.046$\pm$0.040 & 0.057$\pm$0.054 & 0.030$\pm$0.013 & 0.030$\pm$0.013 & 0.030$\pm$0.013 & 0.021$\pm$0.002 \\
\addlinespace[2pt]
Temp. & MMD & 0.142$\pm$0.001 & 0.134$\pm$0.006 & 0.152$\pm$0.010 & 0.213$\pm$0.018 & 0.134$\pm$0.007 & 0.134$\pm$0.007 & 0.134$\pm$0.007 & 0.204$\pm$0.015 \\
 & flat-KL & 0.161$\pm$0.004 & 0.134$\pm$0.024 & 0.109$\pm$0.021 & 0.253$\pm$0.045 & 0.098$\pm$0.016 & 0.098$\pm$0.017 & 0.098$\pm$0.018 & 0.224$\pm$0.039 \\
\addlinespace[2pt]
Rain & MMD & 0.031$\pm$0.001 & 0.046$\pm$0.002 & 0.053$\pm$0.002 & 0.051$\pm$0.004 & 0.036$\pm$0.002 & 0.036$\pm$0.002 & 0.036$\pm$0.002 & 0.040$\pm$0.003 \\
 & flat-KL & 0.007$\pm$0.000 & 0.010$\pm$0.002 & 0.015$\pm$0.004 & 0.050$\pm$0.031 & 0.004$\pm$0.003 & 0.004$\pm$0.003 & 0.004$\pm$0.003 & 0.008$\pm$0.002 \\
\addlinespace[2pt]
NN5 & MMD & 0.180$\pm$0.002 & 0.205$\pm$0.008 & 0.231$\pm$0.007 & 0.292$\pm$0.010 & 0.166$\pm$0.024 & 0.186$\pm$0.007 & 0.186$\pm$0.007 & 0.236$\pm$0.007 \\
 & flat-KL & 0.043$\pm$0.001 & 0.106$\pm$0.036 & 0.193$\pm$0.050 & 0.301$\pm$0.084 & 0.047$\pm$0.004 & 0.051$\pm$0.013 & 0.051$\pm$0.013 & 0.215$\pm$0.043 \\
\addlinespace[2pt]
Fred-MD & MMD & 0.002$\pm$0.000 & 0.003$\pm$0.001 & 0.002$\pm$0.001 & 0.002$\pm$0.001 & 0.003$\pm$0.001 & 0.003$\pm$0.001 & 0.003$\pm$0.001 & 0.004$\pm$0.002 \\
 & flat-KL & 0.244$\pm$0.006 & 0.364$\pm$0.258 & 0.200$\pm$0.026 & 0.236$\pm$0.025 & 0.256$\pm$0.018 & 0.291$\pm$0.106 & 0.288$\pm$0.105 & 0.369$\pm$0.139 \\
\addlinespace[2pt]
Exchange & MMD & 0.124$\pm$0.003 & 0.148$\pm$0.020 & 0.147$\pm$0.021 & 0.148$\pm$0.022 & 0.138$\pm$0.012 & 0.142$\pm$0.010 & 0.143$\pm$0.017 & 0.142$\pm$0.012 \\
 & flat-KL & 1.631$\pm$0.026 & 5.030$\pm$2.159 & 3.068$\pm$0.668 & 3.351$\pm$0.374 & 1.760$\pm$0.108 & 1.842$\pm$0.189 & 1.883$\pm$0.203 & 2.013$\pm$0.091 \\
\bottomrule
\end{tabular}
\end{table*}

\subsection{Generation across sequence lengths}
\label{app:joint-length}

\revisionblue{The length analysis uses the same pretrained MarkovTS encoder, codebooks, flow model, and metric implementations, with $N_{\mathrm{FE}}=16$ Euler steps. For each dataset and seed, $K_{\mathrm{eval}}=100$ references each contribute 50 retained evaluation windows, giving $N=5000$ windows in total. We use seeds 2021--2025.}

\revisionblue{Reference and generation lengths match, $L_{\mathrm{gen}}=L_{\mathrm{ref}}\in\{512,1024,2048\}$. Each output yields 3, 6, or 12 fixed, non-overlapping $L_{\mathrm{eval}}=168$ windows, discarding the final 8, 16, or 32 points, respectively. For each reference, we generate outputs until at least 50 windows are available, order candidates first by Gaussian draw and then by temporal position, and retain the first 50. The comparison therefore measures length-168 window fidelity under references and outputs of different lengths, not full-sequence fidelity at lengths 512, 1024, or 2048.}

\revisionblue{Fred-MD, NN5, Rain, and Temperature contain 504 released training points per channel. Lengths 512, 1024, and 2048 therefore require 8, 520, and 1544 reflection-padded points, respectively.}

\revisionblue{Table~\ref{tab:length-complete} compares reference and generation lengths 512, 1024, and 2048 over five seeds; $\dagger$ marks datasets requiring reflection padding.}
\begin{table}[!htbp]
\small
\caption{Generation quality across reference and output lengths. Lower is better; values are five-seed mean $\pm$ sample standard deviation. All outputs are evaluated through fixed, non-overlapping length-168 windows.}
\label{tab:length-complete}
\centering
\fontsize{8.1}{9.9}\selectfont
\setlength{\tabcolsep}{4pt}
\renewcommand{\arraystretch}{0.824}
\begin{tabular*}{0.92\textwidth}{@{\extracolsep{\fill}}clccc}
\toprule
Metric & Dataset & 512 & 1024 & 2048 \\
\midrule
\multirow{12}{*}{\rotatebox[origin=c]{90}{MMD}}
& Electricity & $0.001\pm0.001$ & $0.002\pm0.001$ & $0.002\pm0.001$ \\
& Solar & $0.033\pm0.003$ & $0.038\pm0.005$ & $0.036\pm0.007$ \\
& Wind & $0.013\pm0.009$ & $0.017\pm0.013$ & $0.021\pm0.017$ \\
& Traffic & $0.087\pm0.002$ & $0.222\pm0.022$ & $0.194\pm0.011$ \\
& Taxi & $0.110\pm0.008$ & $0.102\pm0.003$ & $0.098\pm0.001$ \\
& Pedestrian & $0.032\pm0.006$ & $0.037\pm0.005$ & $0.036\pm0.006$ \\
& Air & $0.015\pm0.002$ & $0.014\pm0.003$ & $0.012\pm0.004$ \\
& Temperature$^{\dagger}$ & $0.211\pm0.002$ & $0.183\pm0.007$ & $0.202\pm0.015$ \\
& Rain$^{\dagger}$ & $0.031\pm0.002$ & $0.029\pm0.002$ & $0.027\pm0.002$ \\
& NN5$^{\dagger}$ & $0.145\pm0.000$ & $0.168\pm0.001$ & $0.215\pm0.000$ \\
& Fred-MD$^{\dagger}$ & $0.001\pm0.000$ & $0.001\pm0.000$ & $0.001\pm0.000$ \\
& Exchange & $0.124\pm0.008$ & $0.130\pm0.008$ & $0.135\pm0.004$ \\
\midrule
\multirow{12}{*}{\rotatebox[origin=c]{90}{flat-KL}}
& Electricity & $0.001\pm0.000$ & $0.064\pm0.044$ & $0.061\pm0.046$ \\
& Solar & $0.009\pm0.002$ & $0.012\pm0.003$ & $0.013\pm0.002$ \\
& Wind & $0.111\pm0.018$ & $0.103\pm0.035$ & $0.114\pm0.055$ \\
& Traffic & $0.007\pm0.002$ & $0.007\pm0.002$ & $0.005\pm0.002$ \\
& Taxi & $0.010\pm0.004$ & $0.009\pm0.005$ & $0.007\pm0.003$ \\
& Pedestrian & $0.007\pm0.001$ & $0.018\pm0.011$ & $0.016\pm0.009$ \\
& Air & $0.026\pm0.011$ & $0.022\pm0.004$ & $0.017\pm0.007$ \\
& Temperature$^{\dagger}$ & $0.137\pm0.041$ & $0.139\pm0.041$ & $0.141\pm0.042$ \\
& Rain$^{\dagger}$ & $0.002\pm0.001$ & $0.002\pm0.001$ & $0.002\pm0.001$ \\
& NN5$^{\dagger}$ & $0.045\pm0.003$ & $0.044\pm0.002$ & $0.051\pm0.002$ \\
& Fred-MD$^{\dagger}$ & $0.176\pm0.001$ & $0.175\pm0.002$ & $0.175\pm0.003$ \\
& Exchange & $1.794\pm0.137$ & $1.855\pm0.144$ & $1.786\pm0.054$ \\
\bottomrule
\end{tabular*}
\end{table}

\section{Qualitative and Structural Generation Analyses}
\label{app:generation-visualizations}

We complement generation metrics with analyses of the learned representation and frozen sampler. Representation analyses use real sequences and deterministic nearest-code assignments. Generation examples use exponential-moving-average flow parameters and independent Gaussian sources unless specified otherwise, showing variation under a fixed reference-derived condition.

\subsection{State and transition code structure}
\label{app:state-transition-structure}

\subsubsection{Shared codebook entries}

\begin{figure}[tbp]
  \centering
  \includegraphics[width=0.8\textwidth]{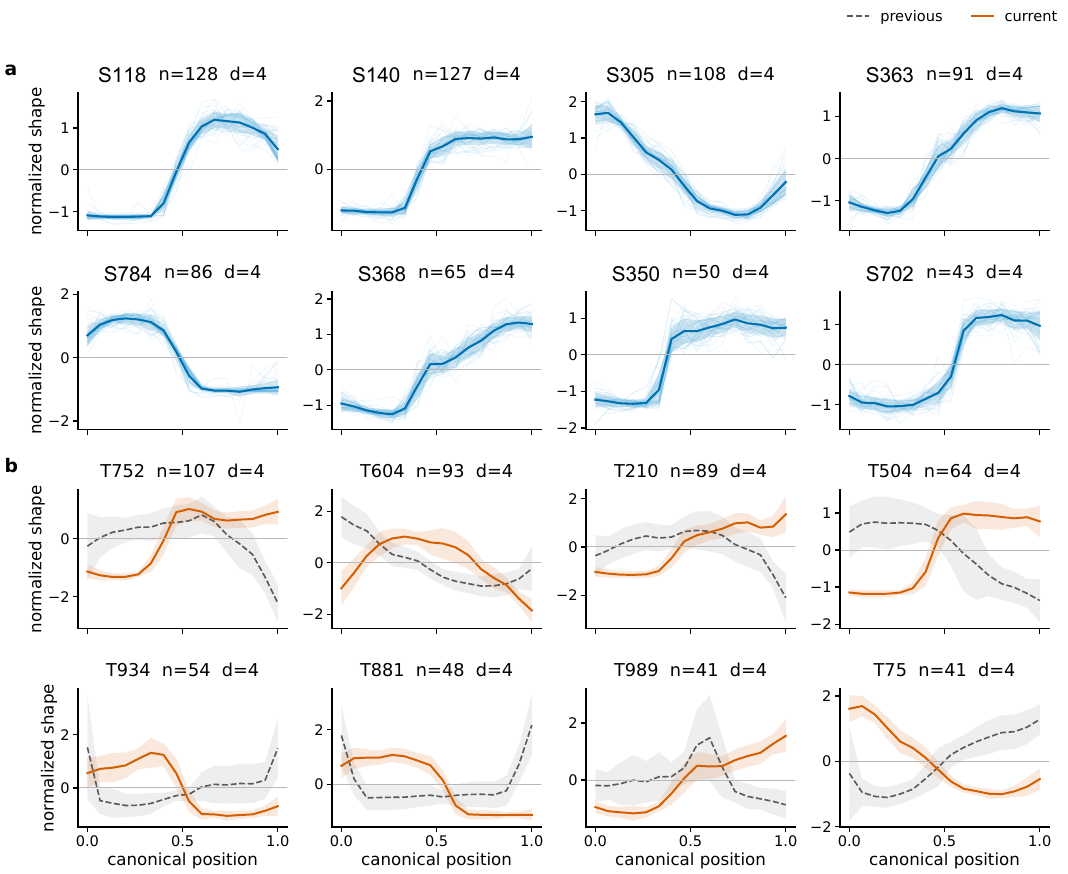}
  \caption{\revisionblue{Representative state and transition codes at temporal span 16.
  \textbf{(a)} Assigned normalized patches and their mean. \textbf{(b)} Mean
  previous/current patches and one-standard-deviation bands. $n$ and $d$
  denote the number of assignments and represented datasets, respectively.}}
  \label{fig:state-transition-semantics}
\end{figure}

\revisionblue{Figure~\ref{fig:state-transition-semantics} illustrates shared local morphology and contextual transitions across held-out datasets. State examples show locally normalized patches assigned to frequent codes; transition examples show previous--current patch pairs sharing a transition code. Deterministic selection prioritizes usage and representation across Electricity, Solar, Traffic, and Exchange. Recurring shapes and ordered changes suggest shared temporal structure across these datasets.}

\subsubsection{State--transition complementarity}
\label{app:transition-complementarity}

Shared morphological states need not specify changes between their continuous patches. Using real length-512 training references and the frozen encoder, we group transition occurrences by dataset $d$, temporal span $w_h$, and ordered state pair. Each reference is standardized once by its mean and population standard deviation, with denominator stabilizer $10^{-8}$. Adjacent intervals are reduced to canonical patches $p,q\in\mathbb R^{P}$, $P=16$, without further patchwise normalization, retaining relative levels and amplitudes.

Let $m(a)$ and $s(a)$ denote the mean and population standard deviation of a canonical patch $a$, and let $u_b=b-(P-1)/2$. The least-squares slope $\beta(a)$ and continuous-relation feature $\phi(p,q)$ are
\begin{equation}
\begin{aligned}
\beta(a)&=\frac{\sum_{b=0}^{P-1}u_ba_b}{\sum_{b=0}^{P-1}u_b^2},\\
\phi(p,q)&=\bigl[m(q)-m(p),\;\log\max(s(q),10^{-6})-\log\max(s(p),10^{-6}),\\
&\hspace{3.5em}\beta(q)-\beta(p),\;q_0-p_{P-1}\bigr]^{\mathsf T}.
\end{aligned}
\label{eq:app-complementarity-features}
\end{equation}
The four coordinates measure level shift, log-amplitude ratio, slope change, and boundary jump. Before group filtering or subsampling, each coordinate is centered and divided by its population standard deviation over all valid transition occurrences in the same dataset--span cell; standard deviations below $10^{-8}$ are replaced by one. Occurrence $j$ has standardized feature $\widetilde\phi_j$.

A state-pair group is eligible when at least two transition codes each occur at least three times. We retain at most 100 eligible groups per dataset--span cell, sampled uniformly without replacement; within each group, we retain up to eight codes ranked by decreasing frequency, breaking ties by increasing code ID, and up to 64 occurrences per code, sampled uniformly without replacement. Sampling uses base seed 20260922. For a retained group $g$, let $I_{g,k}$ contain the retained occurrences of transition code $k$, $n_{g,k}=|I_{g,k}|$, and $\bar\phi_{g,k}$ be their mean standardized feature. Pooled within-code dispersion is
\begin{equation}
W_g^{\mathrm{obs}}=\frac{\sum_k\sum_{j\in I_{g,k}}\|\widetilde\phi_j-\bar\phi_{g,k}\|_2^2}{\sum_k(n_{g,k}-1)}.
\label{eq:app-complementarity-dispersion}
\end{equation}
Thus, each code contributes according to its residual degrees of freedom. Within each group, we permute transition labels 200 times while preserving the occurrence set and every code count. Writing $W_g^{\pi_b}$ for the dispersion under permutation $b$, the observed-to-permuted ratio is
\begin{equation}
R_g=\frac{W_g^{\mathrm{obs}}}{\max\!\left(\frac{1}{200}\sum_{b=1}^{200}W_g^{\pi_b},10^{-12}\right)}.
\label{eq:app-complementarity-ratio}
\end{equation}
Ratios below one indicate greater within-code coherence than count-matched random labels.

Aggregation gives each eligible dataset equal weight. We take median ratios over retained groups $G_{d,h}$ per dataset--span cell, then over datasets with at least five groups:
\begin{equation}
M_{d,h}=\operatorname{median}_{g\in G_{d,h}}R_g,\qquad
M_h=\operatorname{median}_{d:\,|G_{d,h}|\ge5}M_{d,h}.
\label{eq:app-complementarity-aggregation}
\end{equation}
Figure~\ref{fig:transition-complementarity}(c) reports the reduction $100(1-M_h)\%$. At each span, we resample the eligible dataset medians with replacement 5,000 times, retaining the original number of datasets, and recompute the macro-median. The 2.5th and 97.5th percentiles of the resulting reductions define the 95\% interval. Spans $256,128,64,32,16$ include $8,8,10,10,10$ datasets and $547,710,700,772,860$ groups, respectively, totaling 3,589 groups in the macro statistics. The reductions of 18.1\%--29.7\%, with all intervals excluding zero, support transition-code information beyond the morphological state pair.

\begin{figure}[!htbp]
\centering
\includegraphics[page=1,width=0.9\textwidth]{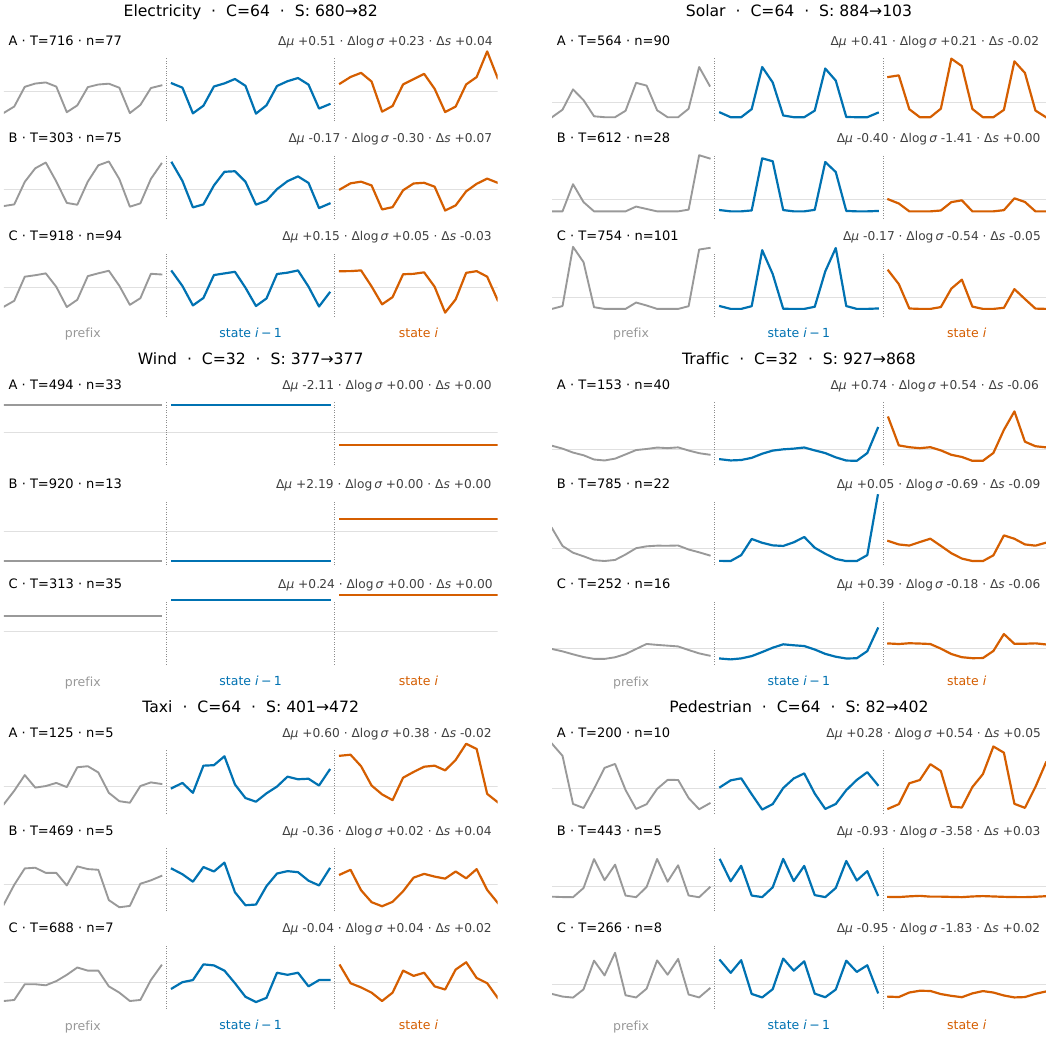}
\caption{Transition-code examples within fixed state pairs on Electricity, Solar, Wind, Traffic, Taxi, and Pedestrian. Each panel shows three real patch-pair medoids. $C$ denotes temporal scale, $S$ the ordered state pair, $T$ the transition code, and $n$ its occurrence count. Annotations report changes in level, log-amplitude, and slope.}
\label{fig:transition-atlas-first}
\end{figure}

\begin{figure}[!htbp]
\centering
\includegraphics[page=2,width=0.9\textwidth]{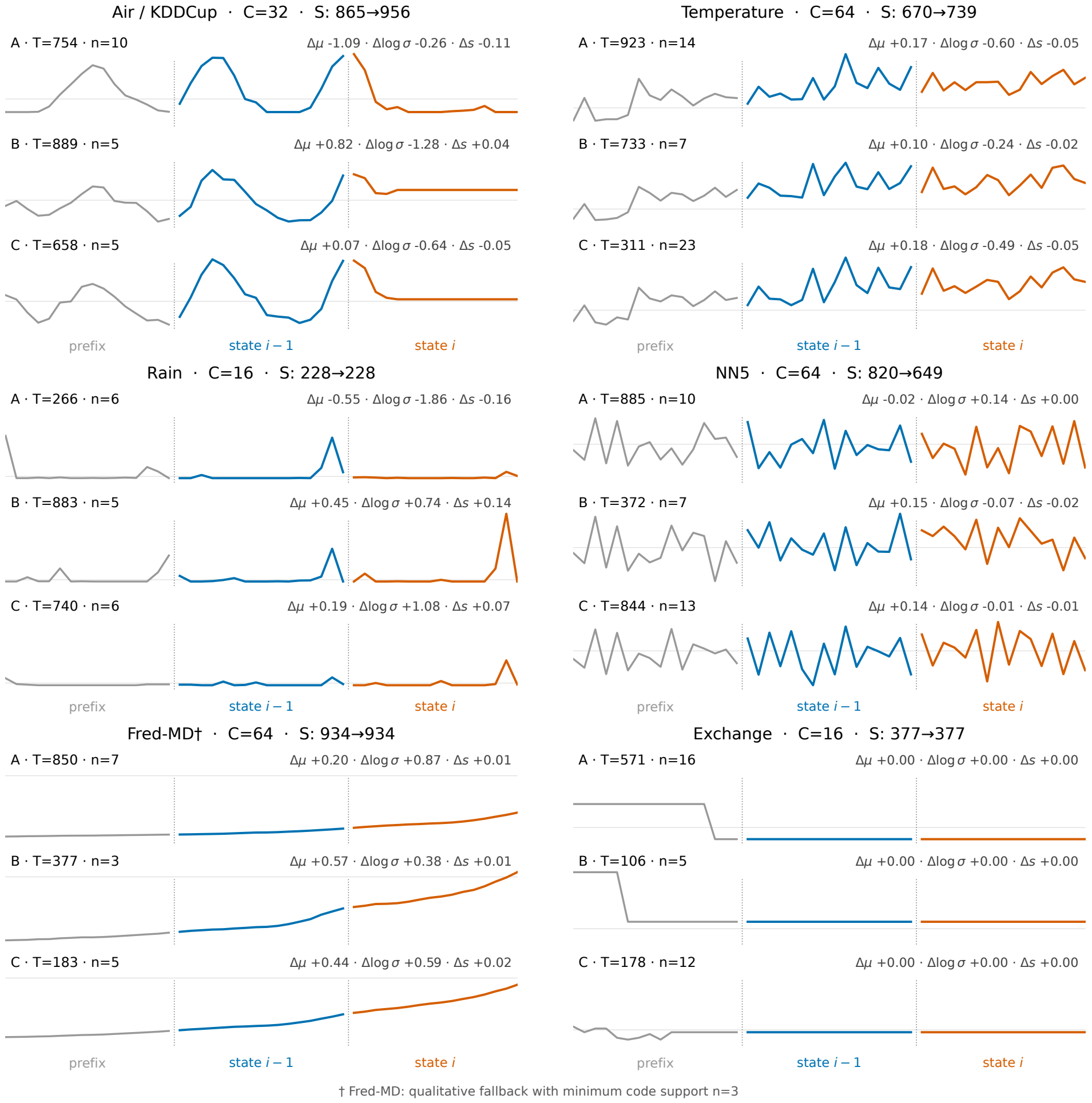}
\caption{Transition-code examples within fixed state pairs on Air, Temperature, Rain, NN5, Fred-MD, and Exchange, using the notation of Figure~\ref{fig:transition-atlas-first}. The dagger marks Fred-MD's lower-support qualitative example.}
\label{fig:transition-atlas-second}
\end{figure}

The examples in Figures~\ref{fig:transition-atlas-first} and~\ref{fig:transition-atlas-second} make this complementarity visible across the 12 datasets. Within a fixed state pair, the Solar examples differ in amplitude change, while the Wind examples distinguish upward and downward level shifts. Each displayed sequence is a feature-space medoid for one transition code, with normalization applied once to the complete length-512 reference so that between-patch level and amplitude differences remain visible. The preceding context is shown in gray, followed by the previous patch in blue and the current patch in orange.

Examples are selected by a support-filtered centroid-diversity rule requiring three transition codes with at least five occurrences each, independently of the aggregate analysis. For Fred-MD, no group meets this threshold; the marked example uses state pair $934\rightarrow934$ at scale 64, with code supports of 7, 3, and 5. This exception covers all datasets qualitatively without altering quantitative results.

\subsection{Multiscale conditioning}

\begin{figure}[!ht]
  \centering
  \includegraphics[width=0.9\textwidth]{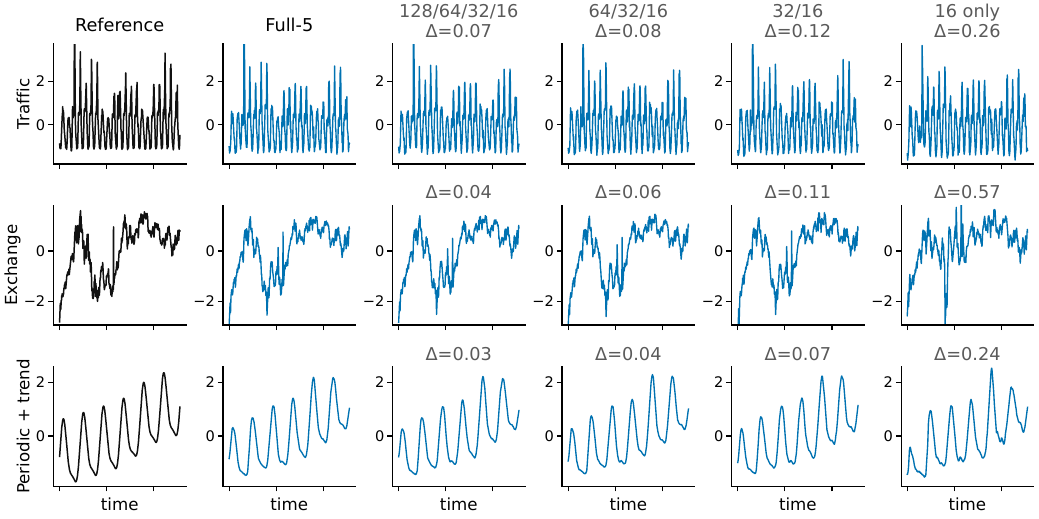}
  \caption{Coarse-to-fine conditioning ablation with a shared Gaussian source.
  Columns progressively remove the coarsest condition; $\Delta$ is the
  standardized RMS distance from the full five-scale output.}
  \label{fig:multiscale-suffix}
\end{figure}

\revisionblue{Figure~\ref{fig:multiscale-suffix} removes coarse conditions while holding the reference, Gaussian source, generator, and integration settings fixed. In the three examples, removing coarse states and transitions alters global organization, while the remaining fine-scale codes retain local waveform detail.}

\subsection{Distributional diagnostics}

\begin{figure}[!ht]
  \centering
  \includegraphics[width=0.72\textwidth]{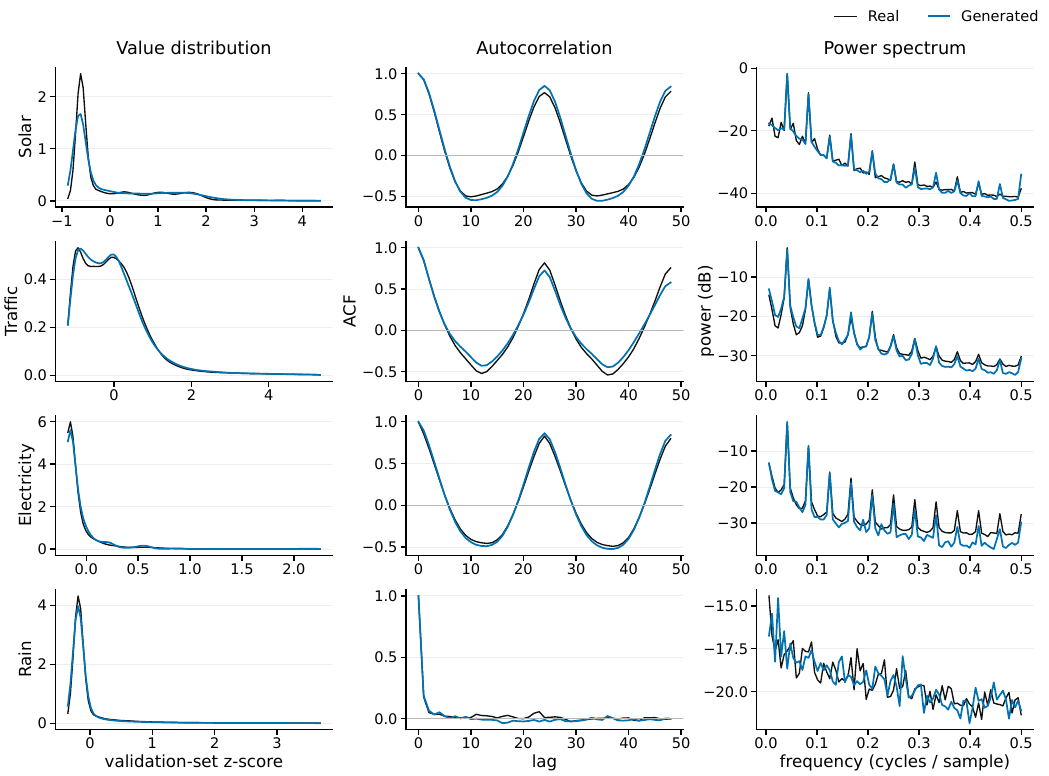}
  \caption{Distributional diagnostics on four held-out datasets. Columns
  compare value density, autocorrelation, and normalized power spectrum for
  real validation data (black) and MarkovTS samples (blue).}
  \label{fig:distribution-diagnostics}
\end{figure}

\revisionblue{Figure~\ref{fig:distribution-diagnostics} compares generated and real validation sequences through marginal value densities, autocorrelation, and power spectra, assessing amplitude statistics, temporal dependence, and periodic content. Using up to 3,000 finite-valued sequences per distribution, these diagnostics complement Table~\ref{tab:generation-l168}.}

\subsection{Conditional flow path}

\begin{figure}[H]
  \centering
  \includegraphics[width=0.8\textwidth]{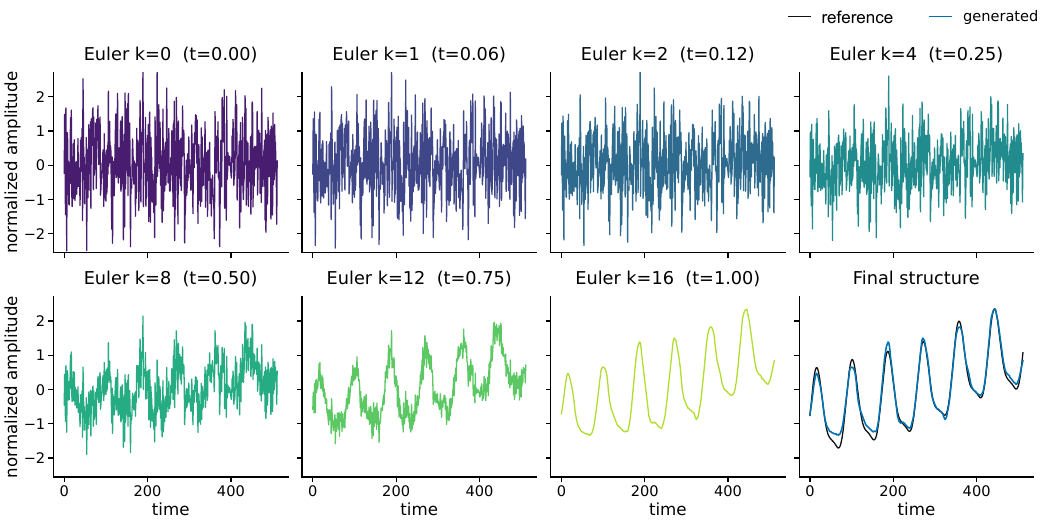}
  \caption{Intermediate states along a conditional flow path. The final
  panel compares the reference with the generated realization.}
  \label{fig:conditional-flow-path}
\end{figure}

\revisionblue{Figure~\ref{fig:conditional-flow-path} shows intermediate Euler states under a fixed state--transition condition. The flow gradually suppresses unstructured Gaussian variation while establishing the reference's periodicity, harmonic content, trend, and smooth level shift.}

\clearpage

\subsection{Conditional sample stability}

\revisionblue{For Figure~\ref{fig:conditional-stability-l168}, we draw 100 length-512 references with replacement from eligible training windows in each dataset and generate 50 samples per reference draw using fixed discrete conditions and independent Gaussian sources. Statistics use the crop $[168,336)$, with the reference and generated samples standardized using the complete length-512 reference. Pointwise empirical quantiles define the median and central 50\% and 95\% intervals; inclusion rate is the fraction of reference values within the 95\% interval. We display the reference draw whose inclusion rate is closest to the dataset's 10th percentile, resolving ties by draw order, to illustrate a lower-coverage case.}

\section{Horizon-Wise Forecasting Results}
\label{app:forecast-aug}

\revisionblue{Table~\ref{tab:forecast-aug-h} compares \emph{Real only}, \emph{Real + MarkovTS} using a fixed generated counterpart pool and 50/50 counterpart-replacement sampling, and online generated-data replacement (\emph{Online}) across forecasting horizons. Training protocols are detailed in Appendix~\ref{app:forecast-protocol}. Bold Imp.\ rows report relative MAE and MSE reductions (\%) from \emph{Real only}.}

\begingroup
\scriptsize
\setlength{\tabcolsep}{1.0pt}
\renewcommand{\arraystretch}{0.72}
\setlength{\aboverulesep}{0.3ex}
\setlength{\belowrulesep}{0.45ex}
\setlength{\LTleft}{\fill}
\setlength{\LTcapwidth}{0.92\textwidth}
\setlength{\LTright}{\fill}
\begin{longtable}{@{}ll*{15}{r}@{}}
\caption{Forecasting MAE and MSE by horizon (lower is better). R: Real only; R+M: Real + MarkovTS; O: Online. Imp.: relative error reduction (\%) from R.}
\label{tab:forecast-aug-h}\\
\toprule
 & & \multicolumn{3}{c}{CATS} & \multicolumn{3}{c}{iTransformer} & \multicolumn{3}{c}{SparseTSF} & \multicolumn{3}{c}{TQNet} & \multicolumn{3}{c}{OLinear} \\
\cmidrule(lr){3-5}\cmidrule(lr){6-8}\cmidrule(lr){9-11}\cmidrule(lr){12-14}\cmidrule(lr){15-17}
Dataset & $H$ & R & R+M & O & R & R+M & O & R & R+M & O & R & R+M & O & R & R+M & O \\
\midrule
\endfirsthead
\caption[]{Horizon-wise forecasting results (continued).}\\
\toprule
 & & \multicolumn{3}{c}{CATS} & \multicolumn{3}{c}{iTransformer} & \multicolumn{3}{c}{SparseTSF} & \multicolumn{3}{c}{TQNet} & \multicolumn{3}{c}{OLinear} \\
\cmidrule(lr){3-5}\cmidrule(lr){6-8}\cmidrule(lr){9-11}\cmidrule(lr){12-14}\cmidrule(lr){15-17}
Dataset & $H$ & R & R+M & O & R & R+M & O & R & R+M & O & R & R+M & O & R & R+M & O \\
\midrule
\endhead
\bottomrule
\endfoot
\multicolumn{17}{c}{\textbf{MAE $\downarrow$}}\\*
\midrule
\multirow{6}{*}{ECL} & 96 & .232 & .217 & .223 & .230 & .233 & .235 & .244 & .237 & .238 & .220 & .218 & .221 & .274 & .273 & .274 \\*
 & 192 & .250 & .233 & .239 & .252 & .248 & .252 & .252 & .248 & .249 & .239 & .237 & .240 & .287 & .288 & .289 \\*
 & 336 & .260 & .250 & .256 & .264 & .264 & .268 & .266 & .264 & .264 & .248 & .252 & .256 & .301 & .303 & .304 \\*
 & 720 & .291 & .281 & .290 & .289 & .299 & .302 & .297 & .295 & .296 & .277 & .280 & .287 & .329 & .323 & .338 \\*
 & Mean & .258 & .245 & .252 & .259 & .261 & .264 & .265 & .261 & .262 & .246 & .247 & .251 & .298 & .297 & .301 \\*
\rowcolor{improw}
 & \textbf{Imp.} &  & \textbf{5.0} & \textbf{2.3} &  & \textbf{-0.8} & \textbf{-1.9} &  & \textbf{1.5} & \textbf{1.1} &  & \textbf{-0.4} & \textbf{-2.0} &  & \textbf{0.3} & \textbf{-1.0} \\
\midrule
\multirow{6}{*}{ETTh1} & 96 & .411 & .387 & .387 & .420 & .395 & .397 & .391 & .388 & .390 & .388 & .389 & .392 & .395 & .392 & .394 \\*
 & 192 & .432 & .413 & .410 & .463 & .420 & .422 & .414 & .410 & .414 & .414 & .414 & .419 & .417 & .418 & .420 \\*
 & 336 & .447 & .432 & .430 & .462 & .439 & .441 & .440 & .425 & .427 & .432 & .436 & .438 & .437 & .436 & .435 \\*
 & 720 & .483 & .462 & .458 & .546 & .479 & .475 & .471 & .443 & .445 & .466 & .471 & .468 & .463 & .470 & .477 \\*
 & Mean & .443 & .423 & .421 & .473 & .433 & .434 & .429 & .417 & .419 & .425 & .428 & .429 & .428 & .429 & .432 \\*
\rowcolor{improw}
 & \textbf{Imp.} &  & \textbf{4.5} & \textbf{5.0} &  & \textbf{8.5} & \textbf{8.2} &  & \textbf{2.8} & \textbf{2.3} &  & \textbf{-0.7} & \textbf{-0.9} &  & \textbf{-0.2} & \textbf{-0.9} \\
\midrule
\multirow{6}{*}{ETTh2} & 96 & .368 & .331 & .330 & .382 & .365 & .362 & .370 & .355 & .357 & .358 & .357 & .360 & .358 & .355 & .356 \\*
 & 192 & .417 & .382 & .378 & .440 & .414 & .414 & .413 & .406 & .409 & .408 & .408 & .406 & .406 & .405 & .402 \\*
 & 336 & .443 & .400 & .403 & .451 & .446 & .444 & .438 & .436 & .443 & .437 & .436 & .429 & .433 & .429 & .428 \\*
 & 720 & .472 & .422 & .425 & .484 & .481 & .483 & .486 & .460 & .462 & .467 & .463 & .457 & .468 & .460 & .460 \\*
 & Mean & .425 & .384 & .384 & .439 & .427 & .426 & .427 & .414 & .418 & .417 & .416 & .413 & .416 & .412 & .412 \\*
\rowcolor{improw}
 & \textbf{Imp.} &  & \textbf{9.6} & \textbf{9.6} &  & \textbf{2.7} & \textbf{3.0} &  & \textbf{3.0} & \textbf{2.1} &  & \textbf{0.2} & \textbf{1.0} &  & \textbf{1.0} & \textbf{1.0} \\
\midrule
\multirow{6}{*}{Exchange} & 96 & .215 & .208 & .205 & .235 & .211 & .215 & .235 & .202 & .206 & .209 & .205 & .207 & .219 & .213 & .216 \\*
 & 192 & .306 & .310 & .304 & .330 & .315 & .323 & .344 & .295 & .298 & .338 & .328 & .304 & .321 & .317 & .313 \\*
 & 336 & .431 & .429 & .435 & .493 & .443 & .447 & .499 & .417 & .420 & .472 & .436 & .434 & .467 & .448 & .431 \\*
 & 720 & .730 & .739 & .752 & .852 & .821 & .838 & .844 & .783 & .827 & .795 & .739 & .787 & .817 & .777 & .831 \\*
 & Mean & .420 & .421 & .424 & .478 & .447 & .456 & .481 & .425 & .438 & .454 & .427 & .433 & .456 & .439 & .448 \\*
\rowcolor{improw}
 & \textbf{Imp.} &  & \textbf{-0.2} & \textbf{-1.0} &  & \textbf{6.5} & \textbf{4.6} &  & \textbf{11.6} & \textbf{8.9} &  & \textbf{5.9} & \textbf{4.6} &  & \textbf{3.7} & \textbf{1.8} \\
\midrule
\multirow{6}{*}{Traffic} & 96 & .263 & .233 & .240 & .270 & .281 & .281 & .285 & .270 & .272 & .249 & .243 & .254 & .316 & .308 & .313 \\*
 & 192 & .283 & .242 & .249 & .281 & .292 & .291 & .290 & .273 & .274 & .261 & .254 & .262 & .323 & .317 & .320 \\*
 & 336 & .287 & .248 & .255 & .291 & .299 & .298 & .296 & .278 & .279 & .267 & .261 & .269 & .329 & .322 & .325 \\*
 & 720 & .305 & .283 & .282 & .311 & .315 & .314 & .311 & .294 & .294 & .284 & .279 & .285 & .343 & .335 & .339 \\*
 & Mean & .284 & .252 & .257 & .289 & .297 & .296 & .295 & .279 & .280 & .265 & .259 & .268 & .328 & .321 & .324 \\*
\rowcolor{improw}
 & \textbf{Imp.} &  & \textbf{11.6} & \textbf{9.8} &  & \textbf{-2.9} & \textbf{-2.6} &  & \textbf{5.7} & \textbf{5.3} &  & \textbf{2.2} & \textbf{-0.9} &  & \textbf{2.2} & \textbf{1.1} \\
\midrule
\multirow{6}{*}{Weather} & 96 & .193 & .182 & .184 & .206 & .194 & .192 & .235 & .217 & .217 & .182 & .183 & .185 & .188 & .188 & .190 \\*
 & 192 & .233 & .225 & .228 & .245 & .235 & .235 & .265 & .254 & .254 & .227 & .229 & .228 & .227 & .230 & .230 \\*
 & 336 & .274 & .266 & .271 & .284 & .276 & .276 & .294 & .291 & .291 & .265 & .271 & .273 & .264 & .271 & .272 \\*
 & 720 & .331 & .323 & .324 & .333 & .328 & .328 & .339 & .341 & .340 & .320 & .324 & .329 & .320 & .326 & .330 \\*
 & Mean & .258 & .249 & .252 & .267 & .258 & .258 & .283 & .276 & .276 & .249 & .252 & .254 & .250 & .254 & .255 \\*
\rowcolor{improw}
 & \textbf{Imp.} &  & \textbf{3.5} & \textbf{2.3} &  & \textbf{3.4} & \textbf{3.4} &  & \textbf{2.5} & \textbf{2.5} &  & \textbf{-1.2} & \textbf{-2.0} &  & \textbf{-1.6} & \textbf{-2.0} \\
\midrule
\multicolumn{17}{c}{\textbf{MSE $\downarrow$}}\\*
\midrule
\multirow{6}{*}{ECL} & 96 & .137 & .130 & .134 & .135 & .138 & .140 & .150 & .144 & .144 & .130 & .127 & .130 & .159 & .157 & .158 \\*
 & 192 & .154 & .149 & .151 & .160 & .155 & .158 & .159 & .157 & .157 & .149 & .148 & .149 & .175 & .176 & .176 \\*
 & 336 & .165 & .165 & .169 & .169 & .171 & .175 & .172 & .173 & .173 & .158 & .162 & .166 & .190 & .193 & .194 \\*
 & 720 & .203 & .201 & .211 & .194 & .212 & .214 & .212 & .211 & .211 & .193 & .195 & .204 & .231 & .219 & .240 \\*
 & Mean & .165 & .161 & .166 & .164 & .169 & .172 & .173 & .171 & .171 & .158 & .158 & .162 & .188 & .186 & .192 \\*
\rowcolor{improw}
 & \textbf{Imp.} &  & \textbf{2.4} & \textbf{-0.6} &  & \textbf{-3.0} & \textbf{-4.9} &  & \textbf{1.2} & \textbf{1.2} &  & \textbf{0.0} & \textbf{-2.5} &  & \textbf{1.1} & \textbf{-2.1} \\
\midrule
\multirow{6}{*}{ETTh1} & 96 & .396 & .376 & .378 & .410 & .373 & .373 & .374 & .371 & .373 & .368 & .370 & .371 & .369 & .366 & .365 \\*
 & 192 & .437 & .418 & .411 & .480 & .414 & .412 & .414 & .410 & .412 & .413 & .415 & .417 & .409 & .410 & .411 \\*
 & 336 & .456 & .444 & .439 & .471 & .435 & .436 & .445 & .438 & .437 & .444 & .453 & .453 & .436 & .435 & .431 \\*
 & 720 & .503 & .464 & .457 & .581 & .460 & .458 & .473 & .428 & .430 & .462 & .477 & .469 & .442 & .451 & .463 \\*
 & Mean & .448 & .425 & .421 & .485 & .421 & .420 & .427 & .412 & .413 & .422 & .429 & .428 & .414 & .415 & .417 \\*
\rowcolor{improw}
 & \textbf{Imp.} &  & \textbf{5.1} & \textbf{6.0} &  & \textbf{13.2} & \textbf{13.4} &  & \textbf{3.5} & \textbf{3.3} &  & \textbf{-1.7} & \textbf{-1.4} &  & \textbf{-0.2} & \textbf{-0.7} \\
\midrule
\multirow{6}{*}{ETTh2} & 96 & .316 & .284 & .283 & .334 & .319 & .317 & .323 & .309 & .314 & .311 & .313 & .317 & .305 & .304 & .304 \\*
 & 192 & .384 & .364 & .350 & .431 & .395 & .395 & .385 & .384 & .396 & .388 & .387 & .380 & .380 & .379 & .375 \\*
 & 336 & .415 & .376 & .388 & .440 & .434 & .430 & .413 & .421 & .436 & .419 & .421 & .411 & .415 & .412 & .404 \\*
 & 720 & .455 & .397 & .400 & .480 & .488 & .475 & .477 & .444 & .447 & .447 & .452 & .443 & .456 & .449 & .449 \\*
 & Mean & .392 & .355 & .355 & .421 & .409 & .404 & .399 & .389 & .398 & .391 & .393 & .388 & .389 & .386 & .383 \\*
\rowcolor{improw}
 & \textbf{Imp.} &  & \textbf{9.4} & \textbf{9.4} &  & \textbf{2.9} & \textbf{4.0} &  & \textbf{2.5} & \textbf{0.3} &  & \textbf{-0.5} & \textbf{0.8} &  & \textbf{0.8} & \textbf{1.5} \\
\midrule
\multirow{6}{*}{Exchange} & 96 & .094 & .084 & .083 & .103 & .084 & .088 & .105 & .081 & .083 & .086 & .081 & .083 & .092 & .087 & .090 \\*
 & 192 & .185 & .183 & .176 & .207 & .182 & .194 & .221 & .166 & .169 & .215 & .202 & .179 & .192 & .187 & .183 \\*
 & 336 & .353 & .350 & .355 & .423 & .359 & .358 & .434 & .325 & .331 & .411 & .360 & .356 & .398 & .366 & .340 \\*
 & 720 & .921 & .932 & .968 & 1.124 & 1.051 & 1.085 & 1.155 & .988 & 1.062 & 1.001 & .903 & 1.025 & 1.059 & .971 & 1.091 \\*
 & Mean & .388 & .387 & .395 & .464 & .419 & .431 & .478 & .390 & .411 & .428 & .387 & .411 & .435 & .402 & .426 \\*
\rowcolor{improw}
 & \textbf{Imp.} &  & \textbf{0.3} & \textbf{-1.8} &  & \textbf{9.7} & \textbf{7.1} &  & \textbf{18.4} & \textbf{14.0} &  & \textbf{9.6} & \textbf{4.0} &  & \textbf{7.6} & \textbf{2.1} \\
\midrule
\multirow{6}{*}{Traffic} & 96 & .382 & .368 & .373 & .375 & .401 & .399 & .424 & .438 & .434 & .377 & .365 & .379 & .436 & .421 & .430 \\*
 & 192 & .418 & .389 & .393 & .403 & .420 & .418 & .437 & .448 & .444 & .403 & .388 & .397 & .453 & .443 & .446 \\*
 & 336 & .431 & .402 & .406 & .420 & .434 & .433 & .448 & .460 & .457 & .417 & .403 & .412 & .468 & .455 & .461 \\*
 & 720 & .454 & .450 & .445 & .449 & .463 & .462 & .470 & .484 & .479 & .442 & .434 & .437 & .502 & .492 & .496 \\*
 & Mean & .421 & .402 & .404 & .412 & .429 & .428 & .445 & .458 & .453 & .410 & .398 & .406 & .465 & .453 & .458 \\*
\rowcolor{improw}
 & \textbf{Imp.} &  & \textbf{4.5} & \textbf{4.0} &  & \textbf{-4.3} & \textbf{-3.9} &  & \textbf{-2.9} & \textbf{-1.9} &  & \textbf{3.0} & \textbf{0.8} &  & \textbf{2.5} & \textbf{1.3} \\
\midrule
\multirow{6}{*}{Weather} & 96 & .145 & .151 & .152 & .158 & .153 & .153 & .185 & .183 & .183 & .144 & .144 & .145 & .145 & .146 & .145 \\*
 & 192 & .184 & .190 & .193 & .200 & .197 & .198 & .224 & .224 & .225 & .188 & .188 & .186 & .186 & .187 & .186 \\*
 & 336 & .236 & .238 & .242 & .251 & .249 & .249 & .267 & .272 & .272 & .237 & .241 & .243 & .233 & .238 & .238 \\*
 & 720 & .316 & .321 & .323 & .323 & .322 & .319 & .333 & .341 & .339 & .317 & .318 & .324 & .318 & .321 & .325 \\*
 & Mean & .221 & .225 & .228 & .233 & .230 & .230 & .252 & .255 & .255 & .222 & .223 & .225 & .221 & .223 & .224 \\*
\rowcolor{improw}
 & \textbf{Imp.} &  & \textbf{-1.8} & \textbf{-3.2} &  & \textbf{1.3} & \textbf{1.3} &  & \textbf{-1.2} & \textbf{-1.2} &  & \textbf{-0.5} & \textbf{-1.4} &  & \textbf{-0.9} & \textbf{-1.4} \\
\midrule
\end{longtable}
\endgroup

\section{Per-Dataset Classification-Augmentation Results}
\label{app:classification-augmentation}

\revisionblue{Table~\ref{tab:classification-augmentation-full} reports \emph{Real only}/\emph{Real + MarkovTS} for every dataset--classifier pair. All metrics except log loss are percentages and higher-is-better; lower log loss is better. Appendices~\ref{app:classification-protocol} and~\ref{app:evaluation-metrics} define training and metrics.}

\begingroup
\scriptsize
\setlength{\tabcolsep}{2.2pt}
\renewcommand{\arraystretch}{0.918}
\setlength{\LTleft}{0pt}
\setlength{\LTcapwidth}{\textwidth}
\setlength{\LTright}{0pt}
\begin{longtable}{@{\extracolsep{\fill}}lrrrrrrr}
\caption{Per-dataset classification: Real only/Real + MarkovTS. Arrows indicate better performance. For multiclass datasets, = Acc. denotes Sens. and Spec. equal to Acc. under the benchmark convention.}\label{tab:classification-augmentation-full}\\
\toprule
Dataset & Acc.$\uparrow$ & BAcc.$\uparrow$ & F1$\uparrow$ & AUROC$\uparrow$ & LL$\downarrow$ & Sens.$\uparrow$ & Spec.$\uparrow$ \\
\midrule
\endfirsthead
\multicolumn{8}{c}{\normalsize\tablename\ \thetable\ (continued)}\\
\toprule
Dataset & Acc.$\uparrow$ & BAcc.$\uparrow$ & F1$\uparrow$ & AUROC$\uparrow$ & LL$\downarrow$ & Sens.$\uparrow$ & Spec.$\uparrow$ \\
\midrule
\endhead
\midrule
\multicolumn{8}{r}{Continued on next page}\\
\endfoot
\bottomrule
\endlastfoot
\multicolumn{8}{l}{\textbf{ResNet1D}} \\
ArticularyWordRecognition & 95.78/88.78 & 95.78/88.78 & 95.78/87.48 & 99.94/99.71 & 0.16/0.42 & \multicolumn{2}{c}{= Acc.} \\
AtrialFibrillation & 11.11/13.33 & 11.11/13.33 & 8.85/10.73 & 37.56/40.22 & 4.18/4.13 & \multicolumn{2}{c}{= Acc.} \\
CharacterTrajectories & 98.84/99.28 & 98.75/99.21 & 98.84/99.28 & 99.98/99.99 & 0.05/0.03 & \multicolumn{2}{c}{= Acc.} \\
Cricket & 98.15/98.15 & 98.15/98.15 & 98.14/98.14 & 99.97/99.99 & 0.06/0.04 & \multicolumn{2}{c}{= Acc.} \\
DuckDuckGeese & 56.67/55.33 & 56.67/55.33 & 55.66/56.01 & 82.37/81.68 & 1.62/2.17 & \multicolumn{2}{c}{= Acc.} \\
ERing & 85.68/86.54 & 85.68/86.54 & 85.50/86.36 & 98.78/98.95 & 0.40/0.36 & \multicolumn{2}{c}{= Acc.} \\
Epilepsy & 95.65/96.14 & 95.81/96.22 & 95.67/96.12 & 99.13/99.53 & 0.20/0.13 & \multicolumn{2}{c}{= Acc.} \\
EthanolConcentration & 26.74/25.48 & 26.72/25.48 & 25.55/23.80 & 53.90/53.04 & 1.69/1.93 & \multicolumn{2}{c}{= Acc.} \\
HandMovementDirection & 35.13/34.69 & 36.15/36.83 & 32.78/33.00 & 61.40/63.20 & 1.49/1.60 & \multicolumn{2}{c}{= Acc.} \\
Handwriting & 52.59/55.41 & 52.15/55.10 & 49.54/52.43 & 93.31/93.16 & 1.76/1.67 & \multicolumn{2}{c}{= Acc.} \\
Heartbeat & 70.57/62.44 & 66.85/70.03 & 50.85/56.35 & 76.23/76.58 & 0.59/0.88 & 58.48/87.14 & 75.22/52.93 \\
JapaneseVowels & 98.29/96.85 & 98.23/96.91 & 98.29/96.87 & 99.98/99.92 & 0.06/0.10 & \multicolumn{2}{c}{= Acc.} \\
LSST & 42.92/60.16 & 39.87/48.02 & 43.51/58.84 & 80.08/88.31 & 3.32/1.90 & \multicolumn{2}{c}{= Acc.} \\
Libras & 92.78/91.85 & 92.78/91.85 & 92.69/91.82 & 99.63/99.65 & 0.26/0.33 & \multicolumn{2}{c}{= Acc.} \\
MotorImagery & 49.33/49.00 & 49.33/49.00 & 34.91/38.57 & 49.11/47.84 & 1.87/1.27 & 27.33/32.67 & 71.33/65.33 \\
NATOPS & 94.63/94.81 & 94.63/94.81 & 94.46/94.76 & 99.77/99.64 & 0.17/0.14 & \multicolumn{2}{c}{= Acc.} \\
PEMS-SF & 70.14/77.65 & 70.70/77.38 & 69.46/77.48 & 94.70/96.64 & 0.99/0.73 & \multicolumn{2}{c}{= Acc.} \\
PhonemeSpectra & 29.14/28.53 & 29.14/28.54 & 28.27/27.98 & 89.02/88.40 & 3.30/3.78 & \multicolumn{2}{c}{= Acc.} \\
RacketSports & 78.73/88.82 & 79.64/89.69 & 78.52/88.83 & 94.31/97.65 & 1.06/0.34 & \multicolumn{2}{c}{= Acc.} \\
SelfRegulationSCP1 & 72.47/72.01 & 72.42/72.02 & 56.14/68.22 & 67.51/87.80 & 0.98/0.90 & 57.53/72.60 & 87.30/71.43 \\
SpokenArabicDigits & 98.51/98.85 & 98.52/98.85 & 98.51/98.85 & 99.98/100.00 & 0.05/0.03 & \multicolumn{2}{c}{= Acc.} \\
StandWalkJump & 31.11/28.89 & 31.11/28.89 & 22.20/19.54 & 51.56/47.11 & 1.13/1.17 & \multicolumn{2}{c}{= Acc.} \\
UWaveGestureLibrary & 82.50/82.71 & 82.50/82.71 & 82.11/82.26 & 97.09/97.12 & 0.58/0.59 & \multicolumn{2}{c}{= Acc.} \\
\midrule
\multicolumn{8}{l}{\textbf{TimesNet}} \\
ArticularyWordRecognition & 94.67/97.33 & 94.67/97.33 & 94.47/97.34 & 99.68/99.93 & 0.85/0.18 & \multicolumn{2}{c}{= Acc.} \\
AtrialFibrillation & 26.67/40.00 & 26.67/40.00 & 21.67/38.89 & 34.67/39.33 & 1.20/1.19 & \multicolumn{2}{c}{= Acc.} \\
CharacterTrajectories & 98.12/97.56 & 97.98/97.37 & 98.12/97.56 & 99.90/99.96 & 0.09/0.09 & \multicolumn{2}{c}{= Acc.} \\
Cricket & 94.44/93.06 & 94.44/93.06 & 94.36/92.17 & 99.96/99.94 & 0.19/0.25 & \multicolumn{2}{c}{= Acc.} \\
DuckDuckGeese & 42.00/52.00 & 42.00/52.00 & 41.05/48.69 & 72.05/80.20 & 1.49/1.35 & \multicolumn{2}{c}{= Acc.} \\
ERing & 75.93/90.74 & 75.93/90.74 & 74.04/90.79 & 96.86/99.03 & 1.18/0.60 & \multicolumn{2}{c}{= Acc.} \\
Epilepsy & 89.86/90.58 & 89.83/90.62 & 89.25/90.24 & 98.63/98.80 & 0.28/0.29 & \multicolumn{2}{c}{= Acc.} \\
EthanolConcentration & 29.66/25.48 & 29.55/25.39 & 19.86/22.66 & 57.46/55.09 & 1.84/1.78 & \multicolumn{2}{c}{= Acc.} \\
HandMovementDirection & 56.76/60.81 & 58.93/62.14 & 56.19/59.41 & 79.85/79.96 & 1.15/1.16 & \multicolumn{2}{c}{= Acc.} \\
Handwriting & 17.41/22.71 & 16.67/22.04 & 13.30/19.81 & 68.58/71.25 & 3.06/3.13 & \multicolumn{2}{c}{= Acc.} \\
Heartbeat & 68.29/72.19 & 64.02/52.70 & 48.82/14.93 & 73.19/73.90 & 0.59/0.57 & 54.39/8.77 & 73.65/96.62 \\
JapaneseVowels & 97.84/97.84 & 97.87/98.15 & 97.84/97.85 & 99.94/99.91 & 0.09/0.12 & \multicolumn{2}{c}{= Acc.} \\
LSST & 57.62/53.73 & 40.62/38.73 & 53.95/51.53 & 85.33/83.98 & 1.43/1.55 & \multicolumn{2}{c}{= Acc.} \\
Libras & 75.00/73.33 & 75.00/73.33 & 74.20/73.20 & 97.19/97.35 & 0.81/0.74 & \multicolumn{2}{c}{= Acc.} \\
MotorImagery & 52.00/52.00 & 52.00/52.00 & 56.36/57.90 & 48.88/50.76 & 1.05/1.20 & 62.00/66.00 & 42.00/38.00 \\
NATOPS & 83.33/92.78 & 83.33/92.78 & 83.13/92.80 & 96.90/99.39 & 0.36/0.18 & \multicolumn{2}{c}{= Acc.} \\
PEMS-SF & 72.83/72.83 & 73.23/72.52 & 72.36/72.57 & 94.88/95.32 & 0.75/0.78 & \multicolumn{2}{c}{= Acc.} \\
PhonemeSpectra & 12.59/11.90 & 12.59/11.90 & 12.04/11.48 & 73.78/74.08 & 7.44/7.03 & \multicolumn{2}{c}{= Acc.} \\
RacketSports & 83.55/81.58 & 84.68/82.94 & 83.10/81.24 & 94.46/92.42 & 0.44/0.51 & \multicolumn{2}{c}{= Acc.} \\
SelfRegulationSCP1 & 81.91/83.96 & 81.96/84.00 & 84.08/85.63 & 95.35/95.27 & 0.39/0.47 & 95.89/95.89 & 68.03/72.11 \\
SpokenArabicDigits & 98.14/97.45 & 98.14/97.45 & 98.14/97.46 & 99.97/99.95 & 0.09/0.11 & \multicolumn{2}{c}{= Acc.} \\
StandWalkJump & 26.67/33.33 & 26.67/33.33 & 25.45/29.10 & 52.00/52.00 & 1.12/1.12 & \multicolumn{2}{c}{= Acc.} \\
UWaveGestureLibrary & 79.06/79.06 & 79.06/79.06 & 77.04/77.72 & 96.89/97.12 & 0.87/0.86 & \multicolumn{2}{c}{= Acc.} \\
\midrule
\multicolumn{8}{l}{\textbf{ConvTran}} \\
ArticularyWordRecognition & 96.33/97.00 & 96.33/97.00 & 96.11/96.94 & 99.92/99.98 & 0.19/0.10 & \multicolumn{2}{c}{= Acc.} \\
AtrialFibrillation & 33.33/33.33 & 33.33/33.33 & 27.32/26.50 & 52.67/55.33 & 1.13/1.11 & \multicolumn{2}{c}{= Acc.} \\
CharacterTrajectories & 99.16/98.89 & 99.12/98.80 & 99.17/98.88 & 100.00/99.99 & 0.03/0.04 & \multicolumn{2}{c}{= Acc.} \\
Cricket & 97.22/98.61 & 97.22/98.61 & 97.21/98.60 & 99.98/99.96 & 0.23/0.07 & \multicolumn{2}{c}{= Acc.} \\
DuckDuckGeese & 52.00/52.00 & 52.00/52.00 & 51.48/51.10 & 83.95/80.40 & 1.23/1.58 & \multicolumn{2}{c}{= Acc.} \\
ERing & 87.04/87.41 & 87.04/87.41 & 86.66/87.31 & 98.10/98.10 & 0.82/0.63 & \multicolumn{2}{c}{= Acc.} \\
Epilepsy & 95.65/96.38 & 95.77/96.50 & 95.64/96.38 & 99.51/99.56 & 0.16/0.18 & \multicolumn{2}{c}{= Acc.} \\
EthanolConcentration & 31.56/26.62 & 31.53/26.57 & 31.33/25.67 & 58.38/56.90 & 1.45/1.50 & \multicolumn{2}{c}{= Acc.} \\
HandMovementDirection & 35.13/33.78 & 36.07/36.43 & 33.93/33.23 & 62.80/61.49 & 1.34/1.35 & \multicolumn{2}{c}{= Acc.} \\
Handwriting & 22.59/23.88 & 22.07/24.09 & 17.13/19.83 & 79.36/80.32 & 2.69/2.74 & \multicolumn{2}{c}{= Acc.} \\
Heartbeat & 74.15/75.61 & 65.38/67.47 & 49.52/52.83 & 74.55/76.90 & 0.52/0.52 & 45.61/49.12 & 85.14/85.81 \\
JapaneseVowels & 99.19/97.84 & 99.30/98.45 & 99.19/97.84 & 99.98/99.99 & 0.05/0.08 & \multicolumn{2}{c}{= Acc.} \\
LSST & 62.13/62.57 & 38.89/40.68 & 58.21/58.28 & 88.08/88.80 & 1.19/1.16 & \multicolumn{2}{c}{= Acc.} \\
Libras & 83.89/88.33 & 83.89/88.33 & 83.81/88.34 & 99.00/99.33 & 0.62/0.37 & \multicolumn{2}{c}{= Acc.} \\
MotorImagery & 50.00/60.00 & 50.00/60.00 & 66.67/55.56 & 51.36/64.84 & 0.80/0.66 & 100.00/50.00 & 0.00/70.00 \\
NATOPS & 95.00/95.00 & 95.00/95.00 & 95.07/94.92 & 99.84/99.77 & 0.14/0.18 & \multicolumn{2}{c}{= Acc.} \\
PEMS-SF & 68.21/68.79 & 68.37/69.36 & 67.33/68.15 & 90.31/92.45 & 1.06/0.99 & \multicolumn{2}{c}{= Acc.} \\
PhonemeSpectra & 26.36/27.35 & 26.37/27.35 & 25.69/25.94 & 90.57/90.65 & 2.38/2.37 & \multicolumn{2}{c}{= Acc.} \\
RacketSports & 82.89/83.55 & 84.27/84.98 & 82.78/83.18 & 92.57/92.53 & 0.58/0.66 & \multicolumn{2}{c}{= Acc.} \\
SelfRegulationSCP1 & 90.10/86.69 & 90.12/86.72 & 90.61/87.93 & 97.36/96.63 & 0.28/0.34 & 95.89/97.26 & 84.35/76.19 \\
SpokenArabicDigits & 98.27/98.95 & 98.27/98.95 & 98.27/98.95 & 99.98/100.00 & 0.07/0.04 & \multicolumn{2}{c}{= Acc.} \\
StandWalkJump & 26.67/33.33 & 26.67/33.33 & 22.22/16.67 & 51.33/38.00 & 1.25/1.11 & \multicolumn{2}{c}{= Acc.} \\
UWaveGestureLibrary & 87.19/89.38 & 87.19/89.38 & 87.08/89.19 & 97.43/98.08 & 0.53/0.43 & \multicolumn{2}{c}{= Acc.} \\
\midrule
\multicolumn{8}{l}{\textbf{Arsenal}} \\
ArticularyWordRecognition & 99.33/99.33 & 99.33/99.33 & 99.33/99.33 & 99.83/99.65 & 0.13/0.23 & \multicolumn{2}{c}{= Acc.} \\
AtrialFibrillation & 6.67/6.67 & 6.67/6.67 & 5.56/5.56 & 27.00/27.33 & 33.64/32.24 & \multicolumn{2}{c}{= Acc.} \\
CharacterTrajectories & 99.30/98.95 & 99.28/98.88 & 99.30/98.95 & 99.78/99.71 & 0.16/0.21 & \multicolumn{2}{c}{= Acc.} \\
Cricket & 100.00/100.00 & 100.00/100.00 & 100.00/100.00 & 100.00/100.00 & 0.00/0.00 & \multicolumn{2}{c}{= Acc.} \\
DuckDuckGeese & 48.00/56.00 & 48.00/56.00 & 48.47/56.10 & 78.17/81.50 & 7.87/6.21 & \multicolumn{2}{c}{= Acc.} \\
ERing & 98.89/97.41 & 98.89/97.41 & 98.89/97.42 & 99.55/99.14 & 0.28/0.53 & \multicolumn{2}{c}{= Acc.} \\
Epilepsy & 98.55/99.28 & 98.53/99.27 & 98.54/99.28 & 100.00/99.99 & 0.01/0.02 & \multicolumn{2}{c}{= Acc.} \\
EthanolConcentration & 42.59/35.36 & 42.68/35.41 & 42.34/35.41 & 67.51/61.05 & 9.38/12.35 & \multicolumn{2}{c}{= Acc.} \\
HandMovementDirection & 50.00/52.70 & 48.69/53.21 & 49.89/52.91 & 72.63/76.03 & 6.59/4.56 & \multicolumn{2}{c}{= Acc.} \\
Handwriting & 59.18/58.94 & 58.50/58.34 & 56.58/56.41 & 85.06/85.31 & 10.36/9.68 & \multicolumn{2}{c}{= Acc.} \\
Heartbeat & 76.10/75.61 & 64.03/63.69 & 46.15/45.65 & 71.40/72.39 & 3.99/3.07 & 36.84/36.84 & 91.22/90.54 \\
JapaneseVowels & 83.24/82.43 & 82.82/81.97 & 82.68/81.74 & 95.99/95.99 & 2.44/2.31 & \multicolumn{2}{c}{= Acc.} \\
LSST & 65.69/63.34 & 37.06/33.39 & 60.32/57.40 & 84.86/84.48 & 6.66/6.16 & \multicolumn{2}{c}{= Acc.} \\
Libras & 90.00/90.00 & 90.00/90.00 & 90.04/90.04 & 96.43/95.55 & 2.47/2.90 & \multicolumn{2}{c}{= Acc.} \\
MotorImagery & 50.00/49.00 & 50.00/49.00 & 47.92/51.43 & 48.04/50.18 & 2.00/1.61 & 46.00/54.00 & 54.00/44.00 \\
NATOPS & 88.33/90.00 & 88.33/90.00 & 87.99/89.79 & 97.40/97.54 & 1.34/1.27 & \multicolumn{2}{c}{= Acc.} \\
PEMS-SF & 82.66/82.66 & 82.01/82.01 & 82.14/82.14 & 94.42/96.20 & 2.94/1.56 & \multicolumn{2}{c}{= Acc.} \\
PhonemeSpectra & 29.88/29.56 & 29.89/29.56 & 29.09/28.91 & 75.84/76.47 & 15.67/14.37 & \multicolumn{2}{c}{= Acc.} \\
RacketSports & 90.13/90.79 & 90.93/91.47 & 90.13/90.75 & 96.06/96.05 & 1.76/1.69 & \multicolumn{2}{c}{= Acc.} \\
SelfRegulationSCP1 & 85.32/85.32 & 85.31/85.31 & 84.81/84.70 & 92.12/93.36 & 1.59/1.07 & 82.19/81.51 & 88.44/89.12 \\
SpokenArabicDigits & 99.14/99.09 & 99.14/99.09 & 99.13/99.09 & 99.88/99.91 & 0.10/0.08 & \multicolumn{2}{c}{= Acc.} \\
StandWalkJump & 46.67/46.67 & 46.67/46.67 & 43.37/43.39 & 61.33/66.00 & 17.02/14.22 & \multicolumn{2}{c}{= Acc.} \\
UWaveGestureLibrary & 94.06/93.75 & 94.06/93.75 & 94.04/93.72 & 97.65/97.66 & 1.52/1.44 & \multicolumn{2}{c}{= Acc.} \\
\end{longtable}
\endgroup

\section{Per-Dataset Anomaly-Detection Augmentation Results}
\label{app:anomaly-augmentation}

\revisionblue{Table~\ref{tab:anomaly-augmentation-full} reports \emph{Real only}/\emph{Real + MarkovTS} per dataset--detector pair to three decimals; all six metrics are higher-is-better. Appendices~\ref{app:anomaly-protocol} and~\ref{app:evaluation-metrics} define protocols and metrics.}

\begingroup
\scriptsize
\setlength{\tabcolsep}{2.4pt}
\renewcommand{\arraystretch}{0.9}
\setlength{\LTleft}{0pt}
\setlength{\LTcapwidth}{\textwidth}
\setlength{\LTright}{0pt}
\begin{longtable}{@{\extracolsep{\fill}}lrrrrrr}
\caption{Per-dataset anomaly detection: Real only/Real + MarkovTS. Higher is better.}\label{tab:anomaly-augmentation-full}\\
\toprule
Dataset & A-P$\uparrow$ & A-R$\uparrow$ & R-A-P$\uparrow$ & R-A-R$\uparrow$ & V-PR$\uparrow$ & V-ROC$\uparrow$ \\
\midrule
\endfirsthead
\multicolumn{7}{c}{\normalsize\tablename\ \thetable\ (continued)}\\
\toprule
Dataset & A-P$\uparrow$ & A-R$\uparrow$ & R-A-P$\uparrow$ & R-A-R$\uparrow$ & V-PR$\uparrow$ & V-ROC$\uparrow$ \\
\midrule
\endhead
\midrule
\multicolumn{7}{r}{Continued on next page}\\
\endfoot
\bottomrule
\endlastfoot
\multicolumn{7}{l}{\textbf{DLinear}} \\
ASD & 0.179/0.172 & 0.762/0.762 & 0.208/0.204 & 0.815/0.814 & 0.205/0.200 & 0.806/0.806 \\
CATSv2 & 0.064/0.063 & 0.567/0.564 & 0.078/0.077 & 0.654/0.650 & 0.079/0.078 & 0.648/0.645 \\
CICIDS & 0.002/0.002 & 0.748/0.749 & 0.002/0.002 & 0.683/0.680 & 0.002/0.002 & 0.699/0.700 \\
CalIt2 & 0.100/0.108 & 0.776/0.800 & 0.101/0.107 & 0.808/0.826 & 0.099/0.105 & 0.799/0.818 \\
Creditcard & 0.070/0.055 & 0.919/0.748 & 0.036/0.006 & 0.847/0.589 & 0.035/0.010 & 0.851/0.622 \\
DAPHNET & 0.196/0.198 & 0.735/0.746 & 0.316/0.320 & 0.816/0.825 & 0.307/0.310 & 0.809/0.818 \\
DLR & 0.160/0.134 & 0.916/0.909 & 0.141/0.123 & 0.904/0.903 & 0.144/0.125 & 0.905/0.904 \\
ECG & 0.550/0.550 & 0.848/0.837 & 0.475/0.468 & 0.790/0.774 & 0.522/0.523 & 0.822/0.811 \\
Exathlon & 0.798/0.794 & 0.900/0.889 & 0.733/0.714 & 0.871/0.861 & 0.740/0.722 & 0.872/0.862 \\
GECCO & 0.116/0.110 & 0.791/0.786 & 0.163/0.163 & 0.891/0.889 & 0.153/0.151 & 0.875/0.870 \\
GHL & 0.010/0.010 & 0.322/0.305 & 0.018/0.017 & 0.492/0.480 & 0.018/0.017 & 0.478/0.465 \\
GutenTAG & 0.103/0.151 & 0.682/0.688 & 0.145/0.198 & 0.763/0.786 & 0.136/0.191 & 0.751/0.772 \\
LTDB & 0.204/0.189 & 0.599/0.563 & 0.310/0.306 & 0.710/0.695 & 0.300/0.292 & 0.695/0.674 \\
MITDB & 0.054/0.059 & 0.583/0.572 & 0.078/0.082 & 0.669/0.660 & 0.078/0.082 & 0.661/0.651 \\
MSL & 0.139/0.134 & 0.601/0.584 & 0.208/0.203 & 0.678/0.660 & 0.206/0.200 & 0.671/0.653 \\
NYC & 0.038/0.032 & 0.699/0.631 & 0.053/0.049 & 0.714/0.682 & 0.053/0.049 & 0.712/0.675 \\
PSM & 0.352/0.342 & 0.559/0.550 & 0.353/0.343 & 0.559/0.550 & 0.353/0.343 & 0.557/0.546 \\
PUMP & 0.120/0.120 & 0.562/0.565 & 0.178/0.175 & 0.653/0.653 & 0.177/0.174 & 0.647/0.648 \\
SKAB & 0.420/0.416 & 0.570/0.564 & 0.658/0.649 & 0.733/0.726 & 0.642/0.635 & 0.725/0.719 \\
SMAP & 0.104/0.104 & 0.391/0.394 & 0.118/0.118 & 0.427/0.430 & 0.119/0.119 & 0.427/0.430 \\
SMD & 0.133/0.135 & 0.708/0.706 & 0.139/0.139 & 0.733/0.729 & 0.138/0.139 & 0.731/0.728 \\
SVDB & 0.052/0.048 & 0.610/0.579 & 0.091/0.087 & 0.710/0.688 & 0.085/0.081 & 0.696/0.673 \\
SWAN & 0.446/0.454 & 0.494/0.507 & 0.369/0.378 & 0.392/0.404 & 0.401/0.410 & 0.436/0.447 \\
SWAT & 0.085/0.630 & 0.304/0.819 & 0.116/0.589 & 0.397/0.813 & 0.114/0.576 & 0.390/0.804 \\
TAO & 1.000/1.000 & 1.000/1.000 & 0.999/0.999 & 1.000/1.000 & 0.955/0.960 & 0.987/0.990 \\
TODS & 0.576/0.573 & 0.828/0.824 & 0.549/0.579 & 0.832/0.834 & 0.535/0.559 & 0.822/0.824 \\
\midrule
\multicolumn{7}{l}{\textbf{TranAD}} \\
ASD & 0.149/0.151 & 0.619/0.607 & 0.133/0.138 & 0.640/0.620 & 0.133/0.139 & 0.635/0.616 \\
CATSv2 & 0.126/0.127 & 0.594/0.603 & 0.128/0.130 & 0.625/0.630 & 0.127/0.129 & 0.617/0.623 \\
CICIDS & 0.001/0.001 & 0.698/0.680 & 0.002/0.002 & 0.627/0.604 & 0.002/0.002 & 0.647/0.629 \\
CalIt2 & 0.053/0.061 & 0.520/0.577 & 0.062/0.070 & 0.554/0.615 & 0.060/0.069 & 0.544/0.604 \\
Creditcard & 0.107/0.109 & 0.952/0.954 & 0.060/0.061 & 0.907/0.911 & 0.057/0.058 & 0.902/0.906 \\
DAPHNET & 0.154/0.135 & 0.571/0.502 & 0.262/0.224 & 0.683/0.597 & 0.253/0.216 & 0.671/0.587 \\
DLR & 0.241/0.253 & 0.895/0.903 & 0.266/0.204 & 0.952/0.955 & 0.268/0.203 & 0.945/0.950 \\
ECG & 0.524/0.484 & 0.872/0.859 & 0.476/0.443 & 0.807/0.793 & 0.493/0.460 & 0.817/0.803 \\
Exathlon & 0.920/0.920 & 0.947/0.947 & 0.844/0.846 & 0.920/0.920 & 0.847/0.847 & 0.919/0.918 \\
GECCO & 0.269/0.300 & 0.719/0.789 & 0.052/0.063 & 0.586/0.673 & 0.052/0.063 & 0.587/0.674 \\
GHL & 0.013/0.015 & 0.303/0.301 & 0.017/0.019 & 0.339/0.349 & 0.017/0.019 & 0.336/0.346 \\
GutenTAG & 0.025/0.035 & 0.569/0.585 & 0.033/0.038 & 0.591/0.607 & 0.033/0.038 & 0.589/0.605 \\
LTDB & 0.217/0.219 & 0.590/0.596 & 0.330/0.321 & 0.719/0.710 & 0.321/0.314 & 0.705/0.699 \\
MITDB & 0.108/0.108 & 0.691/0.687 & 0.126/0.125 & 0.782/0.777 & 0.127/0.126 & 0.773/0.768 \\
MSL & 0.128/0.126 & 0.478/0.487 & 0.180/0.180 & 0.562/0.571 & 0.179/0.179 & 0.556/0.564 \\
NYC & 0.034/0.032 & 0.676/0.675 & 0.048/0.047 & 0.696/0.698 & 0.047/0.046 & 0.688/0.688 \\
PSM & 0.502/0.451 & 0.635/0.624 & 0.438/0.407 & 0.573/0.557 & 0.438/0.407 & 0.574/0.558 \\
PUMP & 0.210/0.211 & 0.802/0.802 & 0.251/0.254 & 0.796/0.796 & 0.249/0.252 & 0.794/0.794 \\
SKAB & 0.750/0.767 & 0.827/0.836 & 0.852/0.860 & 0.871/0.877 & 0.843/0.852 & 0.872/0.878 \\
SMAP & 0.097/0.097 & 0.369/0.374 & 0.112/0.112 & 0.378/0.387 & 0.112/0.112 & 0.376/0.386 \\
SMD & 0.112/0.107 & 0.663/0.654 & 0.105/0.100 & 0.670/0.663 & 0.105/0.101 & 0.669/0.662 \\
SVDB & 0.064/0.062 & 0.577/0.567 & 0.088/0.086 & 0.657/0.650 & 0.086/0.083 & 0.653/0.645 \\
SWAN & 0.530/0.556 & 0.603/0.632 & 0.408/0.432 & 0.423/0.456 & 0.437/0.460 & 0.466/0.496 \\
SWAT & 0.728/0.722 & 0.818/0.817 & 0.527/0.503 & 0.718/0.700 & 0.498/0.478 & 0.691/0.675 \\
TAO & 1.000/1.000 & 1.000/1.000 & 0.999/0.999 & 1.000/1.000 & 0.948/0.948 & 0.980/0.980 \\
TODS & 0.174/0.173 & 0.525/0.525 & 0.134/0.134 & 0.491/0.491 & 0.134/0.134 & 0.487/0.487 \\
\midrule
\multicolumn{7}{l}{\textbf{CATCH}} \\
ASD & 0.170/0.186 & 0.738/0.790 & 0.193/0.226 & 0.790/0.839 & 0.191/0.221 & 0.782/0.830 \\
CATSv2 & 0.090/0.084 & 0.604/0.601 & 0.104/0.100 & 0.682/0.675 & 0.105/0.101 & 0.680/0.674 \\
CICIDS & 0.001/0.001 & 0.669/0.684 & 0.001/0.002 & 0.582/0.593 & 0.002/0.002 & 0.598/0.610 \\
CalIt2 & 0.089/0.101 & 0.772/0.790 & 0.104/0.109 & 0.811/0.818 & 0.101/0.107 & 0.796/0.808 \\
Creditcard & 0.102/0.100 & 0.953/0.951 & 0.052/0.052 & 0.912/0.908 & 0.050/0.050 & 0.909/0.906 \\
DAPHNET & 0.193/0.195 & 0.730/0.739 & 0.310/0.313 & 0.809/0.818 & 0.301/0.303 & 0.801/0.810 \\
DLR & 0.171/0.143 & 0.922/0.904 & 0.162/0.145 & 0.927/0.921 & 0.161/0.144 & 0.921/0.914 \\
ECG & 0.526/0.534 & 0.860/0.860 & 0.458/0.462 & 0.803/0.802 & 0.499/0.505 & 0.828/0.828 \\
Exathlon & 0.782/0.798 & 0.882/0.882 & 0.738/0.730 & 0.864/0.862 & 0.744/0.738 & 0.864/0.862 \\
GECCO & 0.198/0.227 & 0.912/0.941 & 0.275/0.320 & 0.963/0.976 & 0.260/0.300 & 0.953/0.969 \\
GHL & 0.009/0.009 & 0.233/0.246 & 0.016/0.017 & 0.442/0.451 & 0.016/0.017 & 0.428/0.437 \\
GutenTAG & 0.246/0.288 & 0.791/0.828 & 0.428/0.466 & 0.839/0.870 & 0.406/0.444 & 0.832/0.863 \\
LTDB & 0.213/0.207 & 0.596/0.578 & 0.321/0.321 & 0.713/0.709 & 0.310/0.306 & 0.695/0.685 \\
MITDB & 0.096/0.135 & 0.708/0.727 & 0.121/0.152 & 0.756/0.768 & 0.121/0.152 & 0.750/0.763 \\
MSL & 0.154/0.155 & 0.627/0.632 & 0.233/0.228 & 0.709/0.711 & 0.229/0.224 & 0.700/0.702 \\
NYC & 0.021/0.020 & 0.469/0.452 & 0.040/0.040 & 0.606/0.605 & 0.040/0.039 & 0.593/0.591 \\
PSM & 0.385/0.386 & 0.600/0.600 & 0.388/0.389 & 0.600/0.599 & 0.389/0.390 & 0.600/0.600 \\
PUMP & 0.104/0.104 & 0.466/0.476 & 0.169/0.170 & 0.591/0.598 & 0.168/0.169 & 0.583/0.590 \\
SKAB & 0.441/0.426 & 0.589/0.579 & 0.661/0.653 & 0.735/0.729 & 0.649/0.641 & 0.733/0.727 \\
SMAP & 0.115/0.116 & 0.431/0.430 & 0.129/0.129 & 0.459/0.456 & 0.129/0.129 & 0.457/0.455 \\
SMD & 0.164/0.172 & 0.782/0.790 & 0.167/0.174 & 0.804/0.810 & 0.166/0.173 & 0.802/0.808 \\
SVDB & 0.062/0.079 & 0.602/0.623 & 0.091/0.103 & 0.703/0.723 & 0.086/0.098 & 0.686/0.706 \\
SWAN & 0.425/0.437 & 0.485/0.501 & 0.346/0.357 & 0.373/0.386 & 0.376/0.387 & 0.415/0.428 \\
SWAT & 0.082/0.083 & 0.230/0.232 & 0.120/0.122 & 0.346/0.347 & 0.116/0.119 & 0.336/0.338 \\
TAO & 0.998/0.998 & 0.997/0.997 & 0.994/0.994 & 0.994/0.994 & 0.954/0.953 & 0.984/0.984 \\
TODS & 0.610/0.617 & 0.911/0.915 & 0.596/0.605 & 0.907/0.912 & 0.586/0.595 & 0.905/0.909 \\
\midrule
\multicolumn{7}{l}{\textbf{ModernTCN}} \\
ASD & 0.119/0.127 & 0.624/0.627 & 0.121/0.127 & 0.670/0.675 & 0.121/0.127 & 0.665/0.669 \\
CATSv2 & 0.109/0.106 & 0.630/0.633 & 0.127/0.125 & 0.716/0.716 & 0.127/0.125 & 0.708/0.710 \\
CICIDS & 0.001/0.001 & 0.670/0.678 & 0.001/0.002 & 0.607/0.618 & 0.002/0.002 & 0.622/0.632 \\
CalIt2 & 0.063/0.064 & 0.691/0.692 & 0.081/0.081 & 0.754/0.756 & 0.079/0.080 & 0.740/0.741 \\
Creditcard & 0.088/0.090 & 0.958/0.958 & 0.054/0.055 & 0.920/0.921 & 0.051/0.053 & 0.911/0.911 \\
DAPHNET & 0.182/0.180 & 0.710/0.710 & 0.293/0.291 & 0.793/0.792 & 0.284/0.282 & 0.785/0.784 \\
DLR & 0.289/0.272 & 0.961/0.959 & 0.203/0.206 & 0.954/0.955 & 0.206/0.208 & 0.957/0.957 \\
ECG & 0.522/0.524 & 0.865/0.866 & 0.468/0.471 & 0.807/0.808 & 0.502/0.505 & 0.830/0.831 \\
Exathlon & 0.784/0.787 & 0.887/0.887 & 0.755/0.751 & 0.867/0.865 & 0.760/0.757 & 0.867/0.865 \\
GECCO & 0.437/0.484 & 0.954/0.964 & 0.429/0.451 & 0.973/0.982 & 0.431/0.453 & 0.971/0.980 \\
GHL & 0.010/0.010 & 0.290/0.294 & 0.018/0.018 & 0.482/0.486 & 0.018/0.018 & 0.467/0.470 \\
GutenTAG & 0.061/0.064 & 0.607/0.609 & 0.141/0.147 & 0.707/0.710 & 0.124/0.130 & 0.695/0.698 \\
LTDB & 0.230/0.231 & 0.619/0.621 & 0.317/0.318 & 0.702/0.703 & 0.314/0.315 & 0.699/0.701 \\
MITDB & 0.112/0.107 & 0.663/0.670 & 0.129/0.128 & 0.734/0.736 & 0.129/0.128 & 0.727/0.729 \\
MSL & 0.145/0.147 & 0.621/0.625 & 0.224/0.221 & 0.701/0.703 & 0.221/0.218 & 0.693/0.695 \\
NYC & 0.045/0.044 & 0.696/0.711 & 0.061/0.062 & 0.723/0.734 & 0.060/0.060 & 0.716/0.727 \\
PSM & 0.388/0.388 & 0.587/0.590 & 0.387/0.386 & 0.584/0.585 & 0.387/0.386 & 0.584/0.582 \\
PUMP & 0.111/0.113 & 0.420/0.413 & 0.209/0.219 & 0.593/0.589 & 0.205/0.213 & 0.581/0.577 \\
SKAB & 0.439/0.439 & 0.586/0.587 & 0.674/0.674 & 0.746/0.746 & 0.659/0.659 & 0.739/0.740 \\
SMAP & 0.110/0.110 & 0.434/0.433 & 0.124/0.125 & 0.470/0.468 & 0.125/0.125 & 0.468/0.467 \\
SMD & 0.132/0.138 & 0.721/0.728 & 0.131/0.138 & 0.739/0.748 & 0.131/0.138 & 0.738/0.746 \\
SVDB & 0.063/0.064 & 0.593/0.594 & 0.086/0.086 & 0.666/0.666 & 0.084/0.084 & 0.662/0.663 \\
SWAN & 0.454/0.472 & 0.519/0.537 & 0.376/0.392 & 0.413/0.429 & 0.406/0.422 & 0.456/0.472 \\
SWAT & 0.092/0.092 & 0.240/0.242 & 0.135/0.136 & 0.354/0.356 & 0.130/0.130 & 0.345/0.346 \\
TAO & 0.999/0.999 & 1.000/1.000 & 0.997/0.998 & 1.000/1.000 & 0.949/0.950 & 0.983/0.983 \\
TODS & 0.173/0.206 & 0.618/0.634 & 0.183/0.211 & 0.640/0.662 & 0.184/0.211 & 0.640/0.660 \\
\midrule
\multicolumn{7}{l}{\textbf{TimesNet}} \\
ASD & 0.266/0.321 & 0.822/0.837 & 0.285/0.342 & 0.850/0.868 & 0.281/0.338 & 0.845/0.864 \\
CATSv2 & 0.261/0.278 & 0.712/0.732 & 0.266/0.286 & 0.779/0.790 & 0.265/0.284 & 0.772/0.783 \\
CICIDS & 0.002/0.002 & 0.691/0.705 & 0.002/0.002 & 0.589/0.608 & 0.002/0.002 & 0.600/0.615 \\
CalIt2 & 0.064/0.072 & 0.623/0.667 & 0.063/0.081 & 0.661/0.736 & 0.062/0.078 & 0.649/0.720 \\
Creditcard & 0.095/0.090 & 0.932/0.922 & 0.026/0.028 & 0.877/0.857 & 0.029/0.032 & 0.890/0.876 \\
DAPHNET & 0.211/0.205 & 0.748/0.744 & 0.329/0.320 & 0.825/0.823 & 0.319/0.311 & 0.818/0.815 \\
DLR & 0.065/0.066 & 0.910/0.910 & 0.073/0.075 & 0.910/0.910 & 0.073/0.074 & 0.908/0.909 \\
ECG & 0.437/0.393 & 0.808/0.801 & 0.373/0.343 & 0.744/0.735 & 0.421/0.393 & 0.775/0.764 \\
Exathlon & 0.794/0.798 & 0.920/0.916 & 0.759/0.770 & 0.890/0.892 & 0.765/0.774 & 0.891/0.891 \\
GECCO & 0.327/0.305 & 0.898/0.893 & 0.301/0.302 & 0.948/0.946 & 0.297/0.298 & 0.941/0.940 \\
GHL & 0.010/0.010 & 0.298/0.301 & 0.019/0.018 & 0.491/0.495 & 0.019/0.018 & 0.477/0.480 \\
GutenTAG & 0.227/0.295 & 0.753/0.766 & 0.244/0.350 & 0.822/0.827 & 0.238/0.334 & 0.812/0.819 \\
LTDB & 0.289/0.292 & 0.652/0.652 & 0.396/0.400 & 0.760/0.759 & 0.384/0.388 & 0.745/0.744 \\
MITDB & 0.104/0.105 & 0.657/0.656 & 0.141/0.139 & 0.732/0.733 & 0.140/0.139 & 0.725/0.726 \\
MSL & 0.148/0.140 & 0.612/0.581 & 0.229/0.216 & 0.702/0.671 & 0.225/0.212 & 0.693/0.662 \\
NYC & 0.022/0.021 & 0.505/0.469 & 0.056/0.054 & 0.652/0.633 & 0.053/0.051 & 0.636/0.610 \\
PSM & 0.356/0.346 & 0.558/0.549 & 0.359/0.349 & 0.557/0.549 & 0.360/0.350 & 0.558/0.549 \\
PUMP & 0.118/0.113 & 0.451/0.415 & 0.222/0.212 & 0.614/0.580 & 0.216/0.205 & 0.603/0.570 \\
SKAB & 0.441/0.438 & 0.586/0.581 & 0.679/0.676 & 0.747/0.743 & 0.665/0.661 & 0.743/0.738 \\
SMAP & 0.119/0.121 & 0.442/0.445 & 0.136/0.139 & 0.482/0.484 & 0.136/0.138 & 0.480/0.483 \\
SMD & 0.161/0.157 & 0.762/0.755 & 0.159/0.154 & 0.781/0.773 & 0.159/0.153 & 0.780/0.771 \\
SVDB & 0.166/0.130 & 0.708/0.669 & 0.200/0.169 & 0.781/0.756 & 0.192/0.162 & 0.770/0.744 \\
SWAN & 0.604/0.598 & 0.701/0.712 & 0.484/0.491 & 0.559/0.568 & 0.516/0.522 & 0.597/0.604 \\
SWAT & 0.109/0.095 & 0.288/0.253 & 0.186/0.154 & 0.402/0.367 & 0.175/0.147 & 0.392/0.357 \\
TAO & 0.900/0.893 & 0.970/0.971 & 0.831/0.824 & 0.945/0.948 & 0.815/0.811 & 0.944/0.948 \\
TODS & 0.660/0.668 & 0.908/0.916 & 0.619/0.654 & 0.906/0.920 & 0.608/0.641 & 0.900/0.915 \\
\end{longtable}
\endgroup

\raggedbottom
\section{Generation under Partial Code Randomization}
\label{app:sampling-visualizations}

\begin{figure}[H]
\centering
\textbf{Electricity}\par\smallskip
\includegraphics[width=\textwidth]{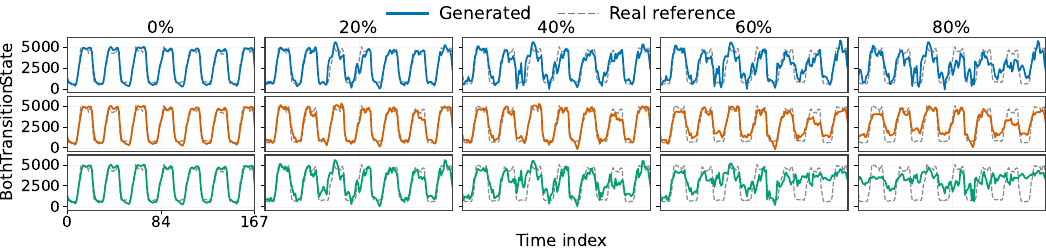}\par\medskip
\textbf{Solar}\par\smallskip
\includegraphics[width=\textwidth]{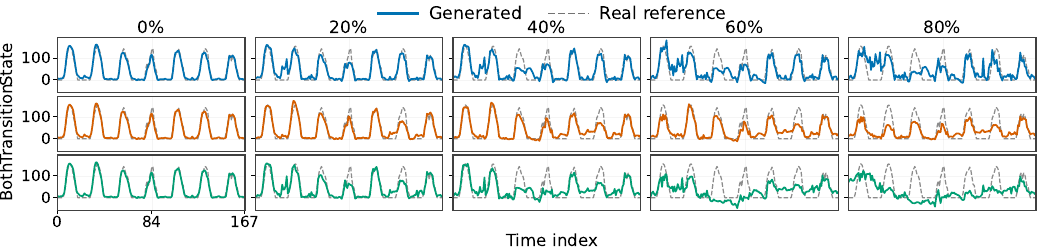}\par\medskip
\textbf{Wind}\par\smallskip
\includegraphics[width=\textwidth]{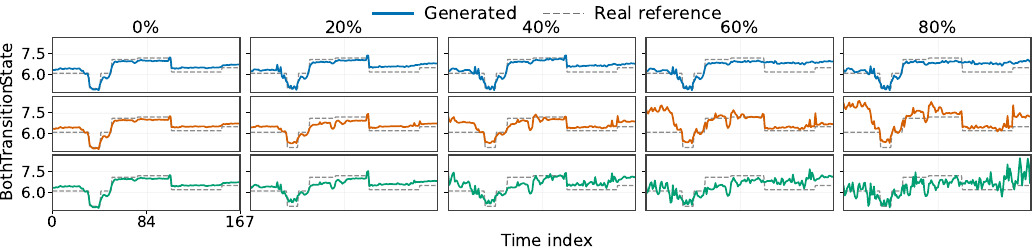}\par\medskip
\textbf{Traffic}\par\smallskip
\includegraphics[width=\textwidth]{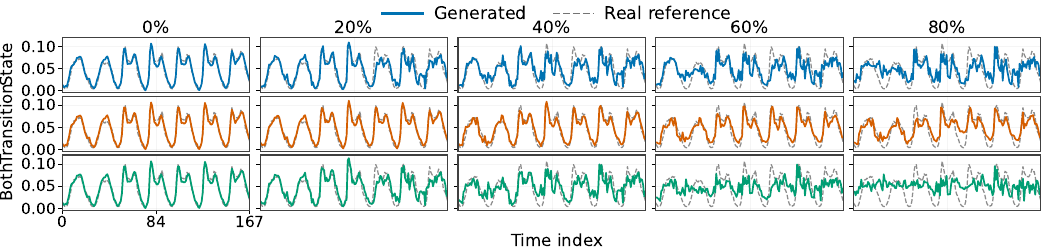}\par\medskip
\caption{Generation under partial code randomization on Electricity, Solar, Wind, and Traffic. Rows randomize states, transitions, or both; columns show replacement fractions. Colored curves are generated sequences; dashed gray curves are references.}
\label{fig:partial-randomization-1}
\end{figure}

\begin{figure}[H]
\centering
\textbf{Taxi}\par\smallskip
\includegraphics[width=\textwidth]{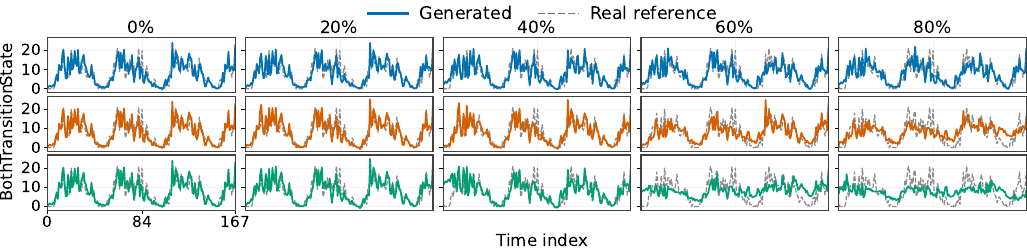}\par\medskip
\textbf{Pedestrian}\par\smallskip
\includegraphics[width=\textwidth]{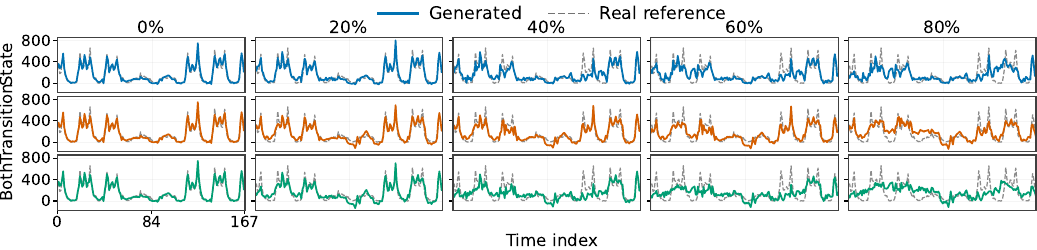}\par\medskip
\textbf{Air (KDDCup)}\par\smallskip
\includegraphics[width=\textwidth]{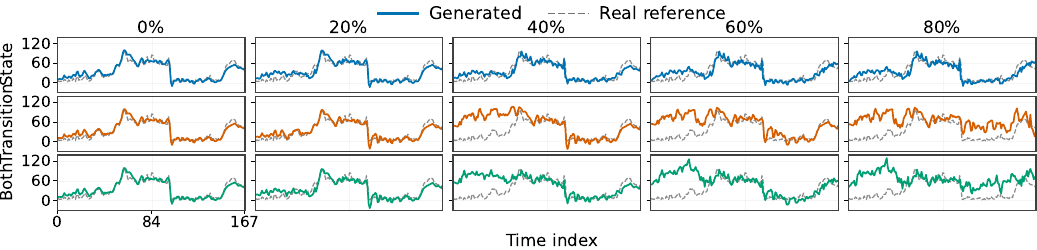}\par\medskip
\textbf{Temperature}\par\smallskip
\includegraphics[width=\textwidth]{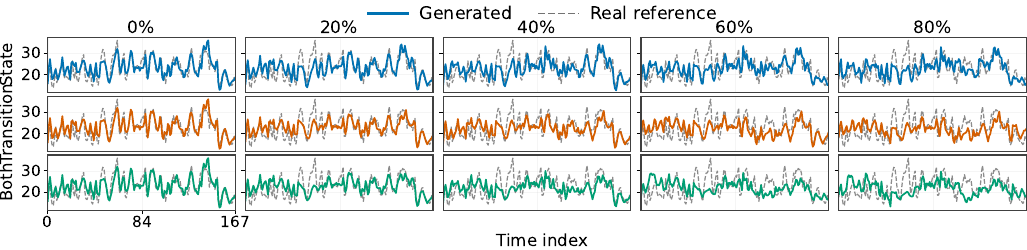}\par\medskip
\caption{Generation under partial code randomization on Taxi, Pedestrian, Air (KDDCup), and Temperature.}
\label{fig:partial-randomization-2}
\end{figure}

\begin{figure}[H]
\centering
\textbf{Rain}\par\smallskip
\includegraphics[width=\textwidth]{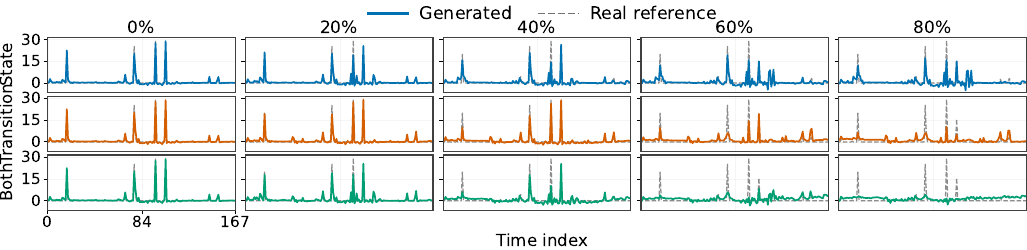}\par\medskip
\textbf{NN5}\par\smallskip
\includegraphics[width=\textwidth]{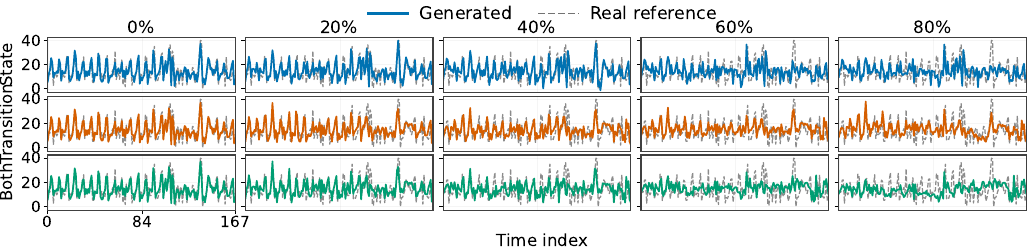}\par\medskip
\textbf{Fred-MD}\par\smallskip
\includegraphics[width=\textwidth]{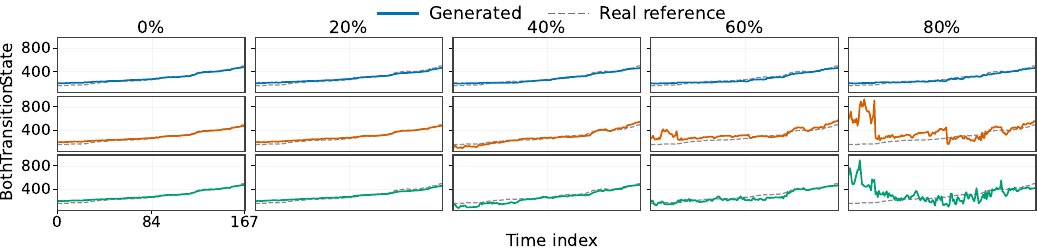}\par\medskip
\textbf{Exchange}\par\smallskip
\includegraphics[width=\textwidth]{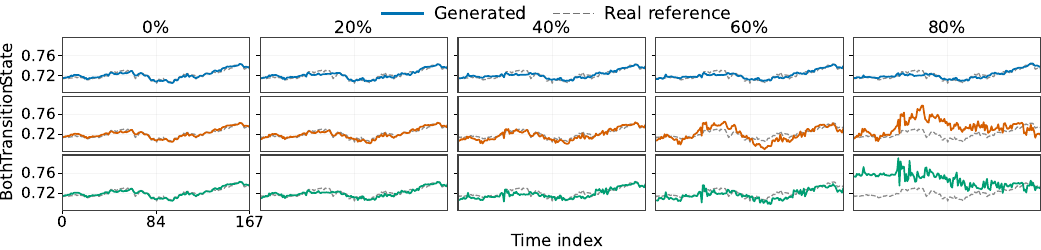}\par\medskip
\caption{Generation under partial code randomization on Rain, NN5, Fred-MD, and Exchange.}
\label{fig:partial-randomization-3}
\end{figure}

\revisionblue{Figures~\ref{fig:partial-randomization-1}--\ref{fig:partial-randomization-3} examine how generation responds to increasing disruption of reference-derived state and transition information. At selected positions, code IDs are replaced uniformly from the full codebook, with replacement fractions from 0\% to 80\%. The three rows perturb states, transitions, or both; single-stream interventions retain the other stream. Each dataset uses the first reference example, with the reference and Gaussian source fixed across panels. The Solar example illustrates changes in peak shape and amplitude as replacement increases, with joint randomization visibly disrupting the reference pattern. These comparisons show the generator's dependence on both streams.}

\section{Limitations}
\label{app:limitations}

\revisionblue{A single MarkovTS generation call supports 512--4096 time steps. Longer generated training counterparts are constructed through segment-wise generation followed by generated-only concatenation, as described in Appendix~\ref{app:experimental-details}. The main distributional benchmark evaluates length-168 windows cut from length-512 outputs and therefore does not assess full-length-512 fidelity or the fidelity of longer concatenated generated sequences. Moreover, the representation remains patch-based, so aggregation within patches can smooth fine-grained within-patch variation.}

\end{document}